\documentclass[journal]{IEEEtran}

\usepackage{times}
\usepackage{epsfig}
\usepackage{graphicx}
\usepackage{amsmath}
\usepackage{amssymb}
\usepackage{subfigure}
\usepackage{xspace}
\usepackage{color}

\usepackage{algorithmic}
\usepackage{algorithm}

\newcommand{\norm}[1]{\left\Vert#1\right\Vert}
\newcommand{\abs}[1]{\left\vert#1\right\vert}

\newcommand\vect[1]{{\bf#1}}
\newcommand\matr[1]{{\bf#1}}

\newcommand\alphabf{{\boldsymbol{\alpha}}}
\newcommand\betabf{{\boldsymbol{\beta}}}
\newcommand{\argmin}{\operatornamewithlimits{argmin}}
\newcommand{\argmax}{\operatornamewithlimits{argmax}}

\newcommand\RR[1]{\mathbb{R}^{#1}}
\DeclareRobustCommand\onedot{\futurelet\@let@token\@onedot}
\def\@onedot{\ifx\@let@token.\else.\null\fi\xspace}

\def\OM {{\mathbf{\Omega}}} 
\def\DIF {{\OM_{\text{\tiny{DIF}}}}} 
\DeclareMathOperator{\diag}{diag}

\newtheorem{thm}{Theorem}[section]

\newtheorem{defn}[thm]{Definition}

\newcommand{\rg}[1]{\textcolor{black}{#1}}

\usepackage{authblk} 

\begin{document}

\title{Sparsity Based Methods for Overparameterized Variational Problems}

\author[*]{Raja Giryes}
\author[**]{Michael Elad}
\author[**]{Alfred M. Bruckstein}
\affil[*]{Department of Electrical and Computer Engineering, 
Duke University, 
Durham, North Carolina, 27708, USA.
{\tt\small raja.giryes@duke.edu}.}
\affil[**]{Department of Computer Science, The Technion - Israel Institute of Technology, 
Haifa, 32000, Israel. ~~~~~ 
{\tt\small \{elad, freddy\}@cs.technion.ac.il}}

\maketitle

\begin{abstract}
Two complementary approaches have been extensively used in signal and image processing leading to novel results, the sparse representation methodology and the variational strategy. Recently, a new sparsity based model has been proposed, the cosparse analysis framework, \rg{which may potentially help in bridging sparse approximation based methods to the traditional total-variation minimization.} Based on this, we introduce a sparsity based framework for solving overparameterized variational problems.  The latter has been used to improve the estimation of optical flow and also for general denoising of signals and images. However, the recovery of the space varying parameters involved was not adequately \rg{addressed} by traditional variational methods. 
We first demonstrate the efficiency of the new framework for one dimensional signals in recovering a piecewise linear and polynomial function. 
Then, we illustrate how the new technique can be used for denoising and segmentation of images.
\end{abstract}

\section{Introduction}

Many successful signal and image processing techniques rely on the fact that the given signals or images of interest belong to a class described by a certain a priori known model. Given the model, the signal is  processed by estimating the ``correct'' parameters of the model. For example, in the sparsity framework the assumption is that the signals belong to a union of low dimensional subspaces \cite{Bruckstein09From, Gribonval03spars, Blumensath09Sampling, Lu08Theory}. In the variational strategy, a model is imposed on the variations of the signal, e.g., its derivatives are required to be smooth \cite{Perona90Scale, Rudin92Nonlinear, Caselles97Geodesic,Weickert98Anisotropic}.

Though both sparsity-based and variational-based approaches are widely used for signal processing and computer vision, they are often viewed as two different methods with little in common between them. 
One of the well known variatonal tools is the total-variation regularization, used mainly for denoising and inverse problems. 
It can be formulated as \cite{Rudin92Nonlinear}
\begin{eqnarray}
\label{eq:ROF}
\min_{\tilde{\vect{f}}}\norm{\vect{g} - \matr{M}\tilde{\vect{f}}}_2^2 + \lambda \norm{{\nabla \tilde{\vect{f}}}}_1,
\end{eqnarray} 
where $\vect{g} = \matr{M}\matr{f}+\vect{e} \in \RR{m}$ are the \rg{given} noisy measurements, $\matr{M} \in \RR{m \times d}$ is a measurement matrix, $\vect{e} \in \RR{m}$ is an additive \rg{(typically white Gaussian)} noise, $\lambda$ is a regularization parameter, $\vect{f} \in \RR{d}$ is the original unknown signal \rg{to be recovered,} and ${\nabla \vect{f}}$ is its gradients vector.

The anisotropic version of \eqref{eq:ROF}, \rg{which we will use in this work,} is
\begin{eqnarray}
\label{eq:TV}
\min_{\tilde{\vect{f}}}\norm{\vect{g} - \matr{M}\tilde{\vect{f}}}_2^2 + \lambda \norm{\DIF \tilde{\vect{f}}}_1,
\end{eqnarray} 
where $\DIF$ is the finite-difference operator that returns the derivatives of the signal. In the $1$D case it applies the filter $[1, -1]$, i.e.,
\begin{eqnarray}
\DIF = \OM_{\text{\tiny{1D-DIF}}}=  \left[\begin{array}{cccccc}
1 & -1 & 0 & \dots & \dots & 0  \\
0 & 1 & -1 & \dots & \dots & 0 \\
\vdots & \ddots & \ddots & \ddots & \ddots & \vdots \\
0 & 0 & \ddots &  1 & -1 & 0 \\
0 & 0 & \dots &  \dots & 1 & -1 
\end{array} \right],
\end{eqnarray} 
For images it returns the horizontal and vertical derivatives \rg{using the filters $[1, -1]$ and $[1, -1]^T$ respectively}. Note that for one dimensional signals there is no difference between \eqref{eq:ROF} and \eqref{eq:TV} as the gradient equals the derivative. 
\rg{However, in the 2D case the first (Eq. \eqref{eq:ROF}) considers the sum of gradients (square root of the squared sum of the directional derivatives), while the second (Eq. \eqref{eq:TV}) considers the absolute sum of the directional derivatives, approximated by finite differences.}


Recently, a very interesting connection has been drawn between the  total-variation minimization problem and the sparsity model.
It has been shown that \eqref{eq:TV} can be viewed as an $\ell_1$-relaxation technique for approximating signals that are sparse in their derivatives domain, i.e., after applying the operator $\DIF$ on them \cite{Needell13Stable, Nam13Cosparse, Giryes14Greedy}.
Such signals are said to be cosparse under the operator $\DIF$ in the analysis (co)sparsity model \cite{Nam13Cosparse}.

Notice that the TV regularization is only one example from the variational framework. Another recent technique, \rg{which is the focus of this paper,} is the overparameterization \rg{idea, which} represents the signal as a combination of
known functions weighted by  space-variant parameters of the model \cite{Nir07Over, Nir08Over}.

\rg{Let us introduce this overparameterized model via an example. 
If a 1D signal $\vect{f}$ is known to be piecewise linear, its
 $i$-th element  can be written as $\vect{f}(i) = \vect{a}(i) + \vect{b}(i)i$, where $\vect{a}(i)$ and $\vect{b}(i)$ are the local coefficients 
describing the local line-curve. As such, the vectors $\vect{a}$ and $\vect{b}$ should be piecewise \emph{constant}, with discontinuities in the same locations. Each constant interval in $\vect{a}$ and $\vect{b}$ corresponds to one linear segment in $\vect{f}$. When put in matrix-vector notation, $\vect{f}$ can be written alternatively as 
\begin{eqnarray}
\vect{f} = \vect{a} + \matr{Z}\vect{b},
\end{eqnarray}
where $\matr{Z} \in \RR{ d \times d}$ is  a diagonal matrix with the values $1,2, \dots , d$ on its main diagonal.}  For images this parameterization would \rg{similarly} be $\vect{f}(i,j) = \vect{a}(i,j) + \vect{b}_1(i,j)i + \vect{b}_2(i,j)j$.

This strategy is referred to as \rg{"overparameterization"} because the number of \rg{representation} parameters is \rg{larger} than the signal \rg{size}. 
In the above 1D example, while the original signal contains $d$ unknown values, the recovery problem that seeks $\vect{a}$ and $\vect{b}$ has twice as many variables. 
Clearly, there are many other parameterization options for signals, beyond the linear one.
Such parameterizations have been shown to improve the denoising performance of \rg{the solution of the problem posed in}
\eqref{eq:ROF} in some cases \cite{Nir07Over}, and to provide very high quality results for optical flow estimation
\cite{Nir08Over,Rosman12Over}.

\subsection{Our Contribution}

The true force behind overparameterization is that while it uses more variables than needed for representing the signals, these are often more naturally suited to describe its structure. For example, if a signal is piecewise linear  then we may impose a constraint on the overparameterization coefficients $\vect{a}$ and $\vect{b}$ to be piecewise constant. 

Note that piecewise constant signals are sparse under the $\DIF$ operator. Therefore, for each of the coefficients we can use the tools developed in the analysis sparsity model \cite{Giryes14Greedy, Giryes11Iterative, Giryes12CoSaMP,  Peleg12Performance, Candes11Compressed}. However, in our case $\vect{a}$ and $\vect{b}$ are jointly sparse, i.e., their change points are collocated and therefore an extension is necessary.

Constraints on the structure in the sparsity pattern of a representation have already been analyzed in the literature. They are commonly referred to as joint sparsity models, and those are found in the literature, \rg{both in the context of handling groups of signals \cite{Cotter05Sparse, Tropp06Algorithms1, Tropp06Algorithms2, Wipf07Empirical, Mishali08ReMBo, Fornasier08Recovery}, or when considering blocks of non-zeros in a single representation vector \cite{Yuan06Model, Eldar09Robust, Stojnic09Reconstruction, Eldar10Block, Baraniuk10Model}}. 
We use these tools to extend the existing analysis techniques to handle the block sparsity  \rg{in our overparameterized scheme}.

In this paper we introduce a general sparsity based framework for solving overparameterized variational problems.
As the structure of these problems enables segmentation while recovering the signal, \rg{we provide an elegant way for recovering a signal from its deteriorated measurements by using an $\ell_0$ approach, which is accompanied by theoretical guarantees.}
We \rg{demonstrate} the efficiency of the new framework for one dimensional functions in recovering piecewise polynomial signals. Then we \rg{shift our view } to images and demonstrate how the new approach can be used for denoising and segmentation.

\subsection{Organization}

\rg{This paper is organized as follows: 
In Section~\ref{sec:OVP} we present the overparameterized variational model with more details.}
In Section~\ref{sec:analysis} we describe briefly the synthesis and analysis sparsity  models. In Sections~\ref{sec:guarantees} and \ref{sec:overparameterized}  we introduce a new framework for solving overparameterized variational problems using sparsity.
Section~\ref{sec:guarantees} proposes a recovery strategy for the 1D polynomial case based on the SSCoSaMP technique with optimal projections \cite{Davenport13Signal,Giryes14GreedySignal,Giryes14NearOracle}. We provide stable recovery guarantees for this algorithm for the case of an additive adversarial noise and denoising guarantees for the case \rg{of} a zero-mean white Gaussian noise. 
In Section~\ref{sec:overparameterized} we extend our scheme beyond the 1D case to higher dimensional polynomial functions such as images. We employ an extension of the GAPN algorithm \cite{Nam11GAPN} for block sparsity for this task.
 In Section~\ref{sec:exp} we present experiments for linear overparameterization of images and one dimensional signals. We \rg{demonstrate} how the proposed method can be used for image denoising and segmentation.
Section~\ref{sec:conc} concludes our work and proposes future directions of research. 


\section{\rg{The Overparameterized Variational Framework}}
\label{sec:OVP}

Considering again the linear relationship between the measurements and the unknown signal, 
\begin{eqnarray}
\label{eq:meas}
\vect{g} = \matr{M}\vect{f} + \vect{e},
\end{eqnarray}
note that without a prior knowledge on $\vect{f}$ we cannot recover it from $\vect{g}$ if $m < d$ or $\vect{e} \ne 0$. 

\rg{In the variational framework, a regularization is imposed on the variations of the signal $\vect{f}$. One popular strategy for recovering the signal in this framework is by solving the following minimization problem:
\begin{eqnarray}
\label{eq:variational}
\min_{\tilde{\vect{f}}}\norm{\vect{g} - \matr{M}\tilde{\vect{f}}}_2^2 + \lambda \norm{\mathcal{A}(\tilde{\vect{f}})}_{p},
\end{eqnarray} 
where $\lambda$ is the regularization weight and $p \ge 1$ is the type of norm used with the  regularization operator $\mathcal{A}$, which is typically a local operator. For example, for $p=1$ and $\mathcal{A} = \nabla$ we get the TV minimization (Eq. \eqref{eq:TV}). Another example for a regularization operator is the Laplace operator $\mathcal{A} = \nabla^2$. Other types of  regularization operators and variational formulations can be found in \cite{Morel95Variational, Chan05Image, Weickert06Mathematical, aubert06mathematical}.}

\rg{Recently, the overameterized variational framework has been introduced as an extension to the traditional variational methodology \cite{Nir07Over, Nir08Over,Rosman12Over,shem13globally}.
Instead of applying a regularization on the signal itself, it is applied on the coefficients of the signal under a global parameterization of the space. Each element of the signal can be modeled as $\vect{f}(i) = \sum_{j=1}^n \vect{b}_j(i)\vect{x}_j(i)$, where $\left\{\vect{b}_j \right\}_{j=1}^n$ are the coefficients vectors and $\left\{\vect{x}_j \right\}_{j=1}^n$  contain the parameterization basis functions for the space.
}

\rg{Denoting by $\matr{X}_i \triangleq \diag(\vect{x}_i)$ the diagonal matrix that has the vector $\vect{x}_i$ on its diagonal, we can rewrite the above as $\vect{f} = \sum_{j=1}^n \matr{X}_j\vect{b}_j$. With these notations, the overparameterized minimization problem becomes
\begin{eqnarray}
\label{eq:variational_ovp}
\min_{\tilde{\vect{b}}_i, 1\le i \le n}\norm{\vect{g} - \matr{M}\sum_{i=1}^n\matr{X}_i\tilde{\vect{b}}_i}_2^2 +  \sum_{i=1}^n \lambda_i\norm{\mathcal{A}_i(\tilde{\vect{b}}_i)}_{p_i},
\end{eqnarray} 
where each coefficient $\vect{b}_i$ is regularized separately by the operator $\mathcal{A}_i$ (which can be the same one for all the coefficients)\footnote{\rg{We note that it is possible to have more than one regularization for each coefficient, as practiced in \cite{shem13globally}.}}. 
}

\rg{Returning to the example of a linear overparameterization, we have that $\vect{f} = \vect{a} + \matr{Z}\vect{b}$, where in this case $\matr{X}_1 = \matr{I}$ (the identity matrix) and $\matr{X}_2 = \matr{Z} = \diag(1, \dots, d)$, a diagonal matrix with $1, \dots, d$ on its diagonal. 
If $\vect{f}$ is a piecewise linear function then the coefficients vectors $\vect{a}$ and $\vect{b}$ should be piecewise constant, and therefore it would be natural to regularize these coefficients with the gradient operator. This leads to  the following minimization problem:
\begin{eqnarray}
\label{eq:variational_linear_ovp}
\min_{\tilde{\vect{a}}, \tilde{\vect{b}}}\norm{\vect{g} - \matr{M}\left(\tilde{\vect{a}} + \matr{Z}\tilde{\vect{b}} \right)}_2^2 + \lambda_1 \norm{\nabla\tilde{\vect{a}}}_{1} + \lambda_2 \norm{\nabla\tilde{\vect{b}}}_{1},
\end{eqnarray} 
which is a special case of \eqref{eq:variational_ovp}. The two main advantages of using the overparameterized formulation are these: (i) the new unknowns have a simpler form (e.g. a piecewise linear signal is treated by piecewise constant unknowns), and thus are easier to recover; and (ii) this formulation leads to recovering the parameters of the signal along with the signal itself.}

\rg{The overparametrization idea, as introduced in \cite{Nir07Over, Nir08Over,Rosman12Over,shem13globally} builds upon the vast work in signal processing that refers to variational methods. As such, there are no known guarantees for the quality of the recovery of the signal, when using the formulation posed in  \eqref{eq:variational_linear_ovp} or its variants. Moreover, it has been shown in \cite{shem13globally} that even for the case of $\matr{M} = \matr{I}$ (and obviously,  $\vect{e} \ne 0$), a poor recovery is achieved  in recovering $\vect{f}$ and its parameterization coefficients. Note that the same happens even if more sophisticated regularizations are combined and applied on $\vect{a}$, $\vect{b}$, and eventually on $\vect{f}$ \cite{shem13globally}.}

\rg{This leads us to look for another strategy to approach the problem of recovering a piecewise linear function from its deteriorated measurement $\vect{g}$. Before describing our new scheme, we introduce in the next section the sparsity model that will aid us in developing this alternative strategy.}

\section{The Synthesis and Analysis Sparsity Models}
\label{sec:analysis}

\rg{A popular prior for recovering a signal $\vect{f}$ from its distorted measurements (as posed in \eqref{eq:meas}) is the sparsity model \cite{Bruckstein09From, Gribonval03spars}. The idea behind it is that if we know a priori that $\vect{f}$ resides in a union of low dimensional subspaces, which do not intersect trivially  with the null space of $\matr{M}$,  then we can estimate $\vect{f}$ stably by selecting the signal  that belongs to this union of subspaces and is the closest to $\vect{g}$ \cite{Blumensath09Sampling, Lu08Theory}.}

In the classical sparsity model, the signal $\vect{f}$ is assumed to have a sparse representation $\alphabf$ under a given dictionary $\matr{D}$, i.e., $\vect{f} = \matr{D}\alphabf, \norm{\alphabf}_0 \le k$, where $\norm{\cdot}_0$ is the $\ell_0$ pseudo-norm that counts the number of non-zero entries in a vector, and $k$ is the sparsity of the signal. Note that each low dimensional subspace in the standard sparsity model, known also as the synthesis model,  is spanned by a collection of $k$ columns from $\matr{D}$. 
With this model we can recover $\vect{f}$ by solving 
\begin{eqnarray}
\label{eq:P0_synthesis_k}
\min_{\alphabf} \norm{\vect{g} - \matr{M}\matr{D}\alphabf}_2^2 &s.t. & \norm{\alphabf}_0 \le k,
\end{eqnarray}
if $k$ is known, or 
\begin{eqnarray}
\label{eq:P0_synthesis_e}
\min_{\alphabf}  \norm{\alphabf}_0 &s.t. & \norm{\vect{g} - \matr{M}\matr{D}\alphabf}_2^2 \le \norm{\vect{e}}_2^2,
\end{eqnarray}
if we have information about the energy of the noise $\vect{e}$.
Obviously, once we get $\alphabf$, the desired recovered signal is simply $\matr{D}\alphabf$.
As both of these minimization problems are NP-hard \cite{Davis97Adaptive-greedy}, many approximation techniques have been proposed to approximate their solution, accompanied with recovery guarantees that depend on the properties of the matrices $\matr{M}$ and $\matr{D}$. These include $\ell_1$-relaxation  \cite{elad07Analysis,Chen98overcomplete,Donoho06OnTheStability}, known also as LASSO \cite{Tibshirani96Regression}, matching pursuit (MP) \cite{MallatZhang93}, 
orthogonal matching pursuit (OMP) \cite{Chen89Orthogonal, Davis94Adaptive}, compressive sampling matching pursuit (CoSaMP) \cite{Needell09CoSaMP}, subspace pursuit (SP) \cite{Dai09Subspace}, iterative hard
thresholding (IHT) \cite{Blumensath09Iterative} and hard thresholding pursuit (HTP) \cite{Foucart11Hard}.

Another framework for modeling a union of low dimensional subspaces is the analysis one \cite{Nam13Cosparse, elad07Analysis}. This model considers the behavior of $\OM\vect{f}$, the signal after applying a given operator $\OM $ on it, and assumes that this vector is sparse. Note that here the zeros are those that characterize the subspace in which $\vect{f}$ resides, as each zero in $\OM\vect{f}$ corresponds to a row in $\OM$ to which $\vect{f}$ is orthogonal to. Therefore,  $\vect{f}$ resides in a subspace orthogonal to the one spanned by these rows.
We say that $\vect{f}$ is cosparse under $\OM$ with a cosupport $\Lambda$ if $\OM_\Lambda\vect{f} =0$, where $\OM_\Lambda$ is a sub-matrix of $\OM$ with the rows corresponding to the set $\Lambda$.

The analysis variants of \eqref{eq:P0_synthesis_k} and \eqref{eq:P0_synthesis_e} for estimating $\vect{f}$ are 
\begin{eqnarray}
\label{eq:P0_analysis_k}
\min_{\tilde{\vect{f}}} \norm{\vect{g} - \matr{M}\tilde{\vect{f}}}_2^2 &s.t. & \norm{\OM\tilde{\vect{f}}}_0 \le k,
\end{eqnarray}
where $k$ is the number of non-zeros in $\OM\vect{f}$,
and
\begin{eqnarray}
\label{eq:P0_analysis_e}
\min_{\tilde{\vect{f}}}  \norm{\OM\tilde{\vect{f}}}_0 &s.t. & \norm{\vect{g} - \matr{M}\tilde{\vect{f}}}_2^2 \le \norm{\vect{e}}_2^2.
\end{eqnarray}
As in the synthesis case, these minimization problems  are also NP-hard \cite{Nam13Cosparse} and approximation techniques have been proposed including  Greedy Analysis Pursuit (GAP) \cite{Nam13Cosparse}, GAP noise (GAPN) \cite{Nam11GAPN}, analysis CoSAMP (ACoSaMP),  analysis SP (ASP), analysis IHT (AIHT) and analysis HTP (AHTP) \cite{Giryes14Greedy}.

\section{Overparameterization via the Analysis Sparsity Model}
\label{sec:guarantees}

\rg{With the sparsity models now defined}, we revisit the overparameterization variational problem. If we know that our signal $\vect{f}$ is piecewise linear, then it is clear that the coefficients parameters should be  piecewise constant with the same \rg{discontinuity locations}, when linear overparameterization is used. \rg{We denote by $k$ the number of these discontinuity locations.}

\rg{As a reminder we rewrite $\vect{f} = \left[ \matr{I}, \matr{Z} \right] \left[\vect{a}^T, \vect{b}^T \right]^T$.} Note that $\vect{a}$ and $\vect{b}$ are jointly sparse under $\DIF$, i.e, $\DIF\vect{a}$ and $\DIF\vect{b}$ have the same non-zero locations. With this observation we can extend the analysis minimization problem 
\eqref{eq:P0_analysis_k} to support the structured sparsity in the vector $\left[ \vect{a}^T, \vect{b}^T\right]^T$, leading to the following minimization problem:
\begin{eqnarray}
\label{eq:P0_analysis_k_linear_overparam}
&& \hspace{-0.5in} \min_{\vect{a}, \vect{b}} \norm{\vect{g} - \matr{M}\left[\matr{I}, \matr{Z}\right] \left[ \begin{array}{c}
\vect{a} \\ \vect{b}
\end{array}\right]}_2^2 \\ \nonumber && \hspace{0.8in} s.t.  ~~ \norm{\abs{\DIF\vect{a}}+\abs{ \DIF \vect{b}}}_0 \le k,
\end{eqnarray}
where $\abs{\DIF\vect{a}}$ denotes applying element-wise absolute value on the entries of $\DIF\vect{a}$.

Note that we can have a similar formulation for this problem also in the synthesis framework using the Heaviside dictionary 
\begin{eqnarray}
\matr{D}_{HS} = \left[\begin{array}{ccccc}
1 & 1 & \dots & 1 & 1  \\
0 & 1 & \dots & \ddots & 1 \\
\vdots & 0 & \ddots & \ddots & \vdots \\
\vdots & \ddots & \ddots & 1 & 1 \\
0 & 0 & \dots & 0 & 1  
\end{array} \right],
\end{eqnarray} 
whose atoms are step functions of different length.
We use the known observation that every one dimensional signal with $k$ change points can be sparsely represented using $k+1$ atoms from $\matr{D}_{HS}$ ($k$ columns for representing  the change points plus one for the DC). One way to observe that is by the fact that $\DIF \tilde{\matr{D}}_{HS} = \matr{I}$, where $\tilde{\matr{D}}_{HS}$ is \rg{a submatrix of ${\matr{D}}_{HS}$ obtained by removing the last column of ${\matr{D}}_{HS}$} (the DC component).
Therefore, one may recover the coefficient parameters $\vect{a}$ and $\vect{b}$, by their sparse representations $\alphabf$ and $\betabf$, solving  
\begin{eqnarray}
\label{eq:P0_synthesis_k_linear_overparam}
&& \hspace{-0.5in} \min_{\alphabf, \betabf} \norm{\vect{g} - \matr{M}\left[\matr{I}, \matr{Z} \right] \left[\begin{array}{cc}
\matr{D}_{HS} & 0 \\
0 & \matr{D}_{HS}
\end{array} \right]\left[ \begin{array}{c}
\alphabf \\ \betabf
\end{array}\right]}_2^2 \\ \nonumber && \hspace{0.8in} s.t.  ~~ \norm{\abs{\alphabf}+\abs{ \betabf}}_0 \le k,
\end{eqnarray}
where $\vect{a} = \matr{D}_{HS}\alphabf$ and $\vect{b} = \matr{D}_{HS}\betabf$.
This minimization problem can be approximated using block-sparsity techniques such as the group-LASSO estimator \cite{Yuan06Model}, the mixed-$\ell_2/\ell_1$ relaxation (extension of the $\ell_1$ relaxation) \cite{Eldar09Robust, Stojnic09Reconstruction}, the Block OMP (BOMP) algorithm \cite{Eldar10Block} or the extensions of CoSaMP and IHT for structured sparsity \cite{Baraniuk10Model}. 
\rg{The joint sparsity framework can also be used with \eqref{eq:P0_synthesis_k_linear_overparam}} \cite{Cotter05Sparse, Tropp06Algorithms1, Tropp06Algorithms2, Wipf07Empirical, Mishali08ReMBo, Fornasier08Recovery}.

The problem with the above synthesis techniques is twofold: (i) No recovery guarantees exist for \rg{this formulation} with the dictionary $\matr{D}_{HS}$; (ii) It is hard to generalize the model in \eqref{eq:P0_synthesis_k} to higher order signals, e.g., images. 
 
The reason that no theoretical guarantees are provided for the $\matr{D}_{HS}$ dictionary is the high correlation between its columns. These create  high ambiguity, causing the classical synthesis techniques to fail in recovering the representations $\alphabf$ and $\betabf$. This problem has been addressed in several contributions that have treated the signal directly and not its representation \cite{Davenport13Signal, Giryes14GreedySignal,Giryes14NearOracle, Giryes13CanP0, Giryes13IHTconf, Giryes13OMP}. 

We \rg{introduce an algorithm that approximates the solutions of}  both \eqref{eq:P0_synthesis_k} and \eqref{eq:P0_analysis_k} and
 has theoretical reconstruction performance guarantees for one dimensional functions $\vect{f}$ with matrices $\matr{M}$ that are near isometric for piecewise polynomial functions.
In the next section we \rg{shall present another} algorithm that does not have such guarantees but is generalizable to higher order functions.

Though till now we have restricted our discussion only to piecewise linear functions, we turn now to look at the more general case of piecewise 1D polynomial functions of degree $n$. Note that this method approximates the following minimization problem, which is a generalization of \eqref{eq:P0_analysis_k_linear_overparam} to any polynomial of degree $n$,
\begin{eqnarray}
\label{eq:P0_analysis_k_poly_overparam}
&& \hspace{-0.5in} \min_{\vect{b}_0, \vect{b}_1, \dots, \vect{b}_n} \norm{\vect{g} - \matr{M}\left[\matr{I}, \matr{Z}, \matr{Z}^2, \dots, \matr{Z}^n \right] \left[ \begin{array}{c}
\vect{b}_0 \\ \vect{b}_1 \\ \vdots \\ \vect{b}_n
\end{array}\right]}_2^2 \\ \nonumber && \hspace{1in} s.t.  ~~ \norm{\sum_{i=0}^n\abs{ \DIF \vect{b}_i}}_0 \le k,
\end{eqnarray} 
\rg{where $\abs{\DIF\vect{b}_i}$ is an element-wise operation that calculates the absolute value of each entry in $\DIF\vect{b}_i$.}

We employ the signal space CoSaMP (SSCoSaMP) strategy \cite{Davenport13Signal, Giryes14GreedySignal}\footnote{In a very similar way we could have used the analysis CoSaMP (ACoSaMP) \cite{Giryes14Greedy, Giryes12CoSaMP}.} to approximate the solution of \eqref{eq:P0_analysis_k_poly_overparam}. 
This algorithm assumes the existence of a projection that for a given signal finds its closest signal (in the $\ell_2$-norm sense) that belongs to the model\footnote{Note that in \cite{Davenport13Signal, Giryes14GreedySignal} the projection might be allowed to be near-optimal in the sense that the projection error is close to the optimal error up to a multiplicative constant.}, where in our case the model is  piecewise polynomial functions with $k$ jump points. 
\rg{This algorithm, along with the projection required, are presented in Appendix~\ref{sec:sscosamp}}.

 \begin{figure*}[htb]
\centering
{\subfigure[Noisy Function $\sigma = 0.1$]{\includegraphics[width=0.33\linewidth]{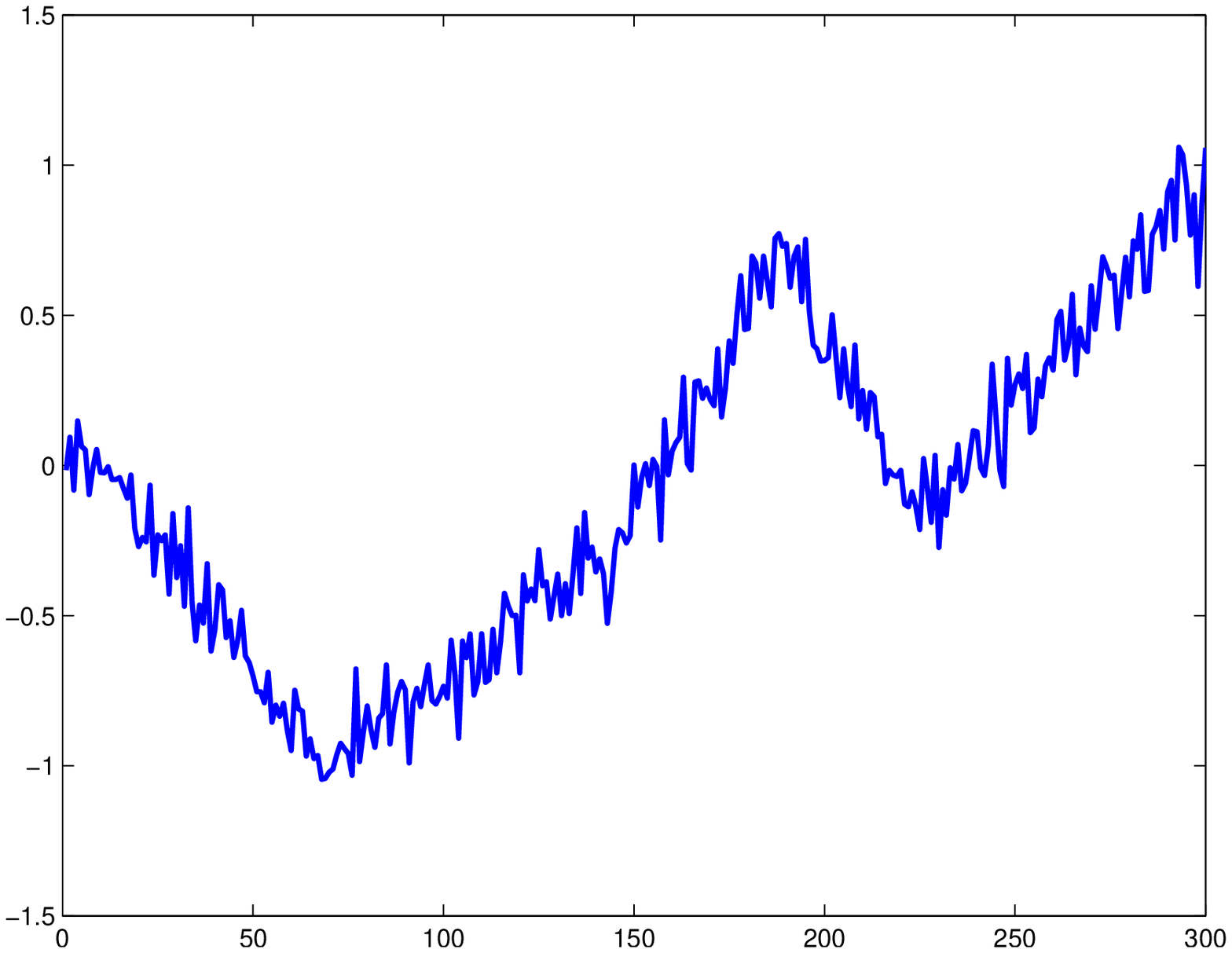}}%
\subfigure[Function Recovery for $\sigma = 0.1$]{\includegraphics[width=0.33\linewidth]{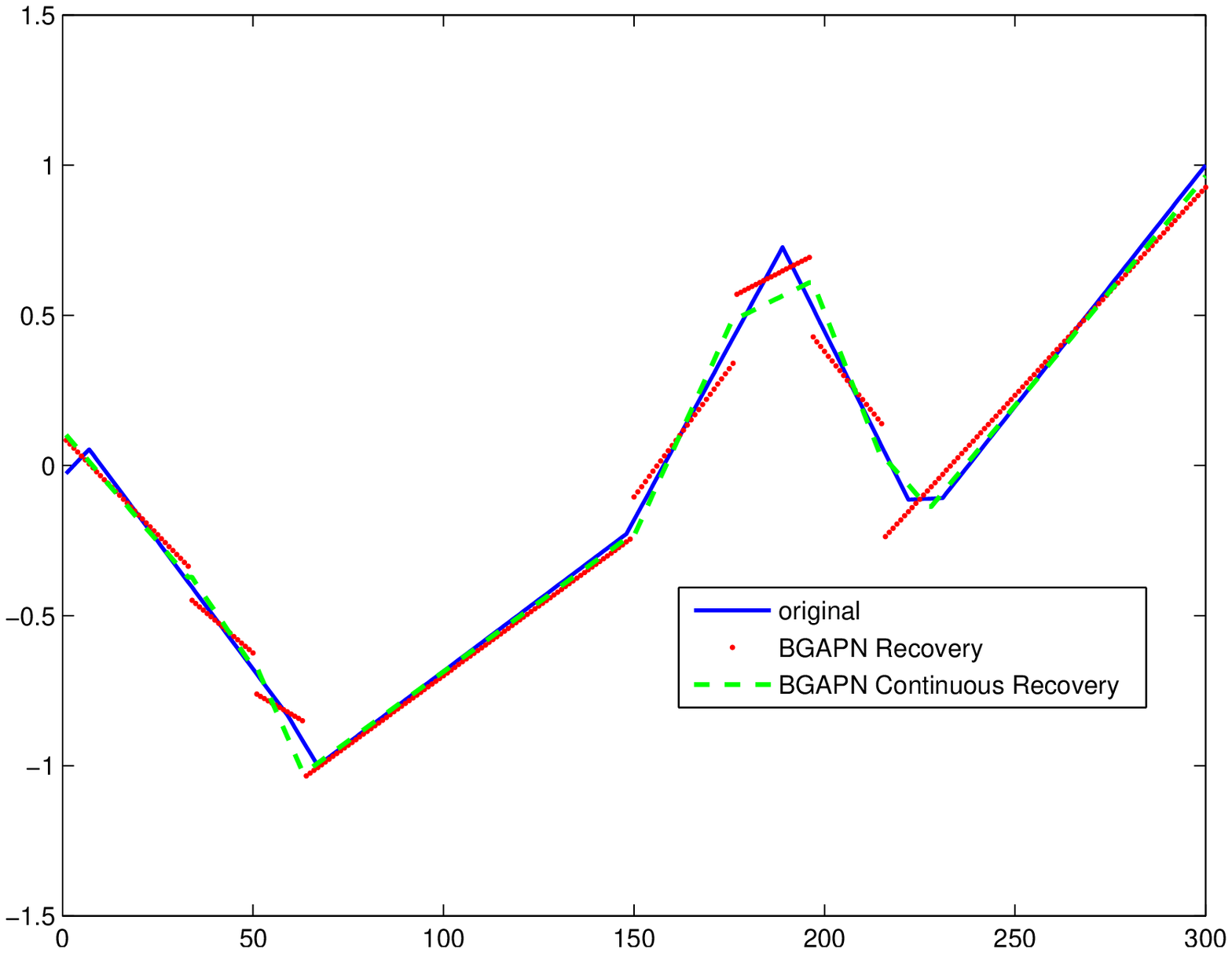}}
\subfigure[Coefficients Parameters Recovery for $\sigma = 0.1$]{\includegraphics[width=0.33\linewidth]{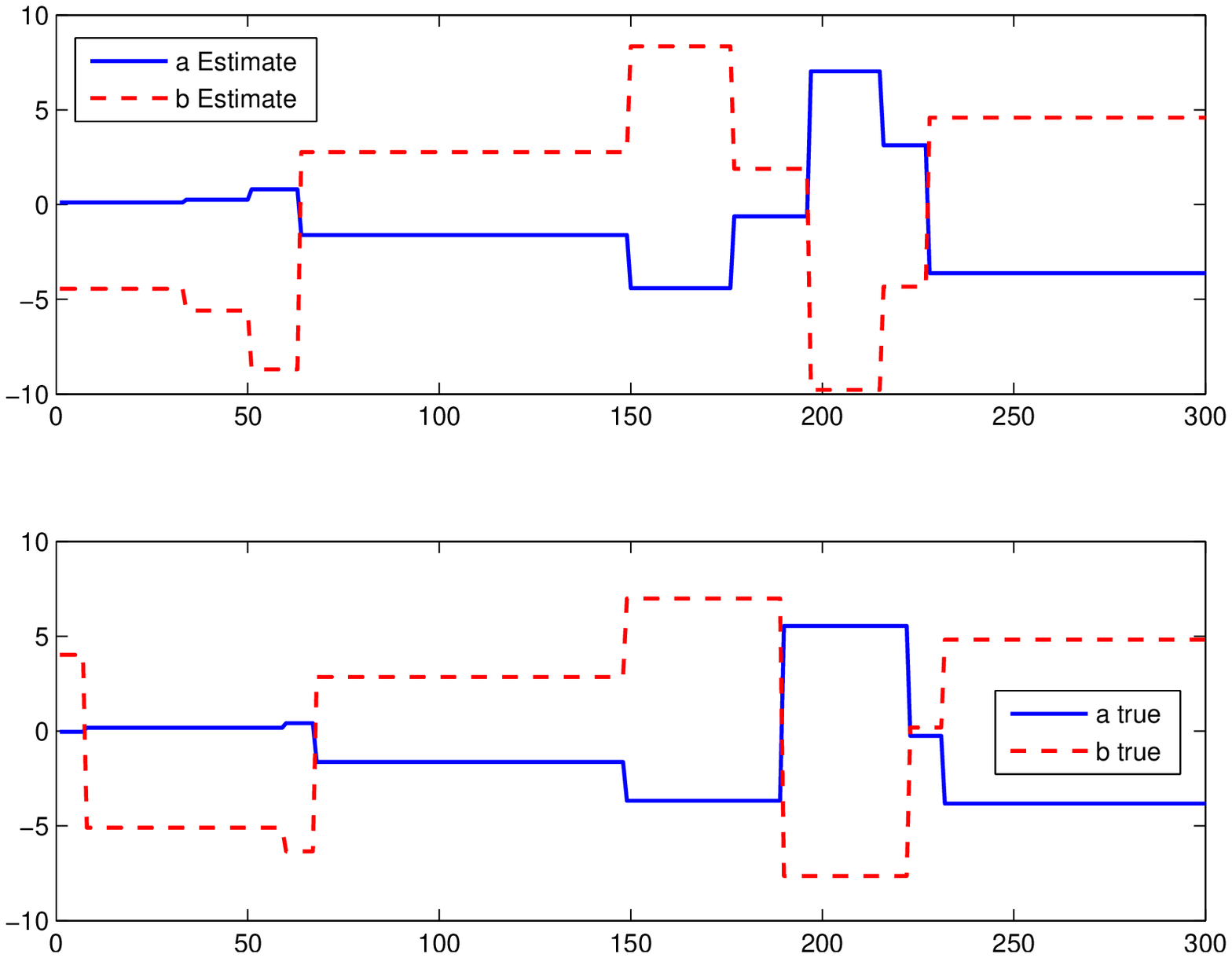}} 
\subfigure[Noisy Function $\sigma = 0.25$]{\includegraphics[width=0.33\linewidth]{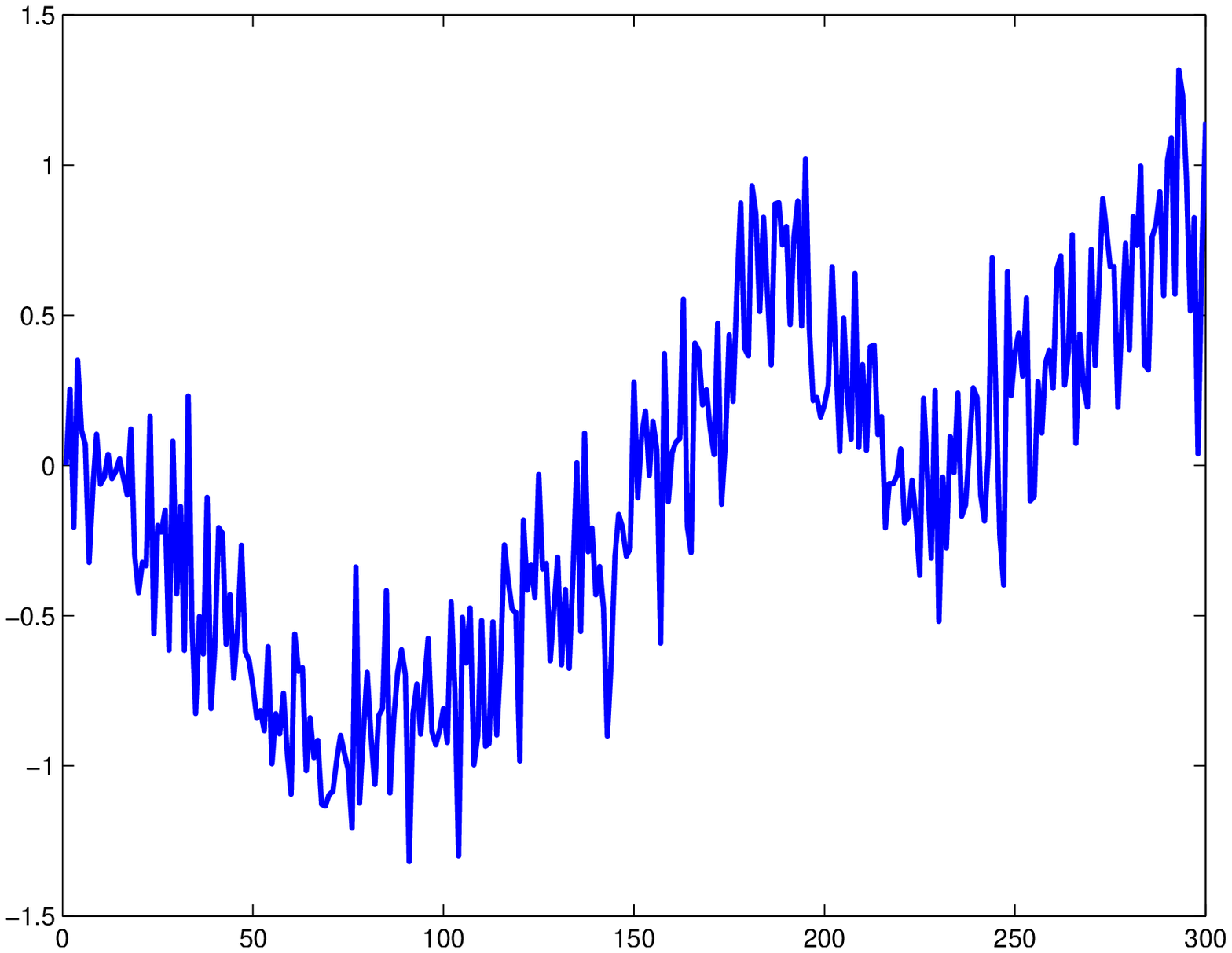}}%
\subfigure[Function Recovery for $\sigma = 0.25$]{\includegraphics[width=0.33\linewidth]{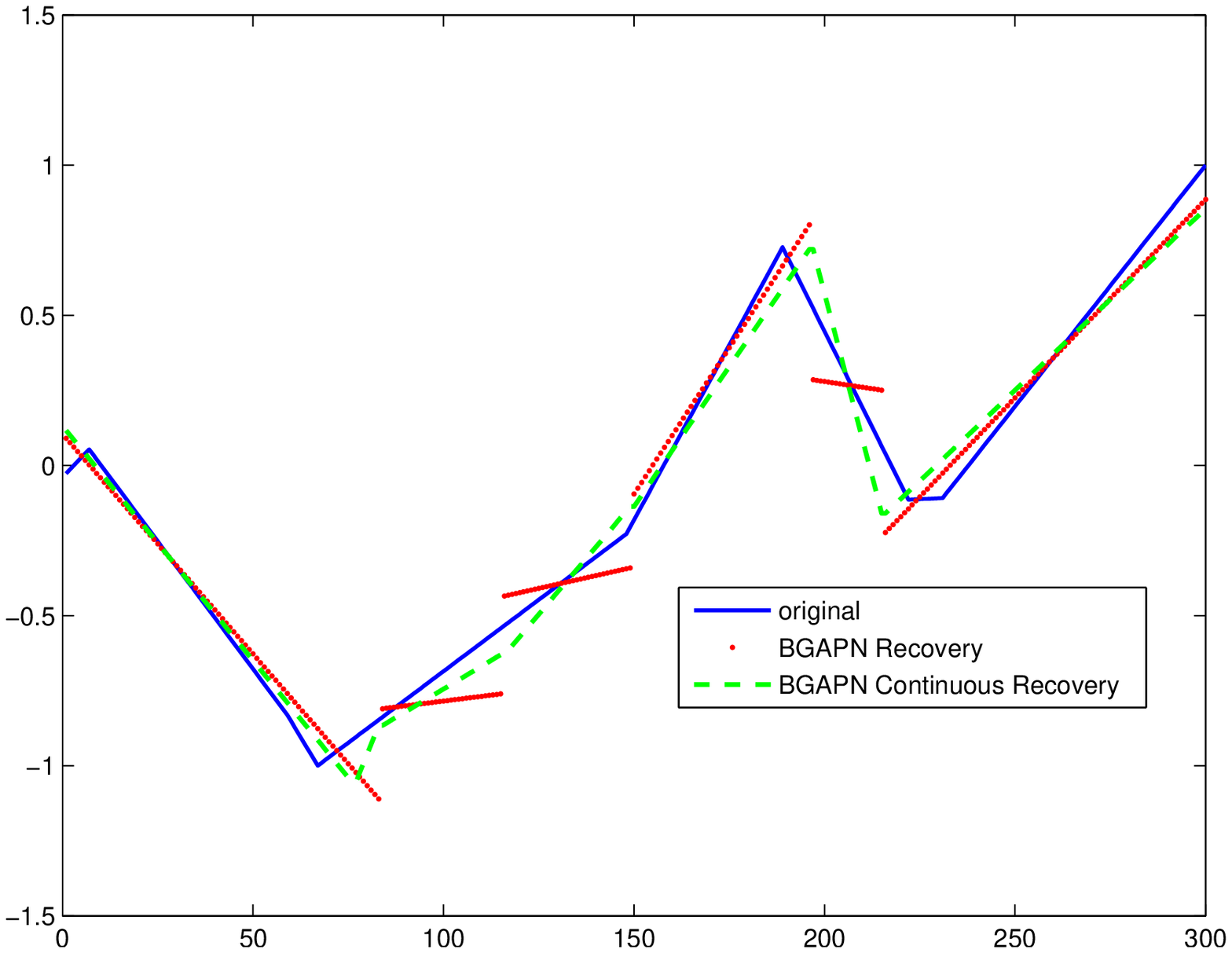}}
\subfigure[Coefficients Parameters Recovery for $\sigma = 0.25$]{\includegraphics[width=0.33\linewidth]{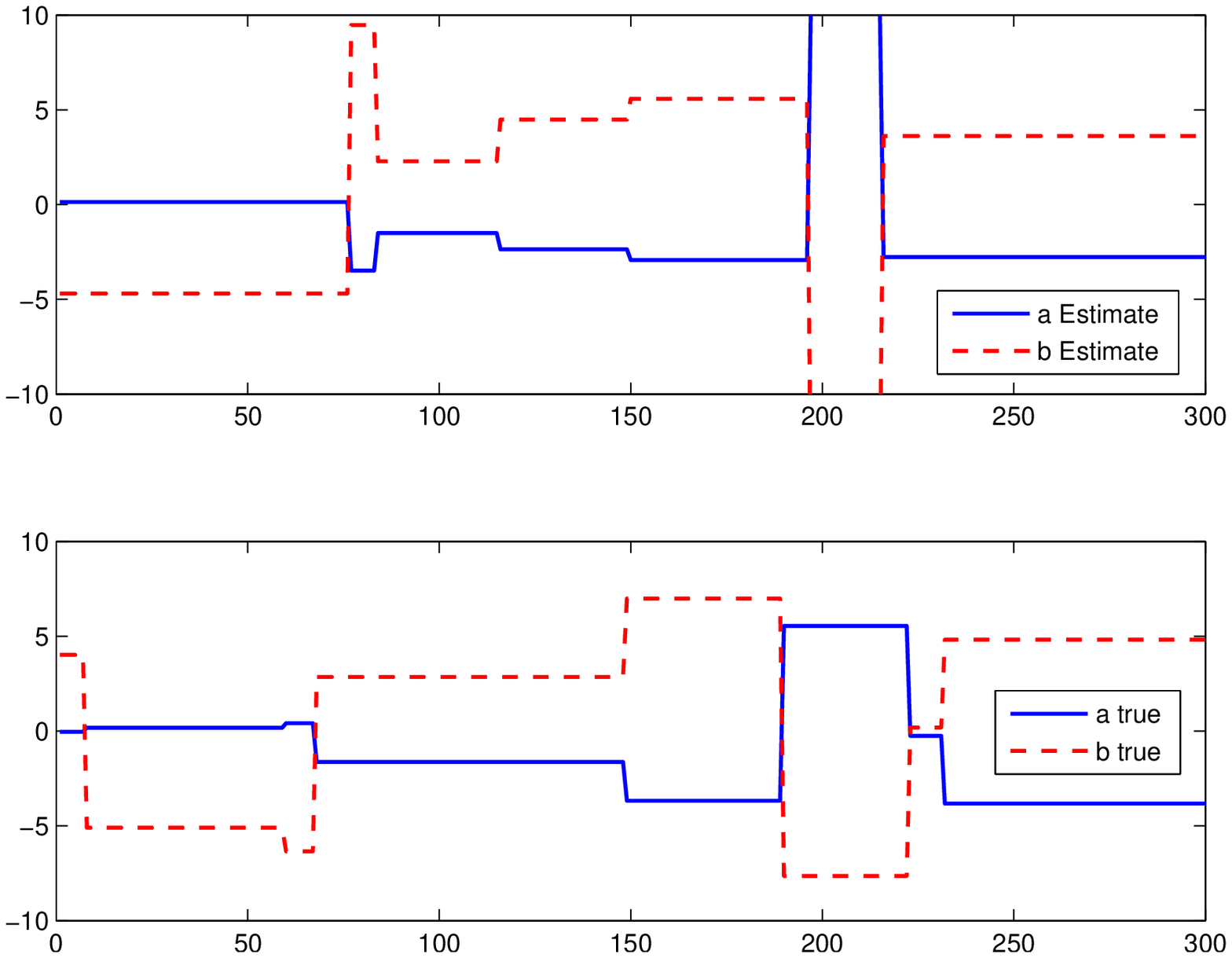}} 
}%
\caption{Recovery of a piecewise linear function using the BGAPN algorithm with and without a constraint on the continuity.}
\label{fig:linear_function_recovery1}
\end{figure*}

\begin{figure*}[htb]
\centering
{\subfigure[Noisy Function $\sigma = 0.1$]{\includegraphics[width=0.33\linewidth]{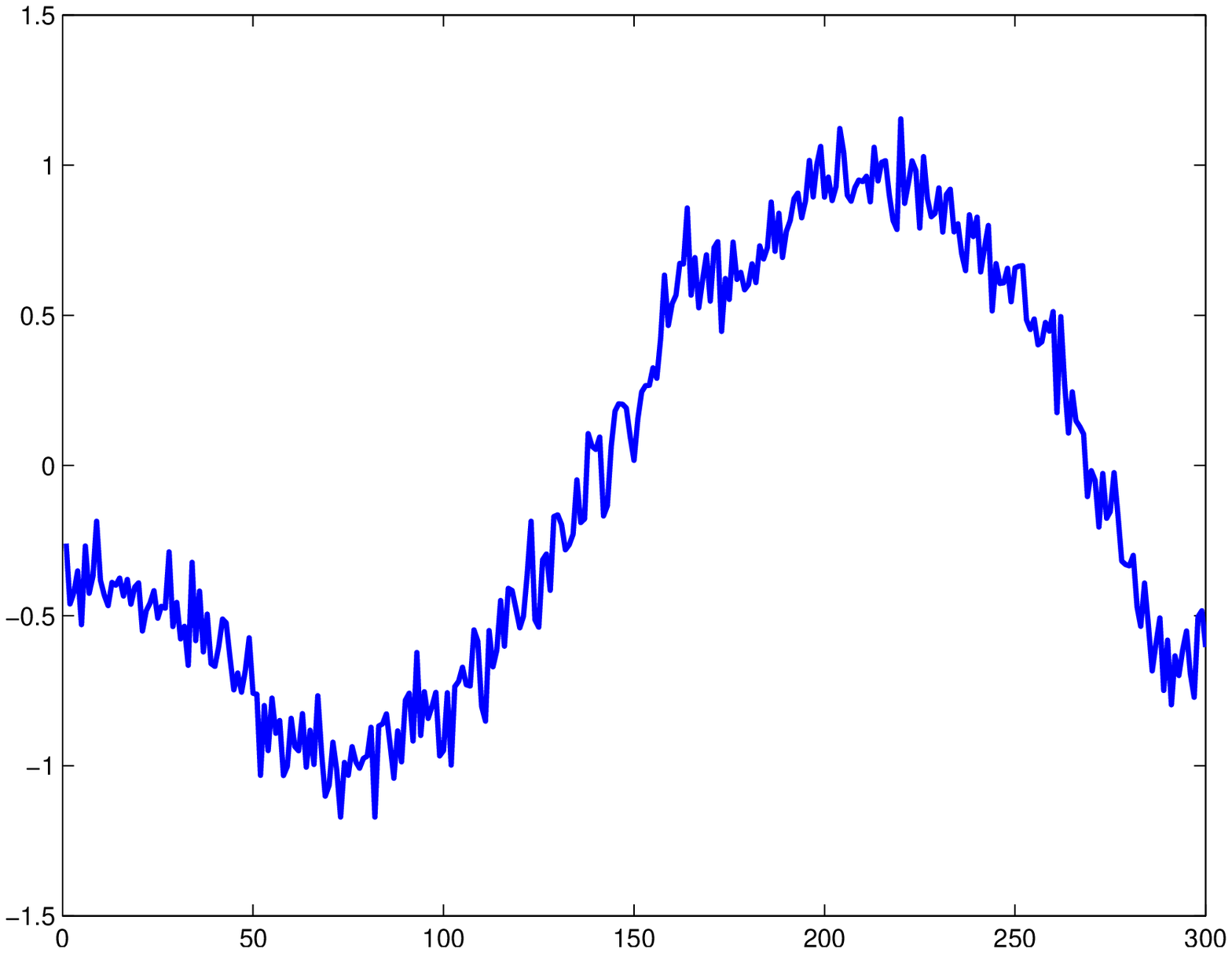}}
\subfigure[Function Recovery for $\sigma = 0.1$]{\includegraphics[width=0.33\linewidth]{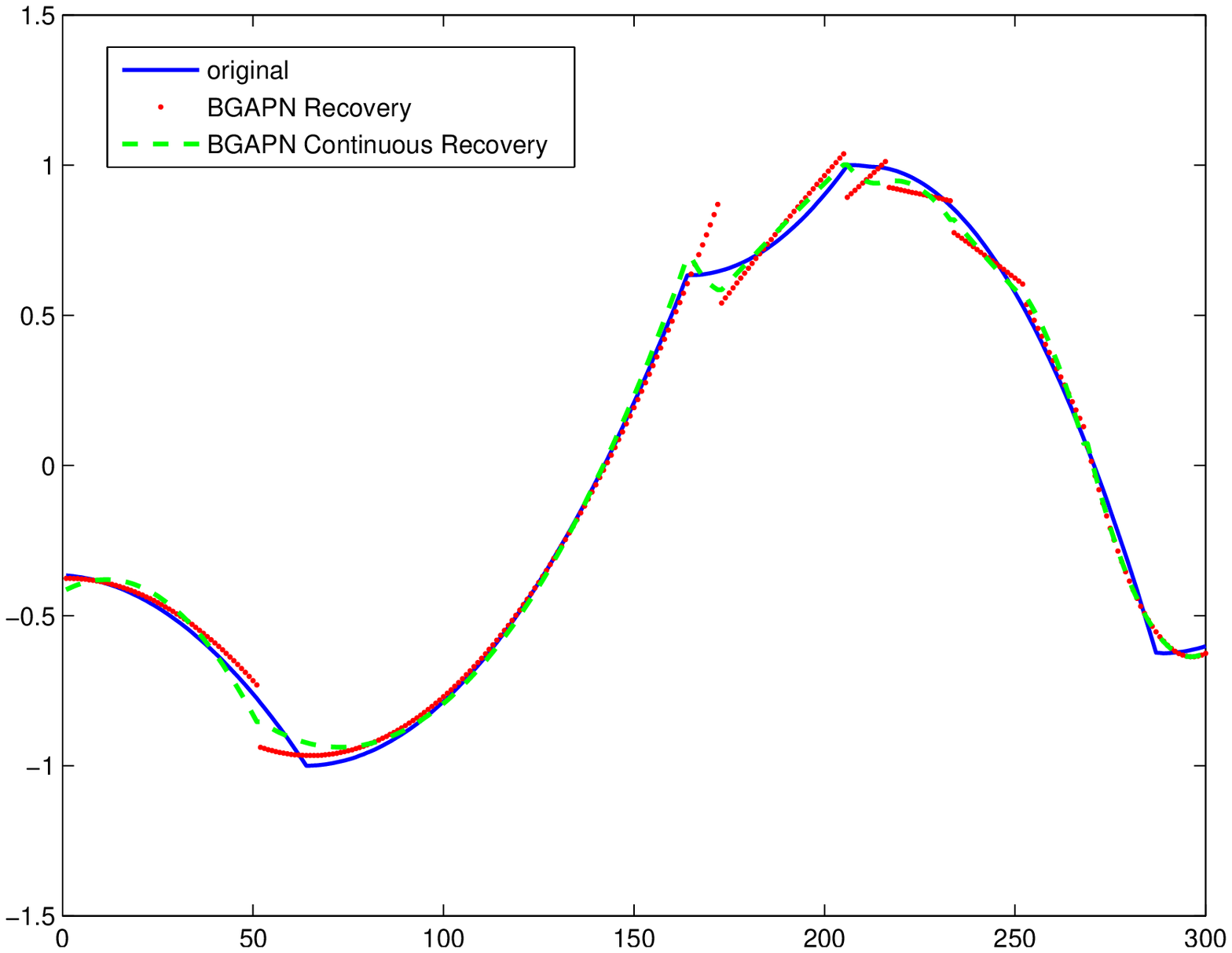}}
\subfigure[Coefficients Parameters Recovery for $\sigma = 0.1$]{\includegraphics[width=0.33\linewidth]{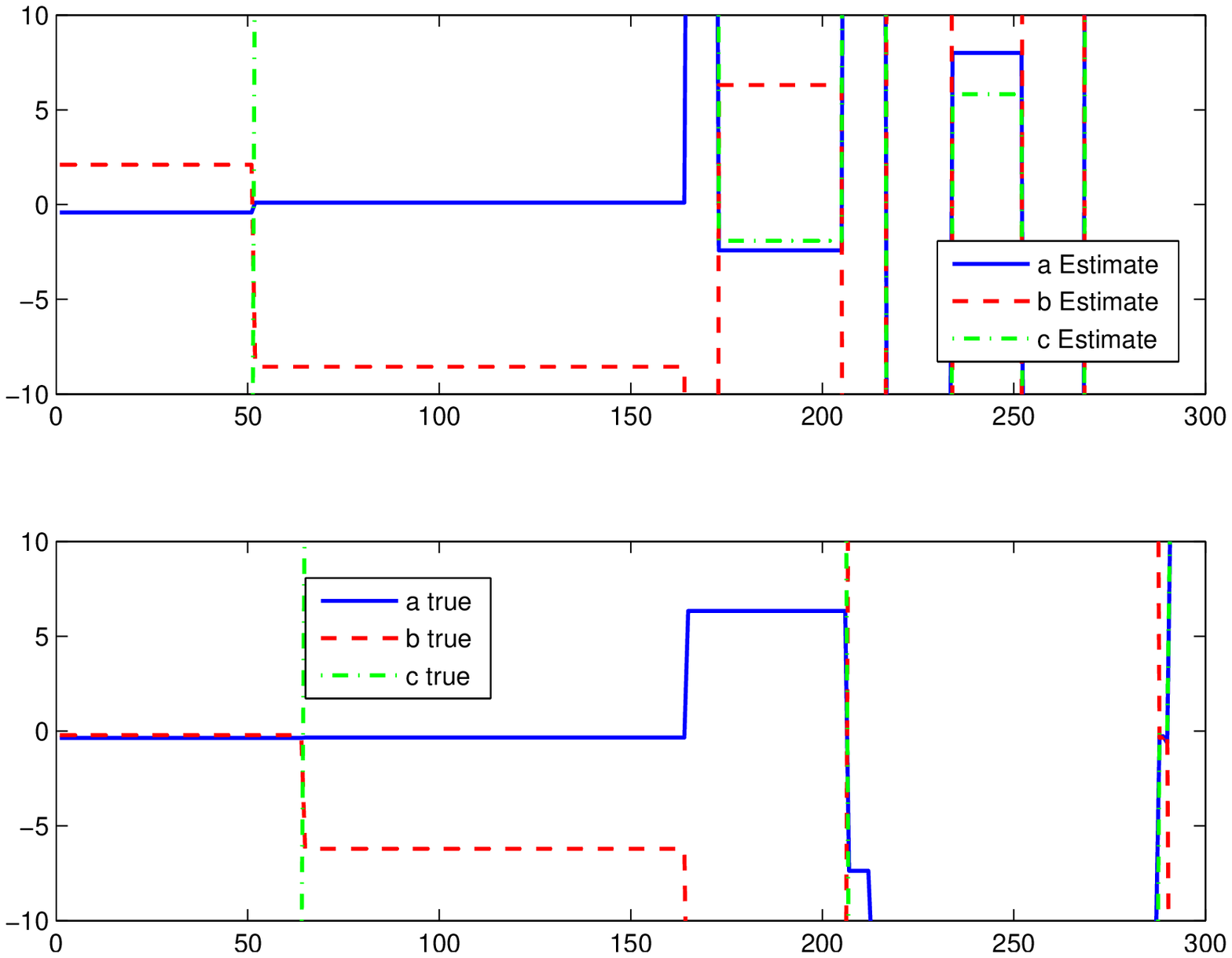}} 
\subfigure[Noisy Function $\sigma = 0.25$]{\includegraphics[width=0.33\linewidth]{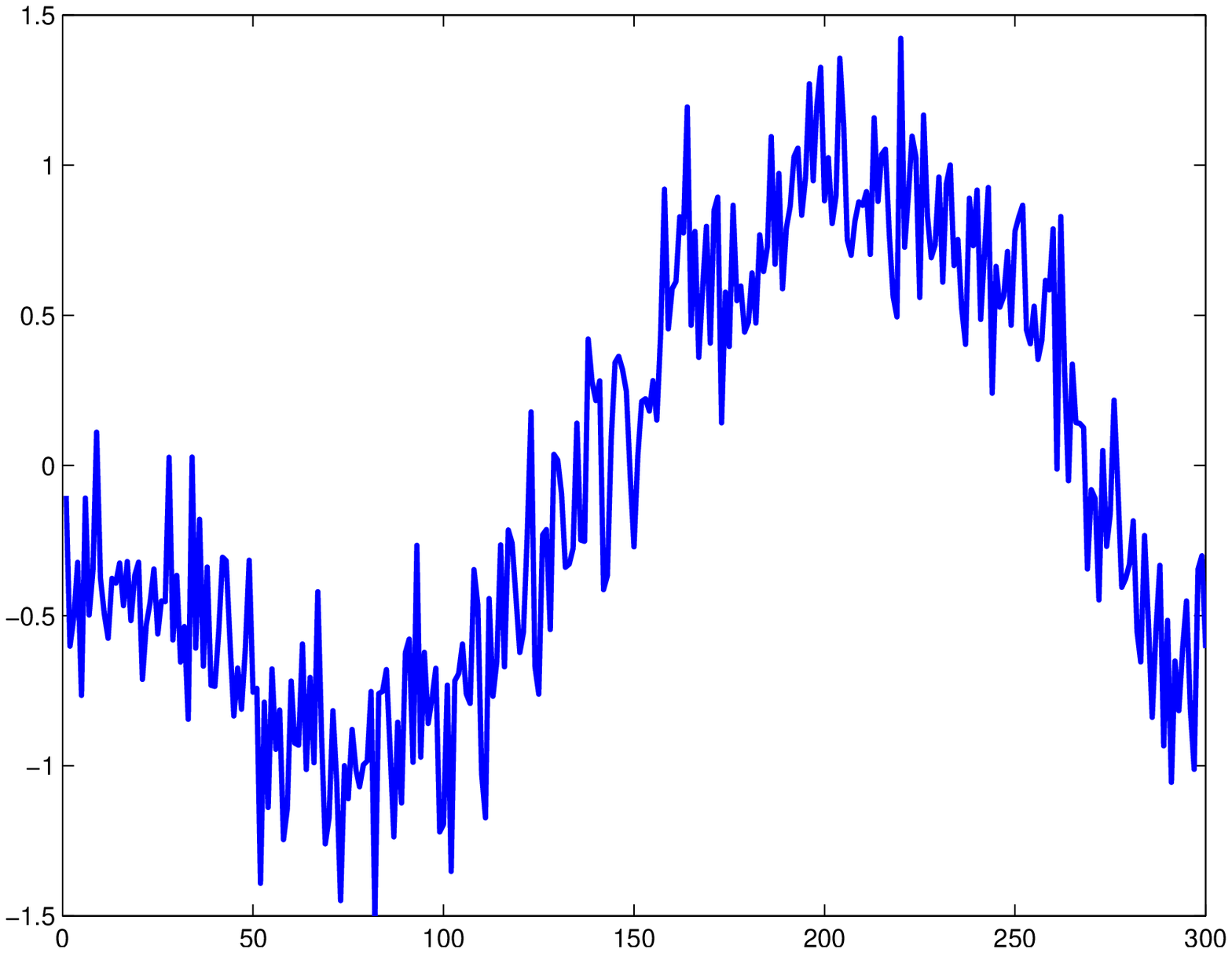}}%
\subfigure[Function Recovery for $\sigma = 0.25$]{\includegraphics[width=0.33\linewidth]{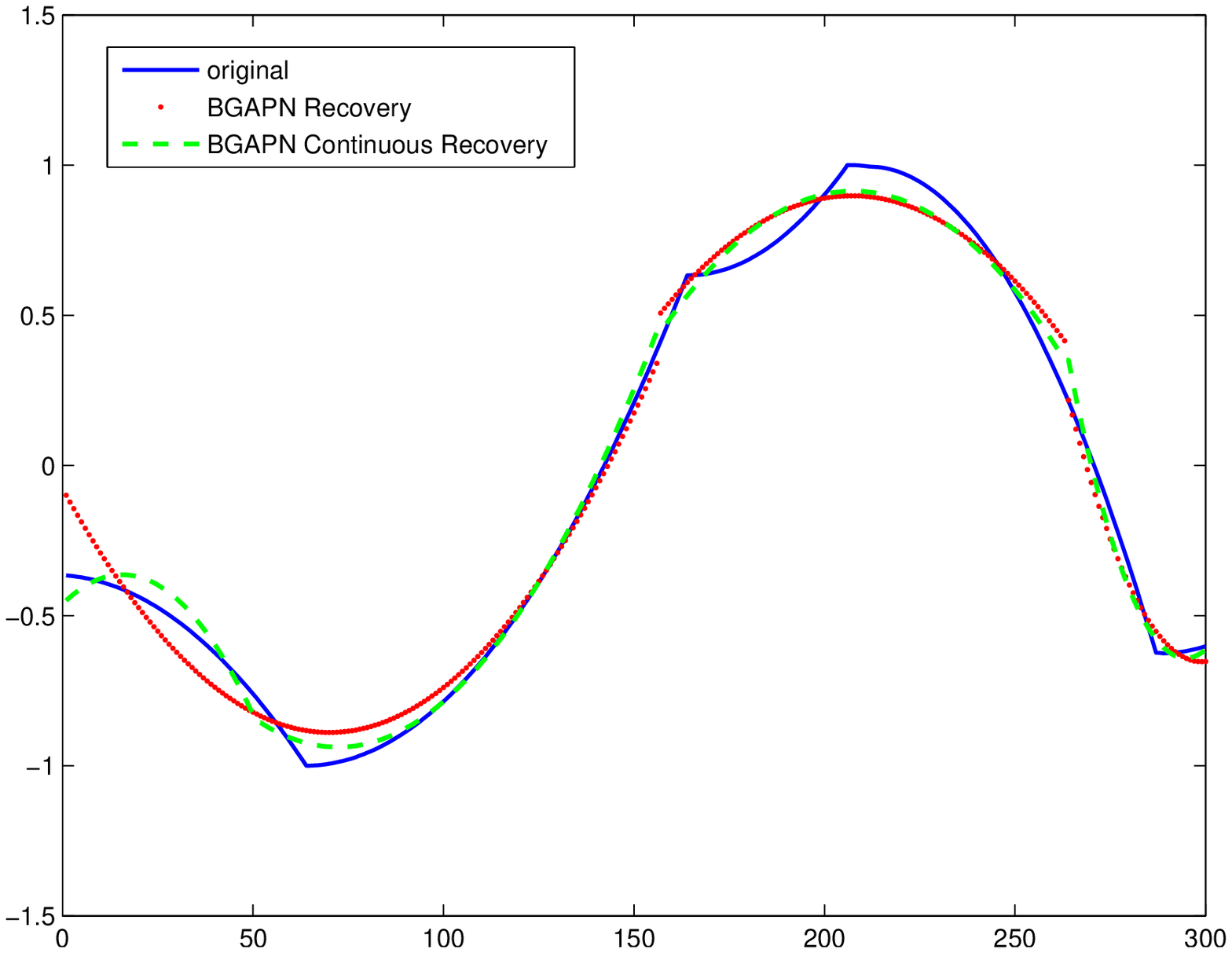}}
\subfigure[Coefficients Parameters Recovery for $\sigma = 0.25$]{\includegraphics[width=0.33\linewidth]{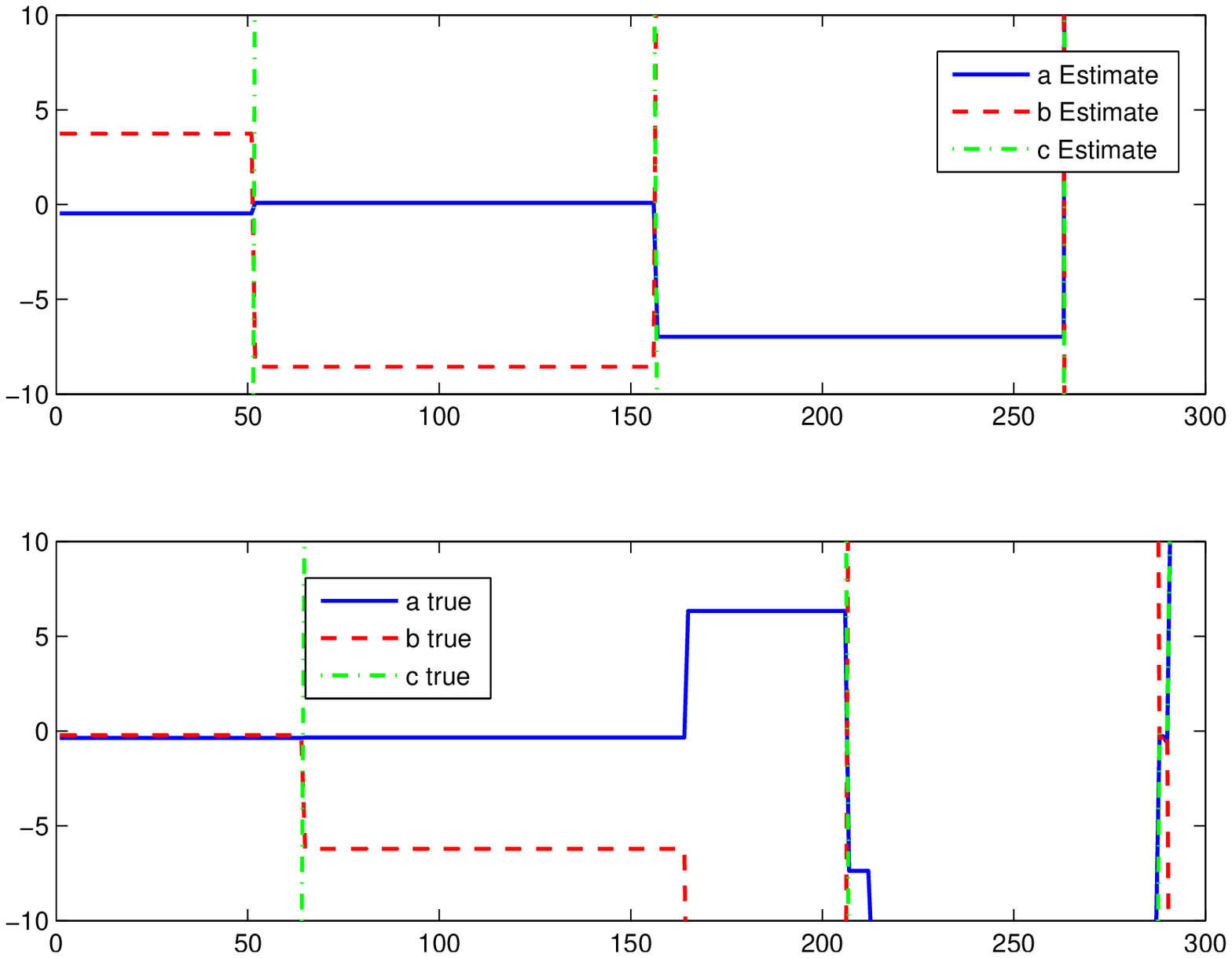}} 
}%
\caption{Recovery of a piecewise second-order polynomial function using the BGAPN algorithm with and without a constraint on the continuity.}
\label{fig:linear_function_recovery2}
\end{figure*}

 \subsection{Recovery Guarantees for Piecewise Polynomial Functions}

To provide theoretical guarantees for the recovery by SSCoSaMP we 
employ two theorems from \cite{Giryes14GreedySignal} and \cite{Giryes14NearOracle}. These lead to reconstruction error bounds for SSCoSaMP that guarantee stable recovery if the noise is adversarial, \rg{and an effective denoising} effect if it is zero-mean white Gaussian. 

Both theorems rely on the following property of the measurement matrix $\matr{M}$, which is a special case of the $\matr{D}$-RIP \cite{Candes11Compressed} and $\OM$-RIP \cite{Giryes14Greedy}.
\begin{defn}
\label{def:P_RIP}
A matrix $\matr{M}$ has a polynomial restricted isometry property of order $n$ ($P_n$-RIP) with a constant $\delta_k$ if for any piecewise polynomial function $\vect{f}$ of order $n$ with $k$ jumps  we have
\begin{eqnarray}
\label{eq:P_RIP}
(1-\delta_k) \norm{\vect{f}}_2^2 \le \norm{\matr{M}\vect{f}}_2^2 \le (1+\delta_k) \norm{\vect{f}}_2^2.
\end{eqnarray}
\end{defn}

Having the $P_n$-RIP definition we turn to present the first theorem, which treats the adversarial noise case.

\begin{thm}[Based on Corollary~3.2 in \cite{Giryes14GreedySignal}\footnote{Corollary~3.2  in \cite{Giryes14GreedySignal} provides stable recovery guarantees for general sparse vectors under a given dictionary $\matr{D}$ with the assumption that there exists a near-optimal projection algorithm that projects any vector to its closest sparse vector under the same dictionary. We can apply the result of Corollary~3.2 due to the optimal projection algorithm \rg{proposed} in Section~\ref{sec:opt_poly_func}.}]
\label{thm:SSCoSaMP_bound}
Let $\vect{f}$ be a piecewise polynomial function of order $n$, \rg{$\vect{e}$ be an adversarial bounded noise}  and 
$\matr{M}$ satisfy the $P_n$-RIP~\eqref{eq:P_RIP} with a constant
$\delta_{4k} < 0.046$. Then after a finite number of iterations, SSCoSaMP yields
\begin{eqnarray}
\label{eq:SSCoSaMP_bound}
&& \hspace{-0.5in} \norm{\hat{\vect{f}} -\vect{f}}_2 \le
 C\norm{\vect{e}}_2,
\end{eqnarray}
where $C > 2$ is a constant depending on $\delta_{4k}$.
\end{thm}

Note that the above theorem implies that we may compressively sense piecewise polynomial functions and achieve  
a perfect recovery in the noiseless case $\vect{e} = 0$. Note also that 
if $\matr{M}$ is a subgaussian random matrix
then it is sufficient to use only $m = O(k (n+\log(d))$ measurements  \cite{Blumensath09Sampling, Giryes14Greedy}. 

Though the above theorem is important for compressed sensing, it does not guarantee noise reduction, even for the case $\matr{M} = \matr{I}$, as $C>2$. The reason for this is that the noise here is adversarial, leading to a worst-case bound.  By introducing a random distribution for the noise, one may get better reconstruction guarantees.
The following theorem assumes that the noise is randomly Gaussian distributed, \rg{this way enabling to 
  provide  effective}  denoising guarantees. 

\begin{thm}[Based on Theorem~1.7 in \cite{Giryes14NearOracle}\footnote{Theorem~1.7  in \cite{Giryes14GreedySignal} provides near-oracle performance guarantees for block-sparse vectors under a given dictionary $\matr{D}$ with the assumption that there exists a near-optimal projection algorithm that projects any vector to its closest sparse vector under the same dictionary. We can apply the result of Theorem~1.7 due to the optimal projection algorithm \rg{proposed} in Section~\ref{sec:opt_poly_func}.}]
\label{thm:SSCoSaMP_bound_near_oracle}
Assume the conditions of Theorem~\ref{thm:SSCoSaMP_bound} such that $\vect{e}$ is a random zero-mean white Gaussian noise  with a variance $\sigma^2$.  Then after a finite number of iterations, SSCoSaMP yields 
\begin{eqnarray}
&&  \hspace{-0.3in} \norm{\hat{\vect{f}} -\vect{f}}_2 \le \\ \nonumber && ~~~~~~~
  C\sqrt{(1+\delta_{3 k})3 k}\left(1+\sqrt{2(1+\beta)\log(nd)} \right)\sigma,
\end{eqnarray}
with probability exceeding $ 1- \frac{2}{(3 k)!}(nd)^{-\beta}$.
\end{thm}

The bound in the theorem can be given on the expected error instead of being given only with high probability using the proof technique in \cite{Giryes12RIP}. 
We remark that if we were given an oracle that foreknows the locations of the jumps in the parameterization, the error we would get would be $O(\sqrt{k}\sigma)$. As the $\log(nd)$ factor in our bound is inevitable \cite{Candes06Modern}, we may conclude that our guarantee is optimal up to a constant factor.

\begin{figure*}[htb]
\centering
{\subfigure{\includegraphics[width=0.45\linewidth]{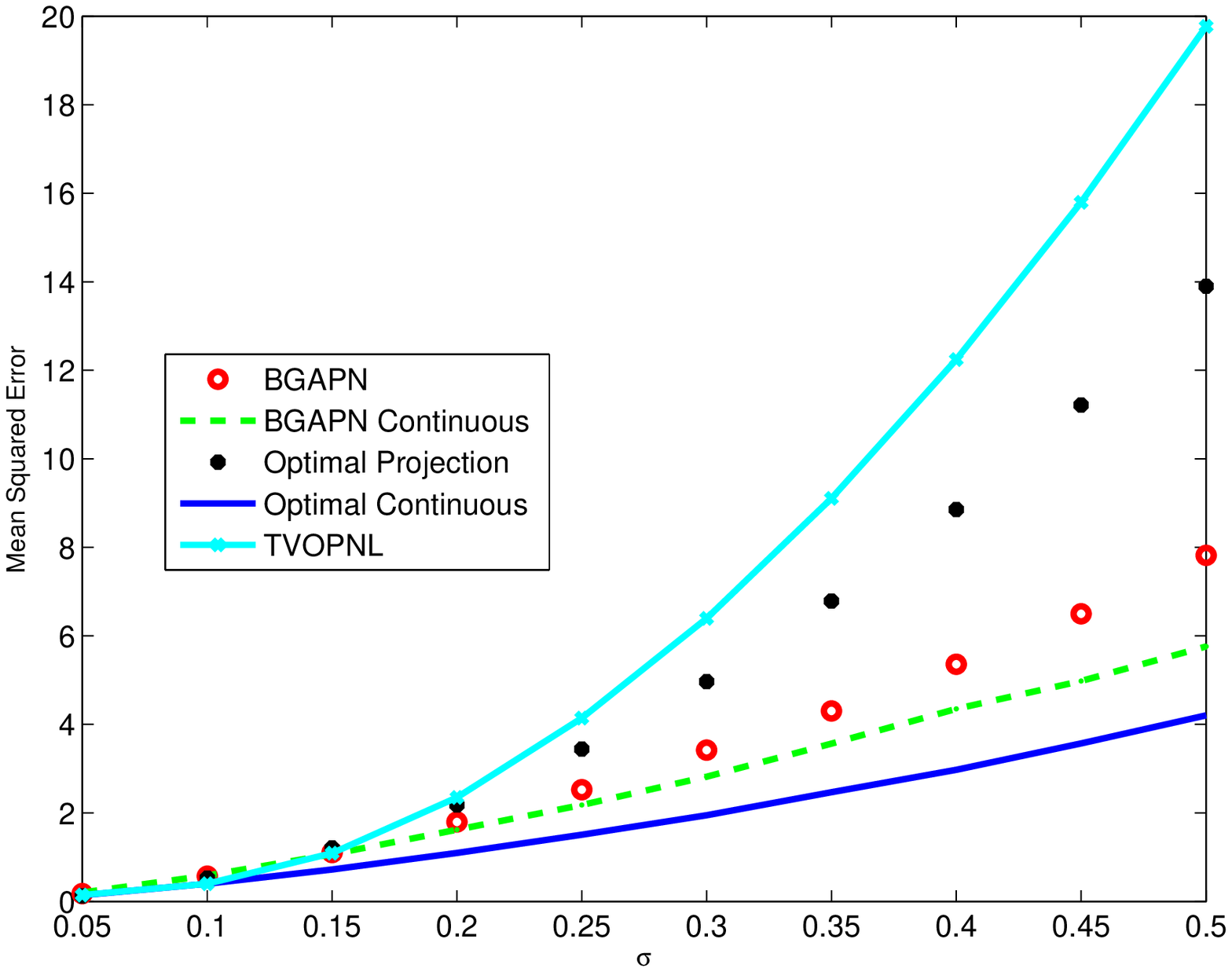}} \hfill
\subfigure{\includegraphics[width=0.45\linewidth]{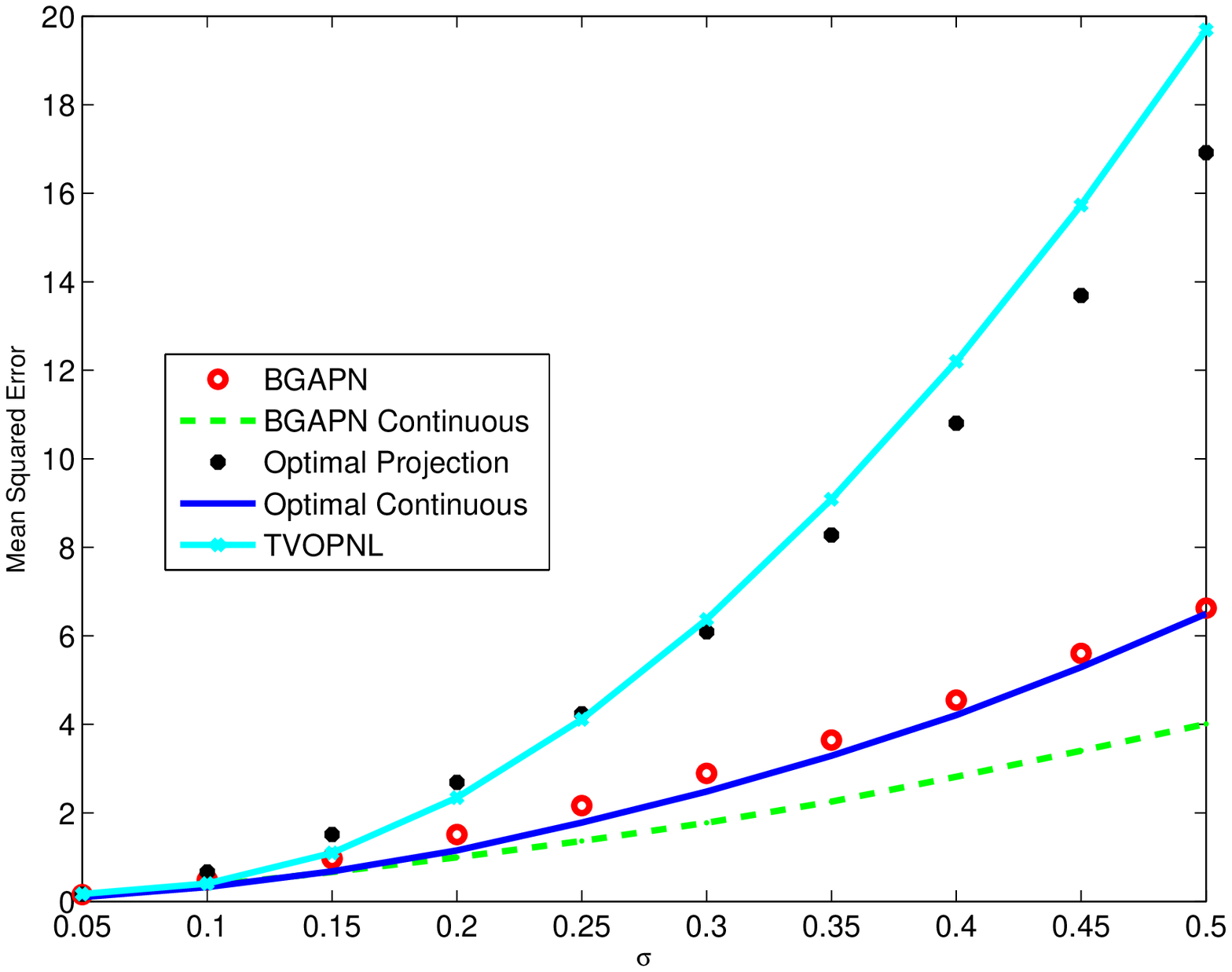}}
} %
\caption{MSE of the recovered piecewise linear  functions (left) and piecewise second-order polynomial functions (right) as a function of the noise variance $\sigma$ for the methods BGAPN with and without  the continuity constraint and the optimal approximation with and without continuity post-processing. As a reference we compare to the non local overparameterized TV (TVOPNL) approach introduced in \cite{shem13globally}.}
\label{fig:function_recovery_sigma}
\end{figure*}
  
\begin{figure*}[htb]
\centering
{\subfigure{\includegraphics[width=0.45\linewidth]{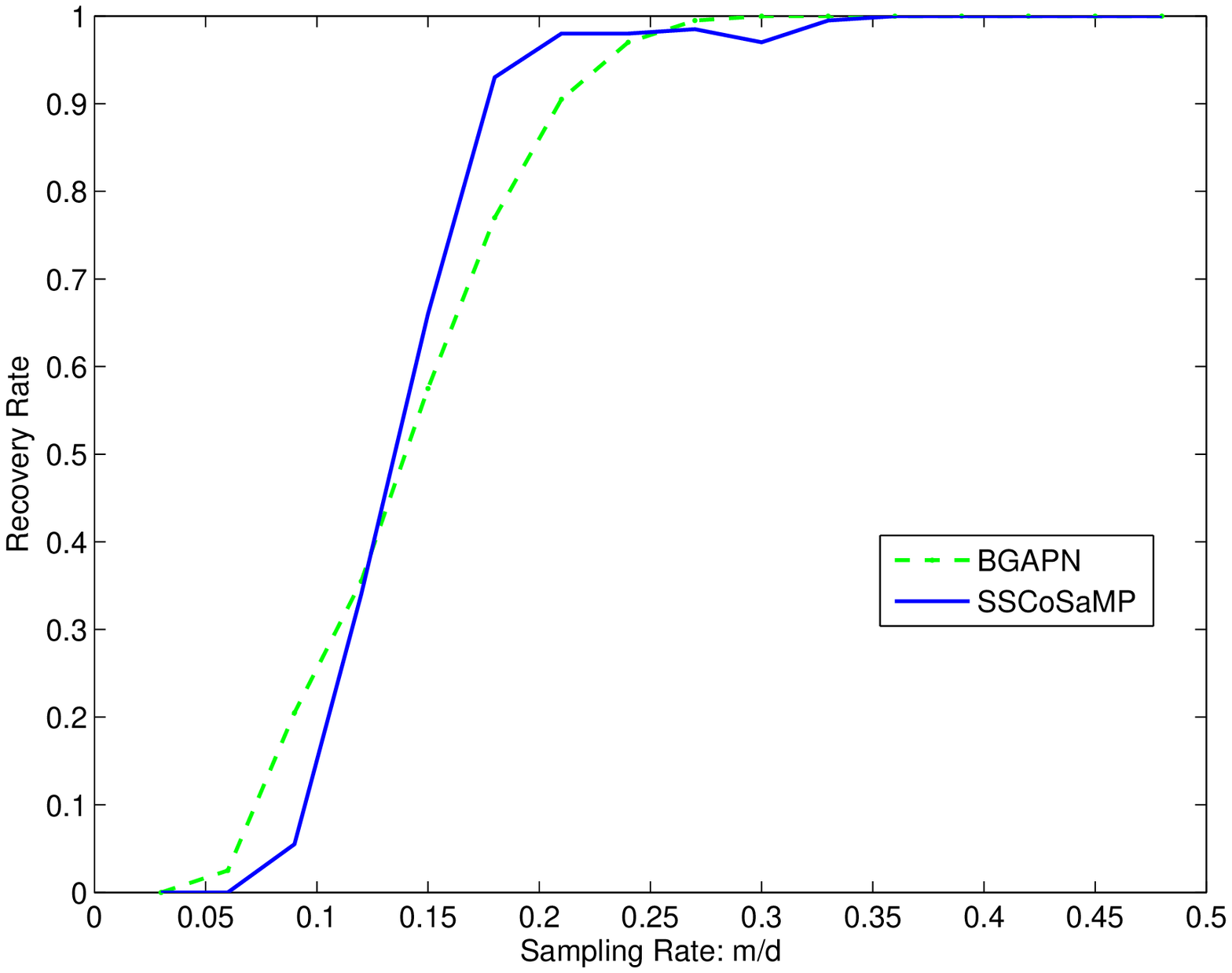}}} %
\caption{Recovery rate of piecewise second-order polynomial functions as a function of the sampling rate $m/d$ for the methods BGAPN and SSCoSaMP.}
\label{fig:function_recovery_CS}
\end{figure*}
   
\section{Sparsity based Overparameterized Variational Algorithm for High Dimensional Functions}
\label{sec:overparameterized}

We now turn to generalize the model in \eqref{eq:P0_analysis_k_linear_overparam} to support other overparameterization forms, including higher dimensional functions such as images. We consider the case where an upper-bound for the noise energy is given and not the sparsity $k$, as is common in many applications. 
Notice that for the synthesis model, such a generalization is not trivial because while it is easy to extend the $\DIF$ operator to high dimensions, 
it is not clear how to do this for the Heaviside dictionary.

Therefore we consider an overparameterized version of \eqref{eq:P0_analysis_e}, where the noise energy is known and the analysis model is used.
Let $\vect{X}_1, \dots \vect{X}_n$ be matrices of the space variables and $\vect{b}_1 \dots \vect{b}_n$ their coefficients parameters. 
\rg{For example, in a 2D (image) case of piecewise linear constant, $\matr{X}_1$ will be the identity matrix, $\matr{X}_2$ will be a diagonal matrix with the values $[1, 2, \dots, d, 1, 2, \dots, d, \dots 1, 2, \dots, d]$ on its main diagonal, and $\matr{X}_3$ will similarly be a diagonal matrix with $[1, 1, \dots, 1, 2, 2, \dots, 2, \dots d, d, \dots, d]$ on its main diagonal.} 
\rg{Assuming that all the coefficient parameters are jointly sparse under a general operator $\OM$, we may recover these coefficients by solving}
\begin{eqnarray}
\label{eq:P0_analysis_e_overparam}
&& \hspace{-0.3in} \left[\hat{\vect{b}}_1^T, \dots,  \hat{\vect{b}}_n^T \right]^T = \min_{\tilde{\vect{b}}_1, \dots, \tilde{\vect{b}}_n} \norm{\sum_{i=1}^n\abs{\OM\tilde{\vect{b}}_i}}_0 \\ \nonumber && \hspace{0.3in} s.t.  ~~ \norm{\vect{g} - \matr{M}\left[\matr{X}_1, \dots \matr{X}_n\right] \left[ \begin{array}{c}
\tilde{\vect{b}}_1 \\ \vdots \\ \tilde{\vect{b}}_n
\end{array}\right]}_2 \le \norm{\vect{e}}_2.
\end{eqnarray}
Having an estimate for all these coefficients, our approximation for the original signal $\vect{f}$ is $\hat{\vect{f}} = \left[\matr{X}_1, \dots, \matr{X}_n \right]\left[\hat{\vect{b}}_1^T, \dots,  \hat{\vect{b}}_n^T \right]^T$.

As the minimization problem in \eqref{eq:P0_analysis_e_overparam} is NP-hard we suggest to solve it by a generalization of the GAPN algorithm \cite{Nam11GAPN} --  the block GAPN (BGAPN). We introduce this extension  in Appendix~\ref{sec:bgapn}.

This algorithm aims at finding in a greedy way the rows of $\OM$ that are orthogonal to the space variables $\vect{b}_1 \dots \vect{b}_n$. Notice that once we find the \rg{indices of these rows}, the set $\Lambda$ that satisfies $\OM_{\Lambda}{\vect{b}}_i =0$ for $i= 1\dots n$ ($\OM_{\Lambda}$ is the submatrix of $\OM$ with the rows corresponding to the set  $\Lambda$), we may approximate $\vect{b}_1 \dots \vect{b}_n$ by solving 
\begin{eqnarray}
\label{eq:GAPN_e_overparam}
&& \hspace{-0.3in} \left[\hat{\vect{b}}_1^T, \dots,  \hat{\vect{b}}_n^T \right]^T = \min_{\tilde{\vect{b}}_1, \dots, \tilde{\vect{b}}_n} {\sum_{i=1}^n\norm{\OM_{\Lambda}\tilde{\vect{b}}_i}_2^2} \\ \nonumber && \hspace{0.3in} s.t.  ~~ \norm{\vect{g} - \matr{M}\left[\matr{X}_1, \dots \matr{X}_n\right] \left[ \begin{array}{c}
\tilde{\vect{b}}_1 \\ \vdots \\ \tilde{\vect{b}}_n
\end{array}\right]}_2 \le \norm{\vect{e}}_2.
\end{eqnarray}
Therefore, BGAPN \rg{approximates $\vect{b}_1 \dots \vect{b}_n$} by finding $\Lambda$ \rg{first}. It starts with $\Lambda$ that \rg{includes} all the rows \rg{of} $\OM$ and \rg{then} gradually \rg{removes} elements from it by solving \rg{the problem posed in \eqref{eq:GAPN_e_overparam}} at each iteration and then finding the row in $\OM$ \rg{that has the largest correlation} with the current temporal solution  $ \left[\hat{\vect{b}}_1^T, \dots,  \hat{\vect{b}}_n^T \right]^T $.

Note that there are no known recovery guarantees for BGAPN of the form  we have had for SSCoSaMP before. Therefore, we present its efficiency in several experiments in the next section. 
As explained in Appendix~\ref{sec:bgapn}, the advantages of BGAPN over SSCoSaMP, despite the lack of theoretical guarantees, are that (i) it does not need $k$ to be foreknown and (ii) it is easier to use with higher dimensional functions.

Before we move to the next section we note that one of the advantages of the above formulation and the BGAPN algorithm is the relative ease of adding to it new constraints. For example, we may encounter piecewise polynomial functions that are also continuous. However, we do not have such a continuity constraint in the current formulation. As we shall see in the next section, the absence of such a constraint allows jumps in the \rg{discontinuity} points between the polynomial segments and therefore it is important to add it to the algorithm to get a better reconstruction.

One possibility to solve this problem is to add a continuity constraint on the jump points of the signal. In Appendix~\ref{sec:bgapn} we present also a modified version of the BGAPN algorithm that imposes such a continuity constraint, \rg{and in the next section we shall see how this handles} the problem. Note that this is only one example of a constraint that one may add to the BGAPN technique. For example, in images one may add a \rg{smoothness constraint}  on the edges' directions.

\section{Experiments}
\label{sec:exp}

For demonstrating the efficiency of the proposed method we perform several tests. We start with the one dimensional case, testing our polynomial fitting approach with the continuity constraint and without it for continuous \rg{piecewise polynomials  of first and second degrees.} We \rg{compare these results with} the optimal polynomial approximation scheme presented in Section~\ref{sec:guarantees} \rg{and to the variational approach in \cite{shem13globally}}. We continue with a compressed sensing experiment for discontinuous piecewise polynomials and compare BGAPN with SSCoSaMP.
Then we perform some tests on images using BGAPN. We start by denoising cartoon images using the piecewise linear model. We compare our outcome with the one of TV denoising \cite{Rudin92Nonlinear} and show that \rg{our result does} not suffer from a staircasing effect \cite{Savage07Multigrids}. \rg{We compare also to a TV denoising version with overparameterization \cite{Nir07Over}.}
 Then we show how our framework may be used for image segmentation, drawing a connection to the Mumford-Shah functional \cite{Mumford89Optimal, Ambrosio90Approximation}. 
 \rg{We compare our results with the ones obtained by the popular graph-cuts based segmentation \cite{Felzenszwalb04Efficient}.}

\subsection{Continuous Piecewise Polynomial Functions Denoising}

In order to check the performance of the polynomial fitting, we generate random continuous piecewise-linear and second-order polynomial functions with $300$ samples, $6$ jumps and a dynamic range $[-1,1]$. Then we contaminate the signal with a white Gaussian noise with a standard deviation from the set $\left\{0.05, 0.1, 0.15, \dots, 0.5 \right\}$. 

We compare the recovery result of BGAPN with and without the continuity   constraint with the one of the optimal approximation\footnote{We have done the same experiment with the BOMP algorithm \cite{Eldar10Block}, adopting the synthesis framework, with and without the continuity constraint, \rg{and observed that it performs very similarly to}  BGAPN.}. 
Figs.~\ref{fig:linear_function_recovery1} and \ref{fig:linear_function_recovery2} present BGAPN reconstruction results for the linear and second order polynomial cases respectively, for two different noise levels. 
It can be observed that the addition of the continuity constraint is essential for the correctness of the recovery. Indeed, without it we get jumps between the segments. Note also that the number of jumps in our recovery may be different than the one of the original signal as BGAPN does not have a preliminary information about it. However, it still manages to recover the parameterization in a good way,  especially in the lower noise case. 

The possibility to provide a parametric representation is one of the advantages of our method. Indeed, one may achieve good denoising results without using the linear model in terms of mean squared error (MSE) using methods such as free-knot spline \cite{Jupp78Siam}. However, the approximated function is not guaranteed to be piecewise linear and therefore learning the change points from it is sub-optimal. See \cite{shem13globally} and the references therein for more details. 

To evaluate our method with respect to its MSE we compare it with the optimal approximation for piecewise polynomial function presented in Appendix~\ref{sec:opt_poly_func}. Note that the target signals are continuous while this algorithm does not use this assumption. Therefore, we add the continuity constraint to this method as a post processing (unlike BGAPN that \rg{merges this} in its steps). We take the changing points it has recovered and project the noisy measurement $\vect{g}$ to its closest \rg{continuous piecewise polynomial function with the same discontinuities}.

Figure~\ref{fig:function_recovery_sigma} presents the recovery performance of BGAPN and the projection algorithm with and without the continuous constraint. Without the constraint, it can be observed that  BGAPN achieves better recovery performance. This is due to the fact that it is not restricted to the number of change points in the initial signal and therefore it can use more points and thus adapt itself better to the signal, achieving lower MSE. However, after adding the constraint in the piecewise linear case the optimal projection achieves a better recovery error. The reason is that, as the optimal projection uses the exact number of points, it finds the changing locations more accurately. Note though that in the case of second order polynomial functions, BGAPN gets better recovery. This happens because  this program uses the continuity constraint also within its iterations and not only at the final step, as is the case with the projection algorithm. As the second order polynomial case is more  complex than the piecewise linear one, the impact of the usage of the continuity prior is higher and more significant than the information on the number of change points.

We compare also to the non-local opverapameterized TV algorithm (TVOPNL) in \cite{shem13globally}\footnote{Code provided by the authors.}, \rg{which was shown to be better for the task of line segmentation, when compared with several alternatives including the ones  reported in \cite{Nir07Over} and \cite{Nir08Over}.} Clearly, our proposed scheme achieves better recovery performance than TVOPNL, demonstrating the supremacy of our line segmentation strategy.

\begin{figure}[htb]
\centering
{\subfigure[Original Image]{\includegraphics[width=0.48\linewidth]{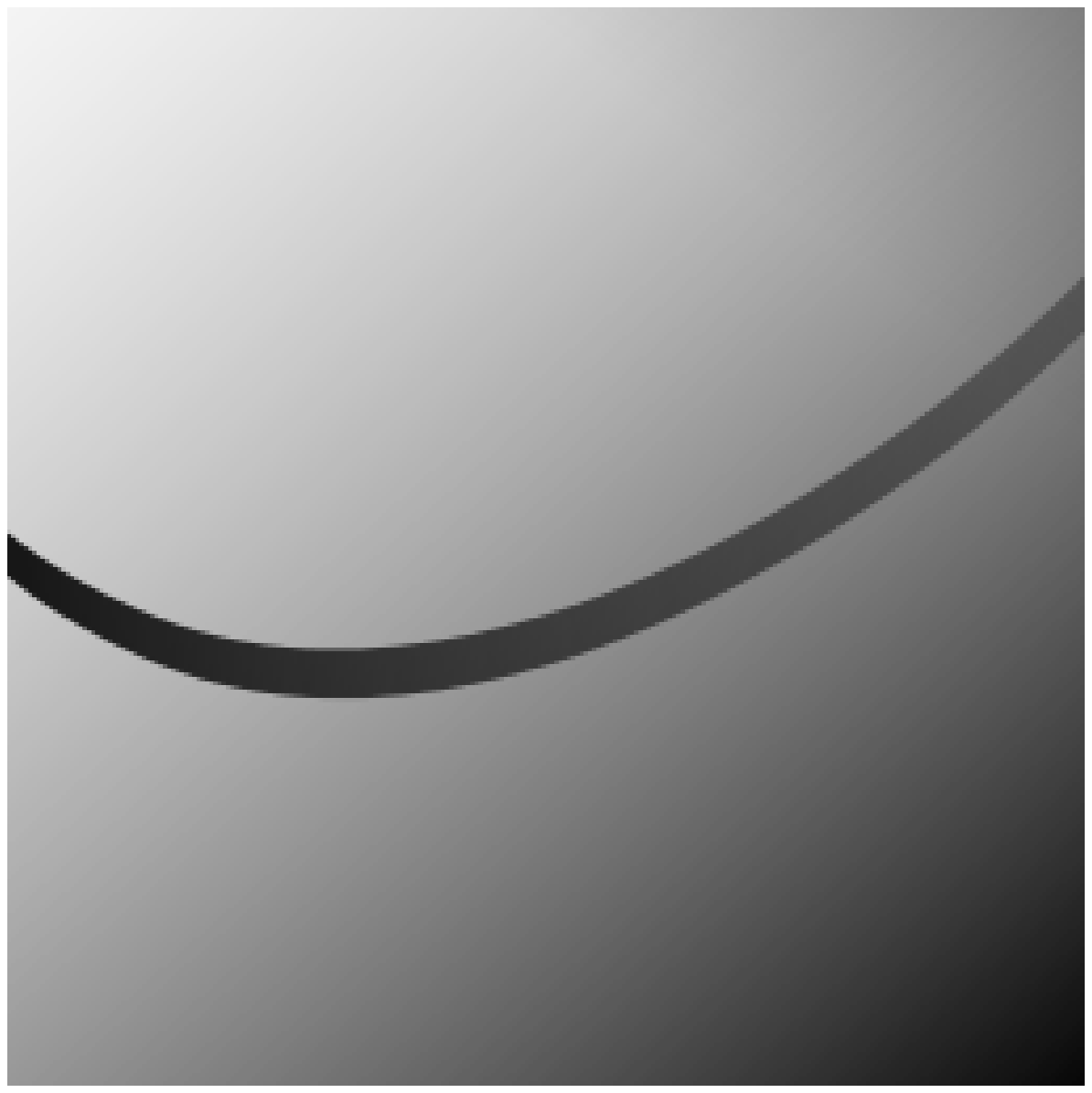}}%
\subfigure[Noisy Image $\sigma = 20$]{\includegraphics[width=0.48\linewidth]{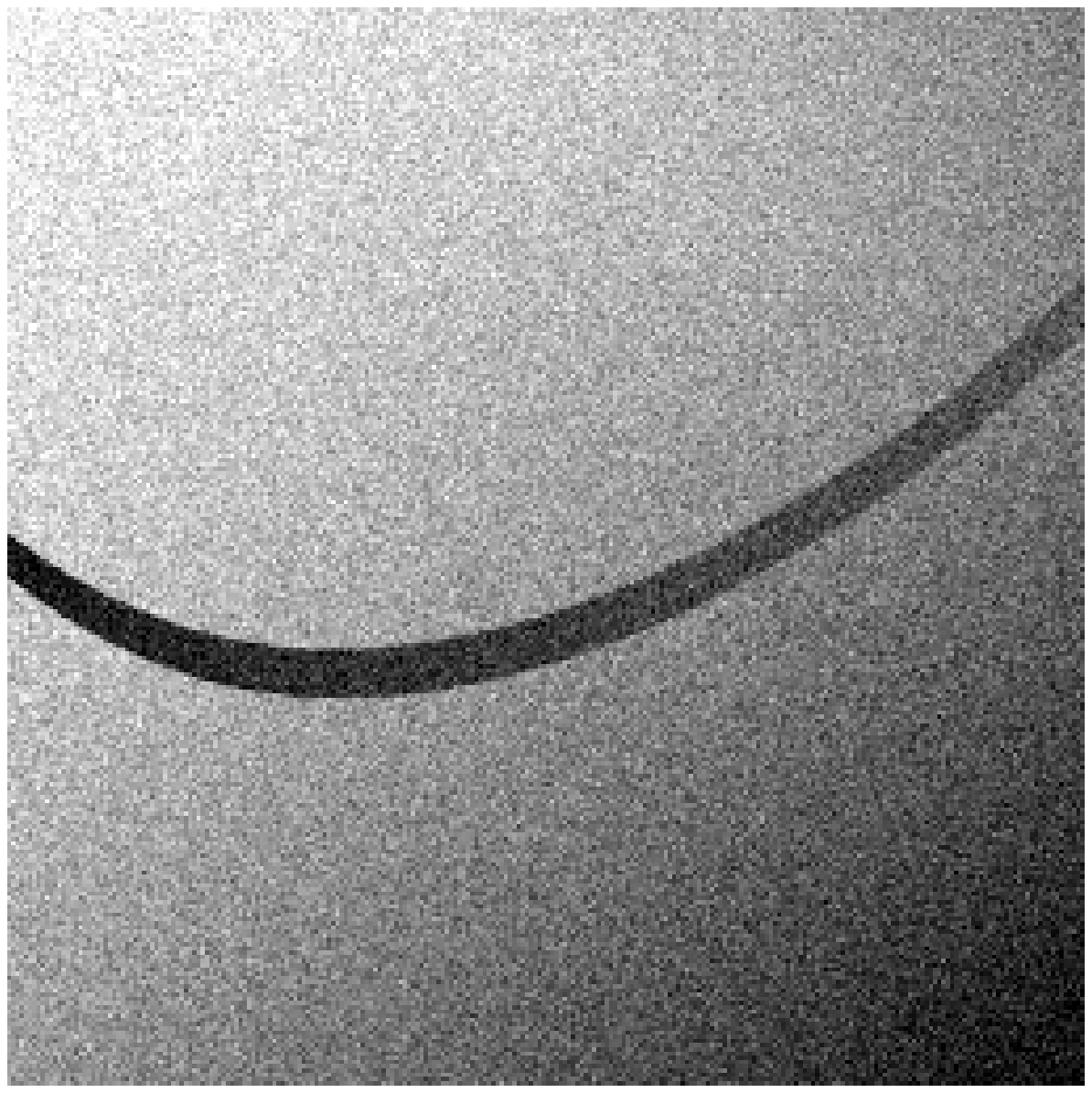}}
\subfigure[BGAPN with $\DIF$. PSNR =  40.09dB.]{\includegraphics[width=0.48\linewidth]{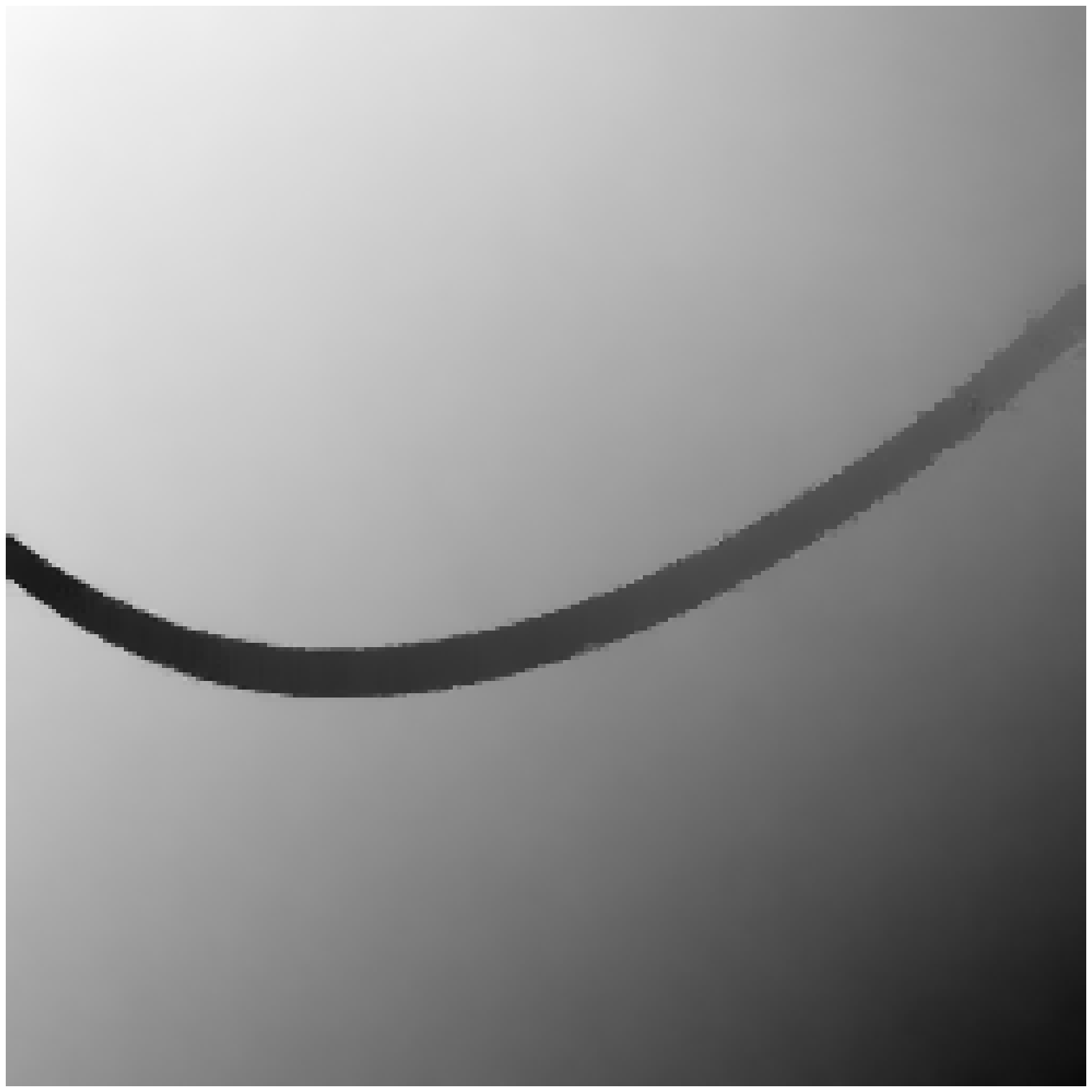}} 
\subfigure[TV recovery. PSNR = 38.95dB.]{\includegraphics[width=0.48\linewidth]{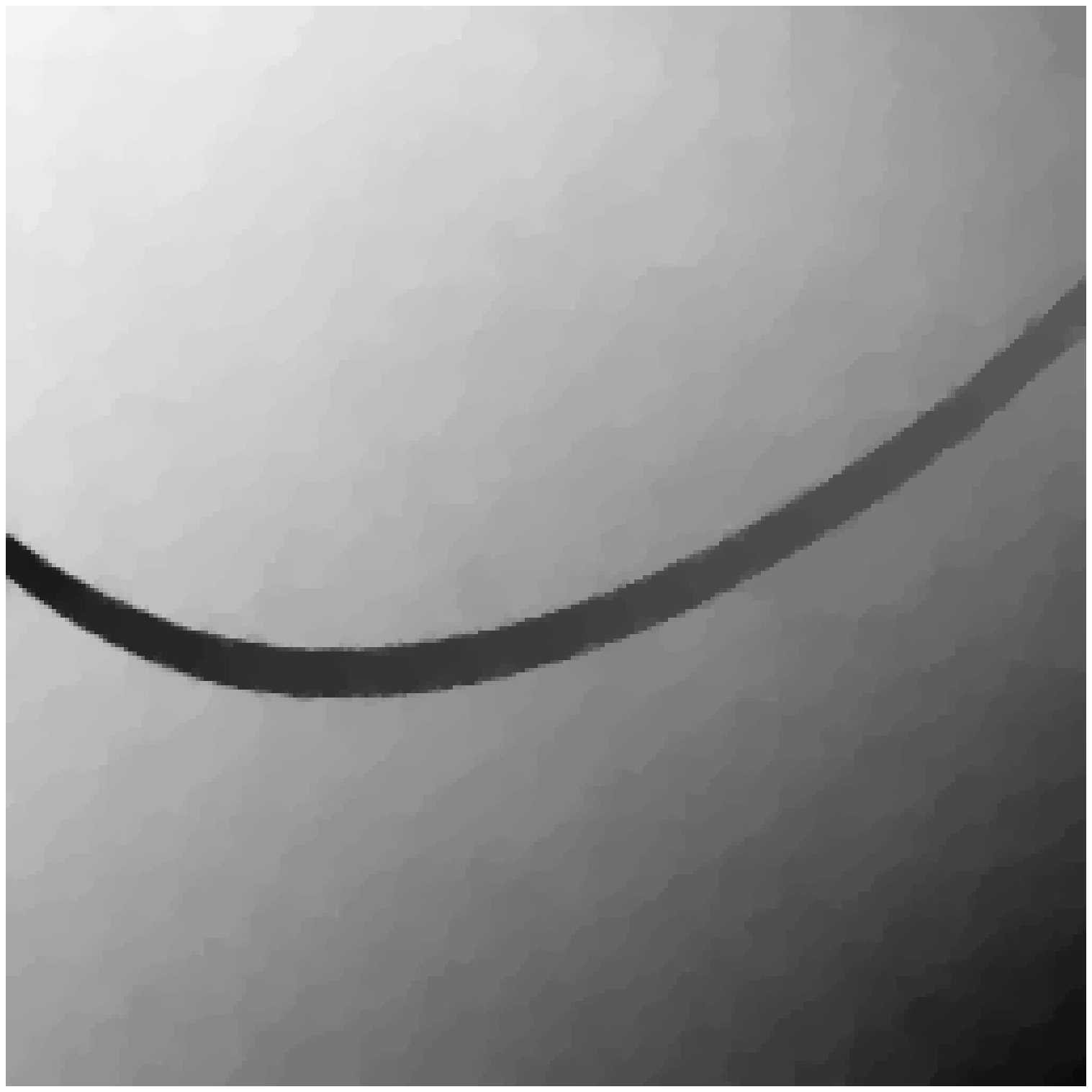}}%
\subfigure[TV OP recovery. PSNR = 37.41dB.]{\includegraphics[width=0.48\linewidth]{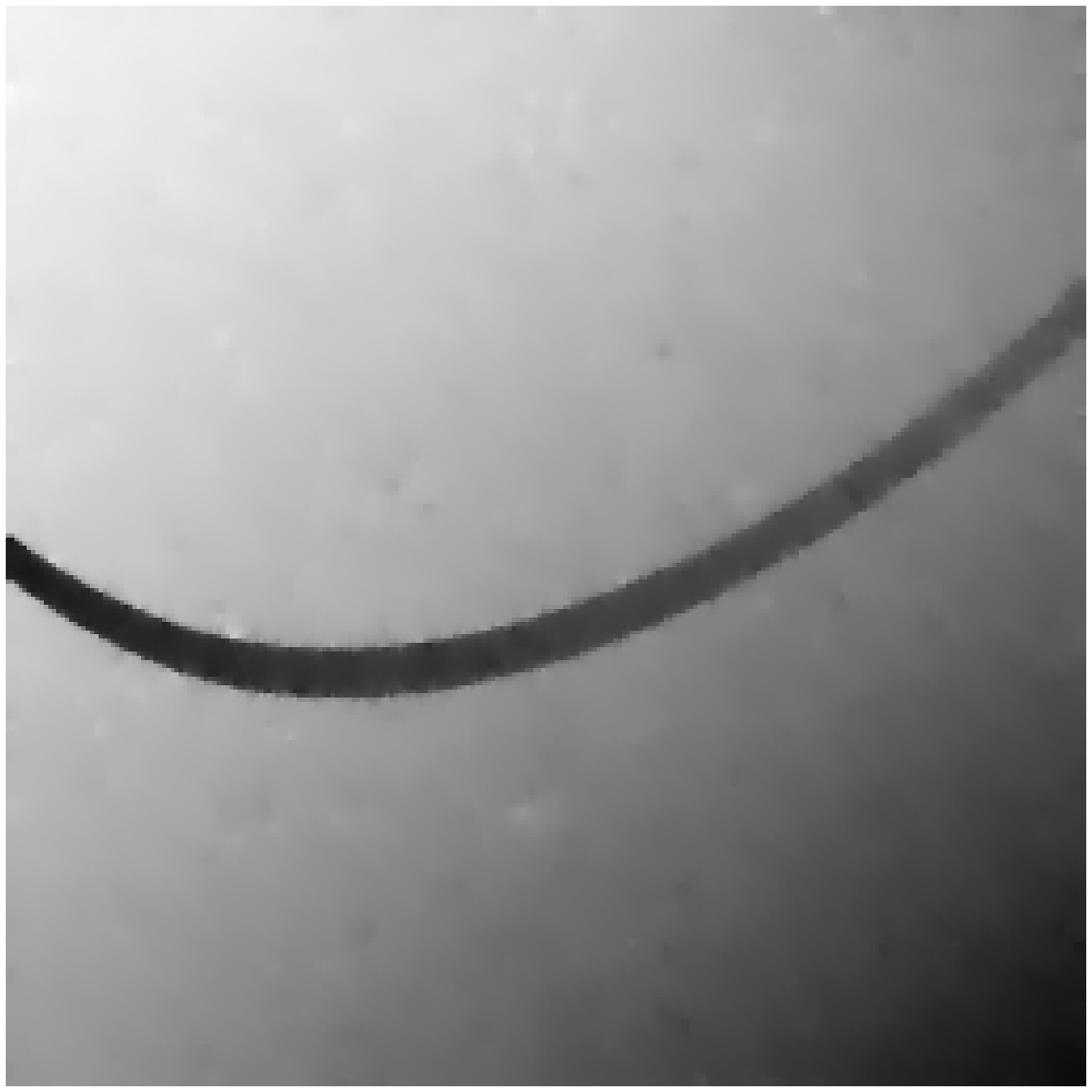}}%
}%
\caption{Denoising of {\em swoosh} using the BGAPN algorithm with and without diagonal derivatives. Notice that we do not have the staircasing effect that appears in the TV reconstruction.}
\label{fig:swoosh_recovery}
\end{figure}

\begin{figure}[htb]
\centering
{\subfigure[Original Image]{\includegraphics[width=0.48\linewidth]{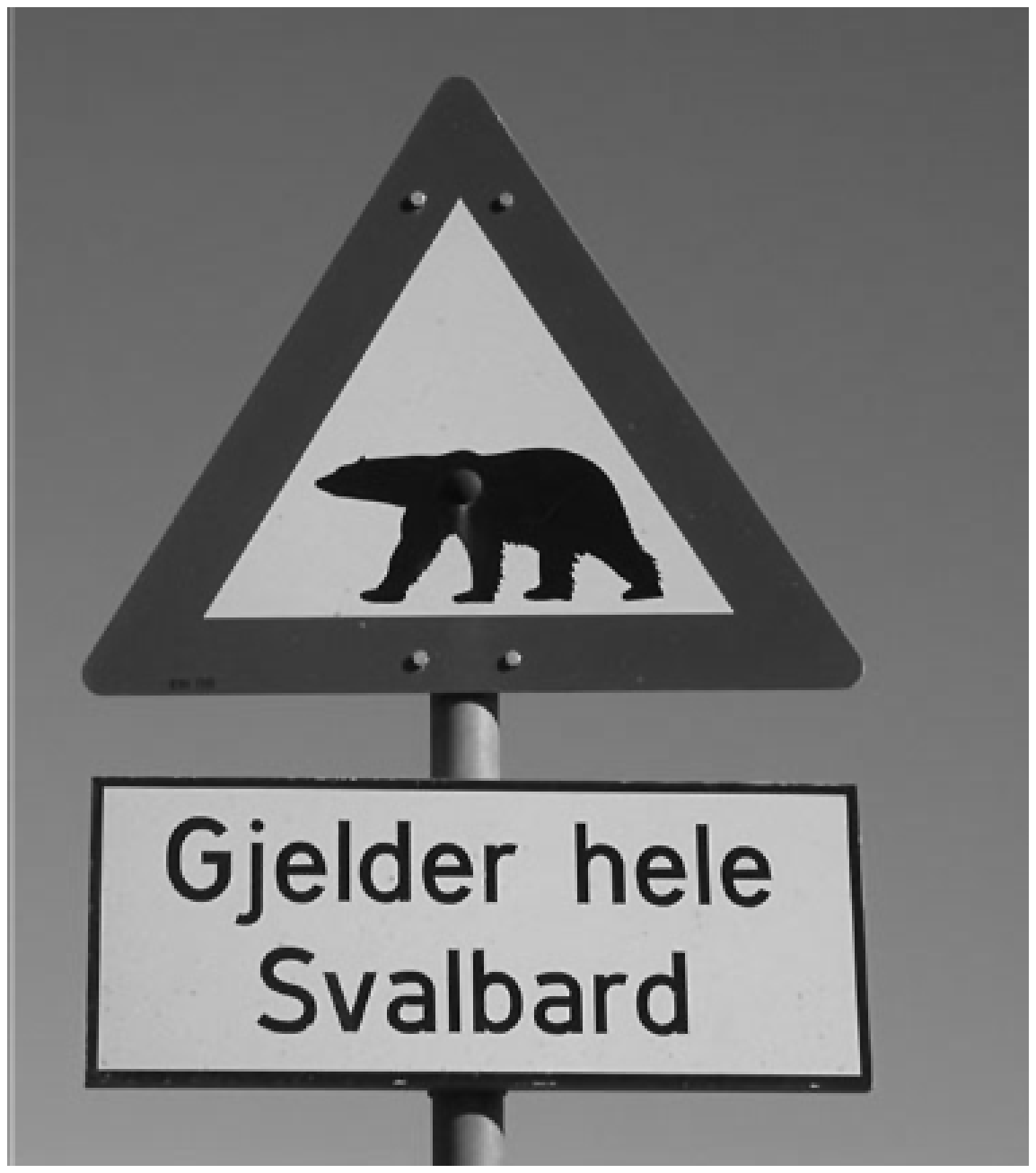}}%
\subfigure[Noisy Image $\sigma = 20$]{\includegraphics[width=0.48\linewidth]{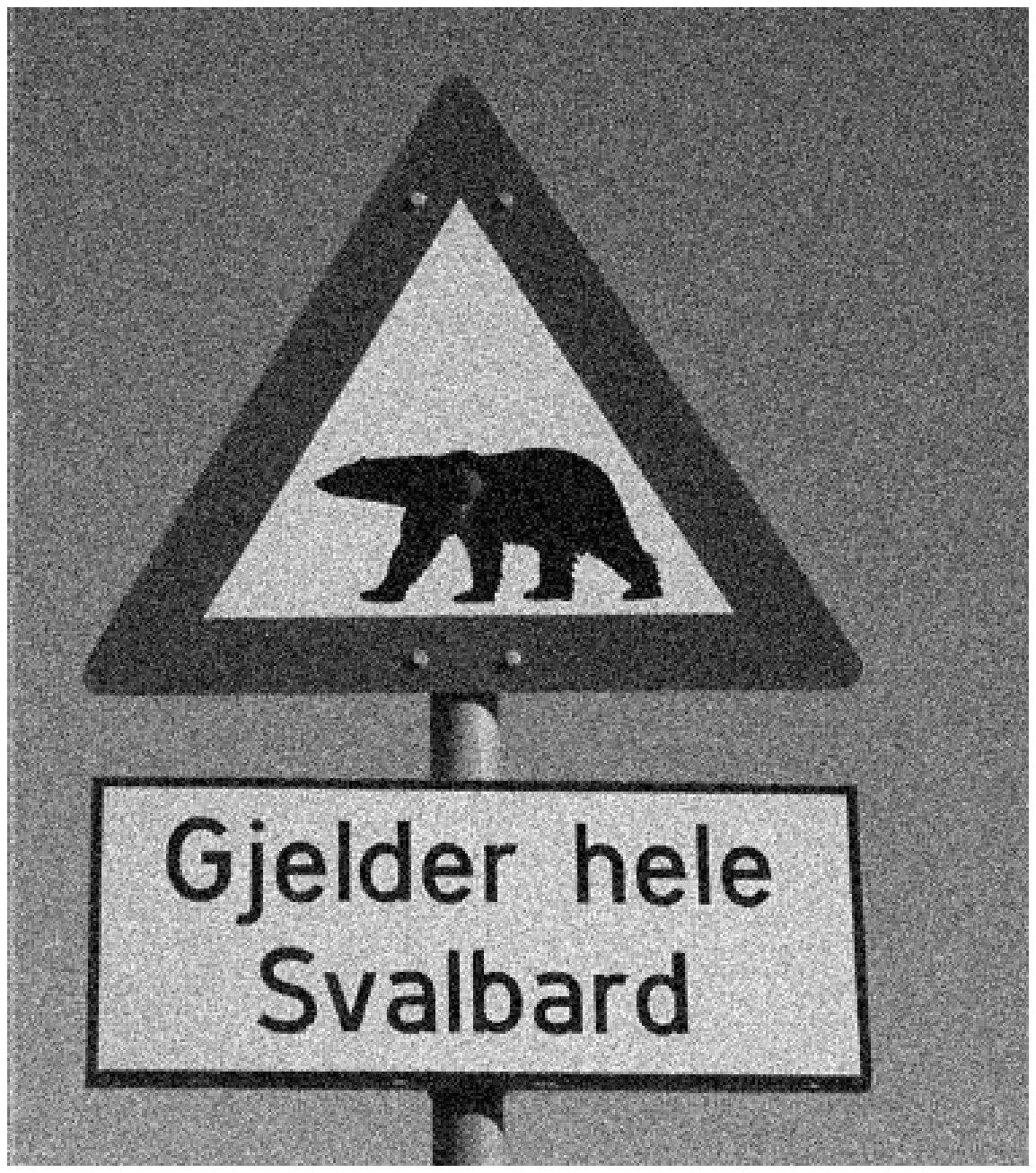}}
\subfigure[BGAPN with $\DIF$. PSNR =  34.02dB.]{\includegraphics[width=0.48\linewidth]{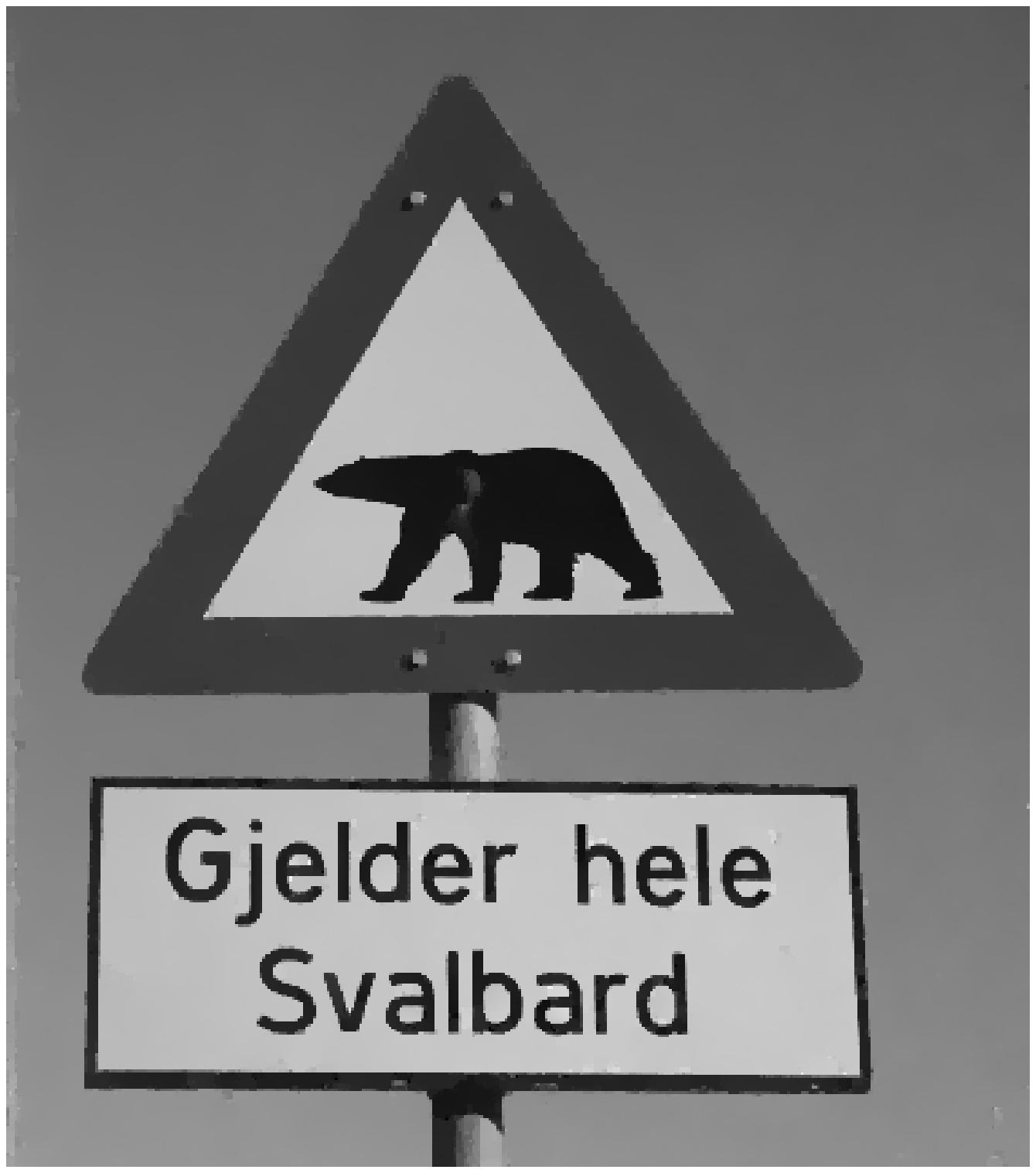}} 
\subfigure[TV recovery. PSNR =33.69dB.]{\includegraphics[width=0.48\linewidth]{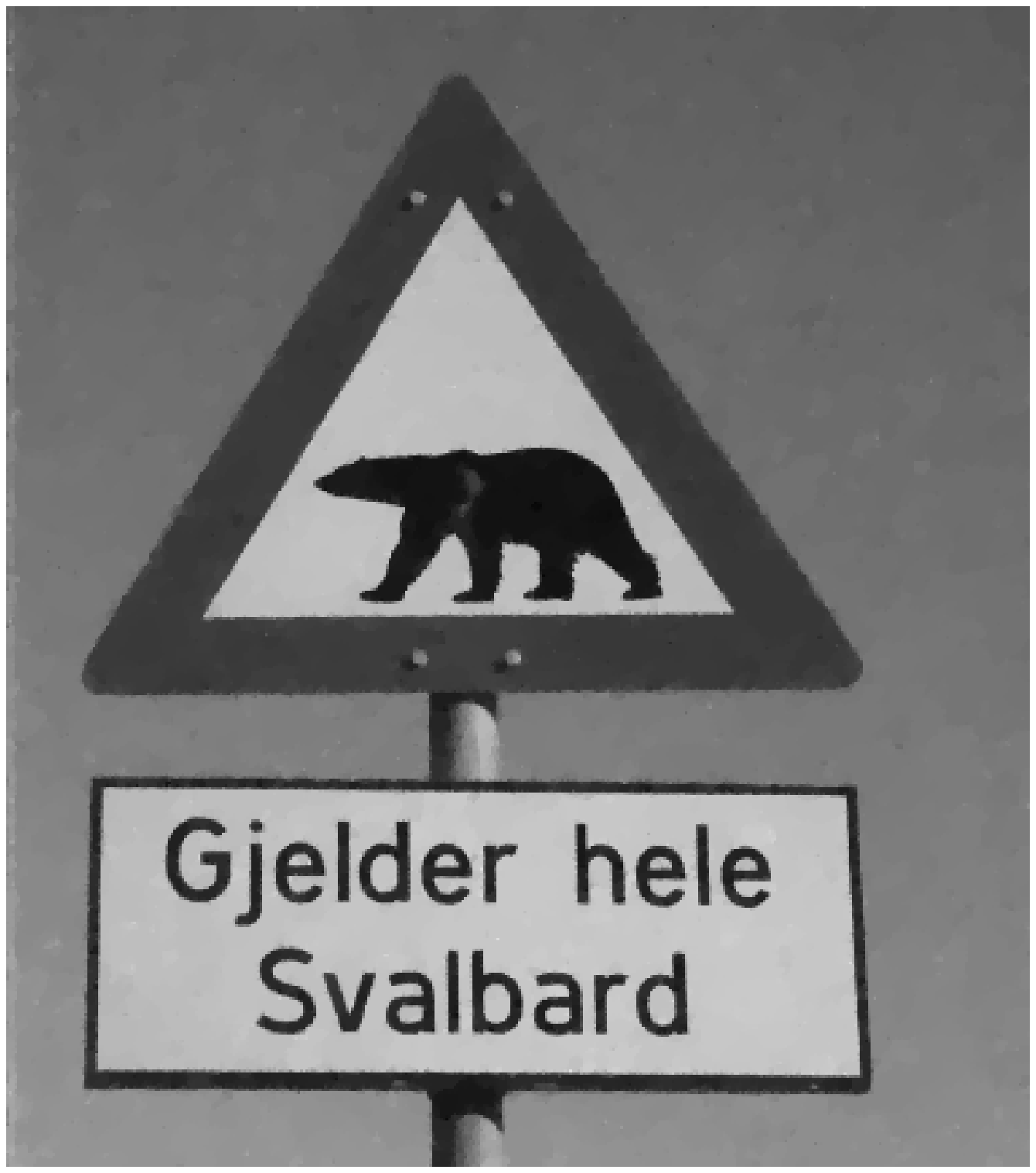}}%
\subfigure[TV OP recovery. PSNR =31.83dB.]{\includegraphics[width=0.48\linewidth]{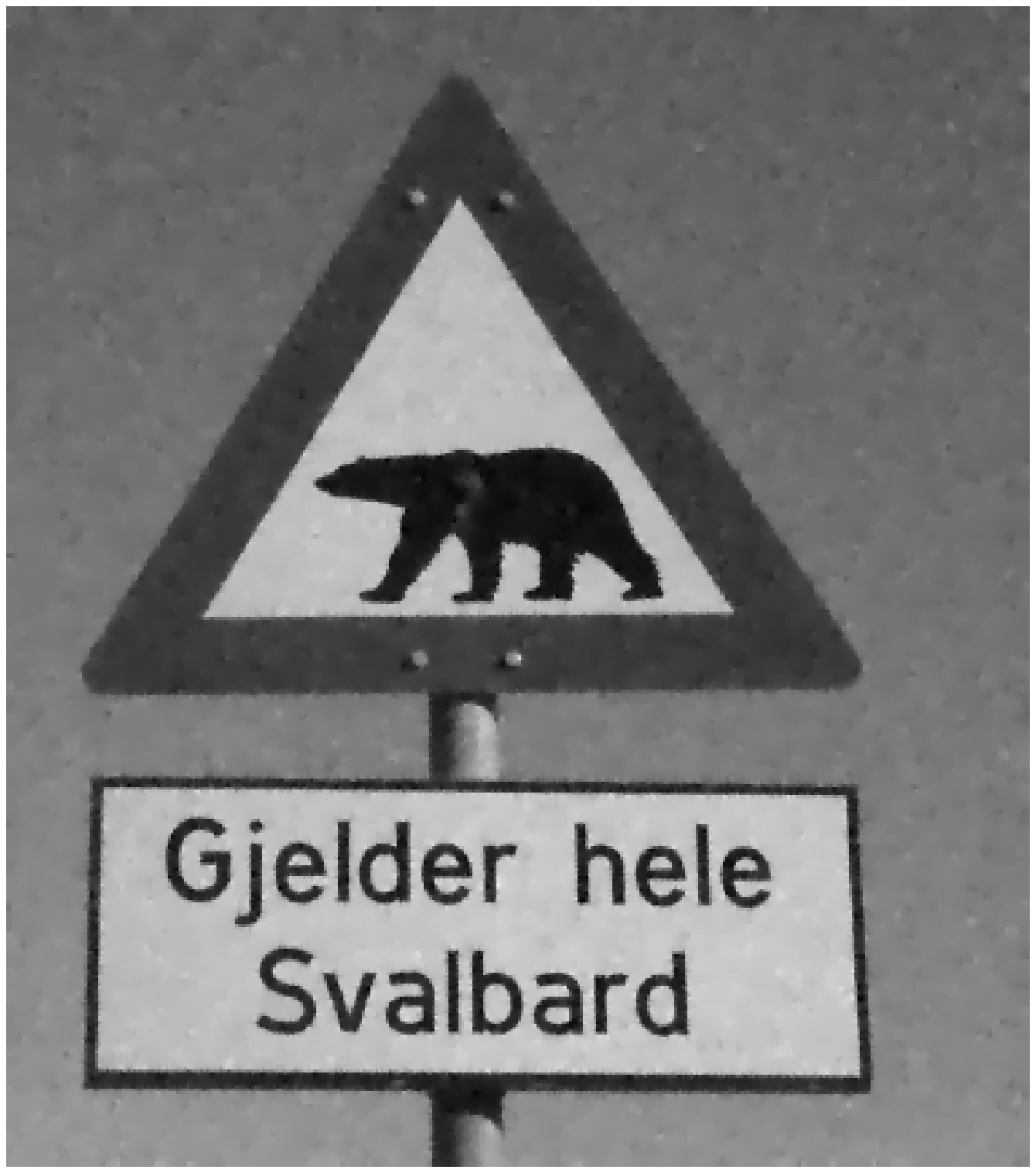}}%
}%
\caption{Denoising of {\em sign} using the BGAPN algorithm. The \rg{results of TV and OP-TV are} presented as a reference.}
\label{fig:sign_recovery}
\end{figure}

\begin{figure}[htb]
\centering
{\subfigure[Original Image]{\includegraphics[width=0.48\linewidth]{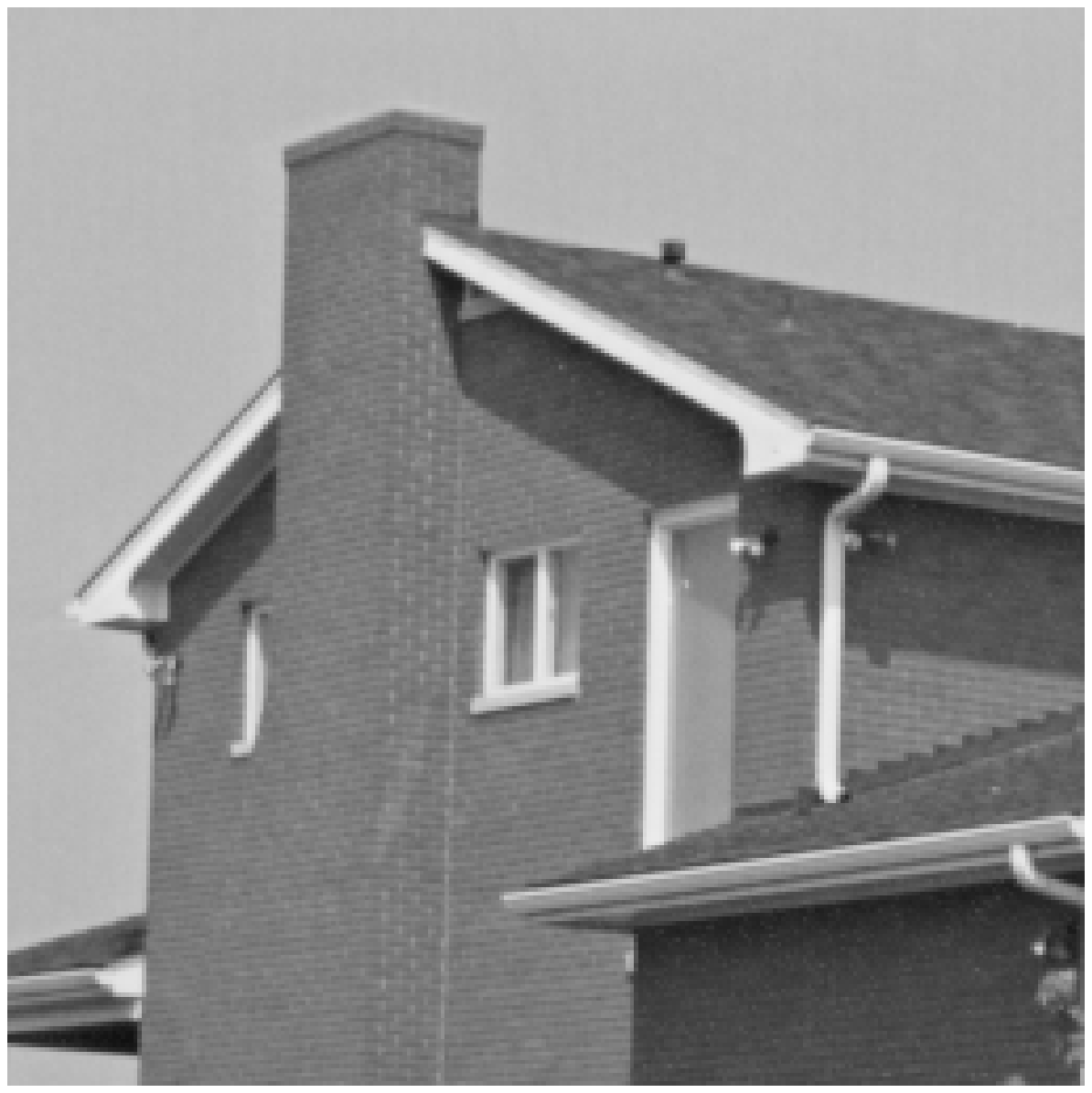}}%
\subfigure[Noisy Image $\sigma = 20$]{\includegraphics[width=0.48\linewidth]{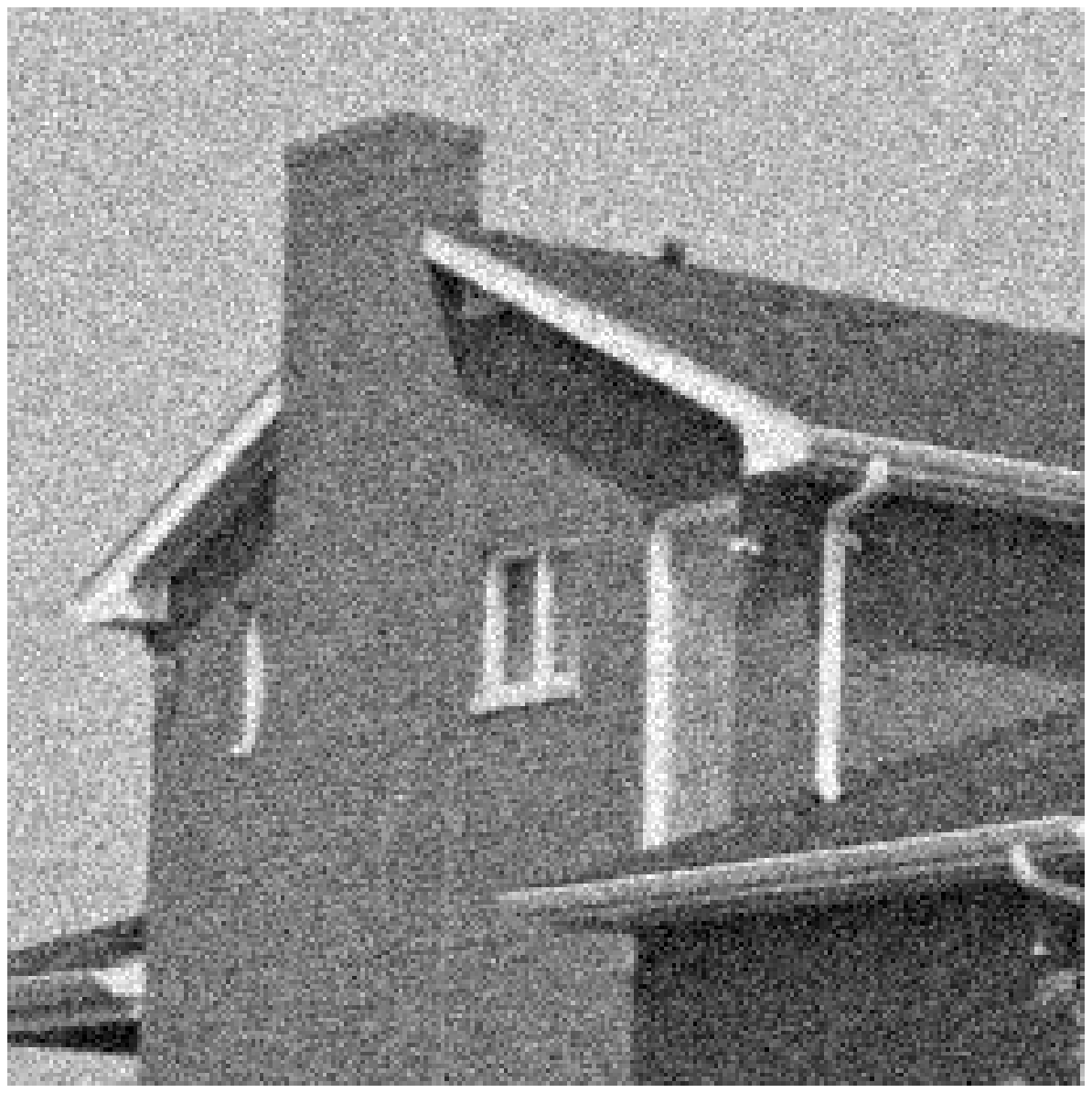}}
\subfigure[BGAPN with $\DIF$. PSNR =  30.77dB.]{\includegraphics[width=0.48\linewidth]{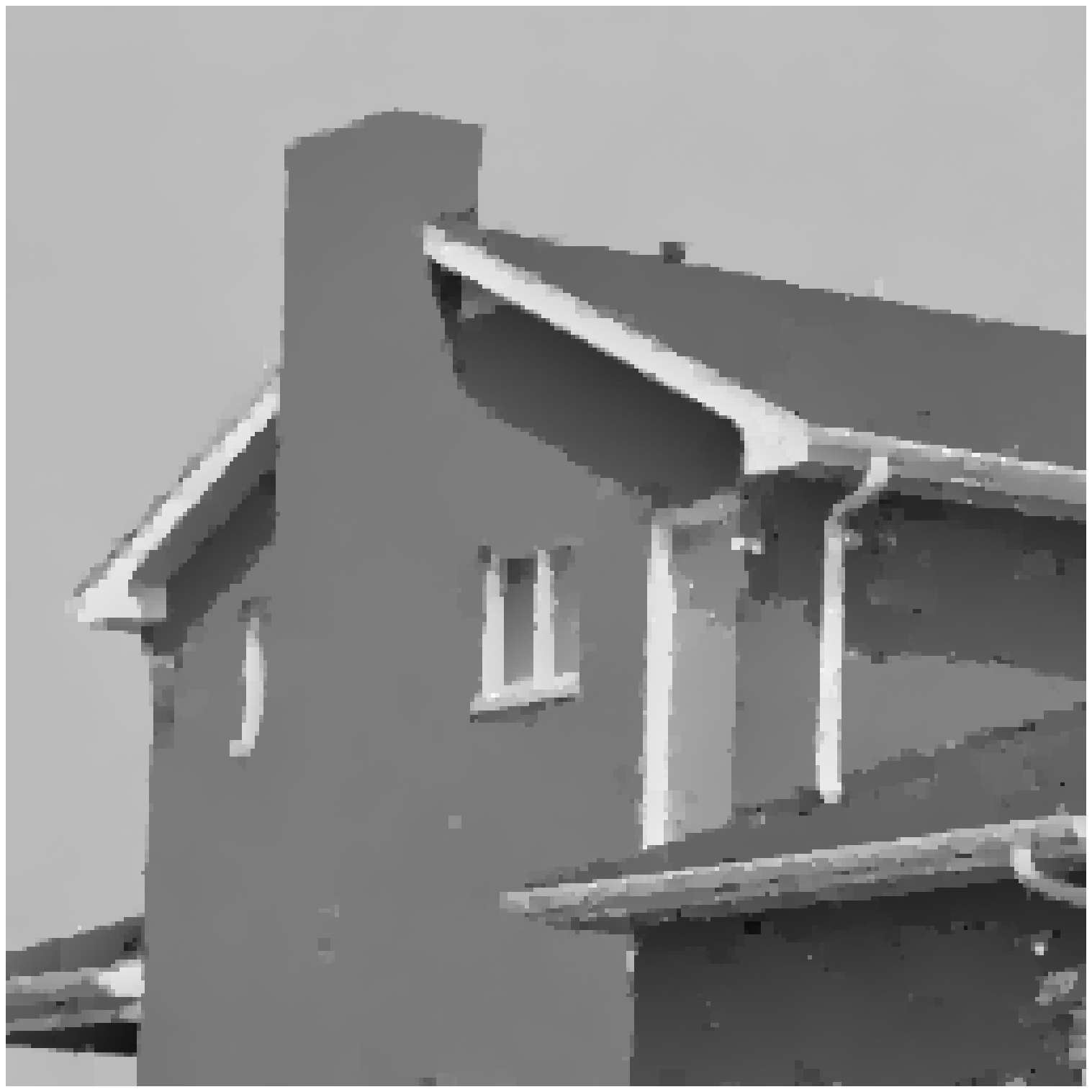}} 
\subfigure[TV recovery. PSNR = 31.44dB.]{\includegraphics[width=0.48\linewidth]{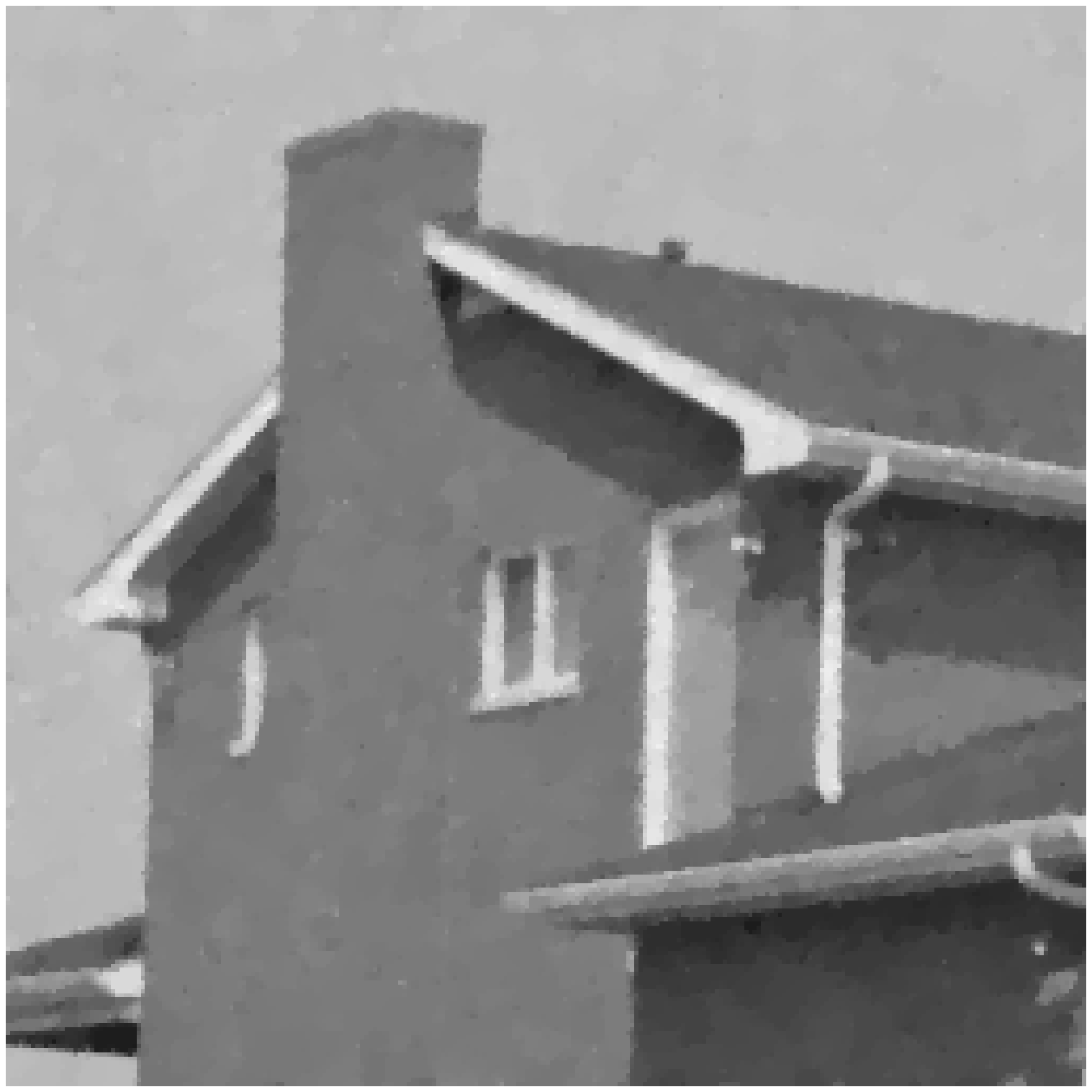}}%
\subfigure[TV OP recovery. PSNR = 30.6dB.]{\includegraphics[width=0.48\linewidth]{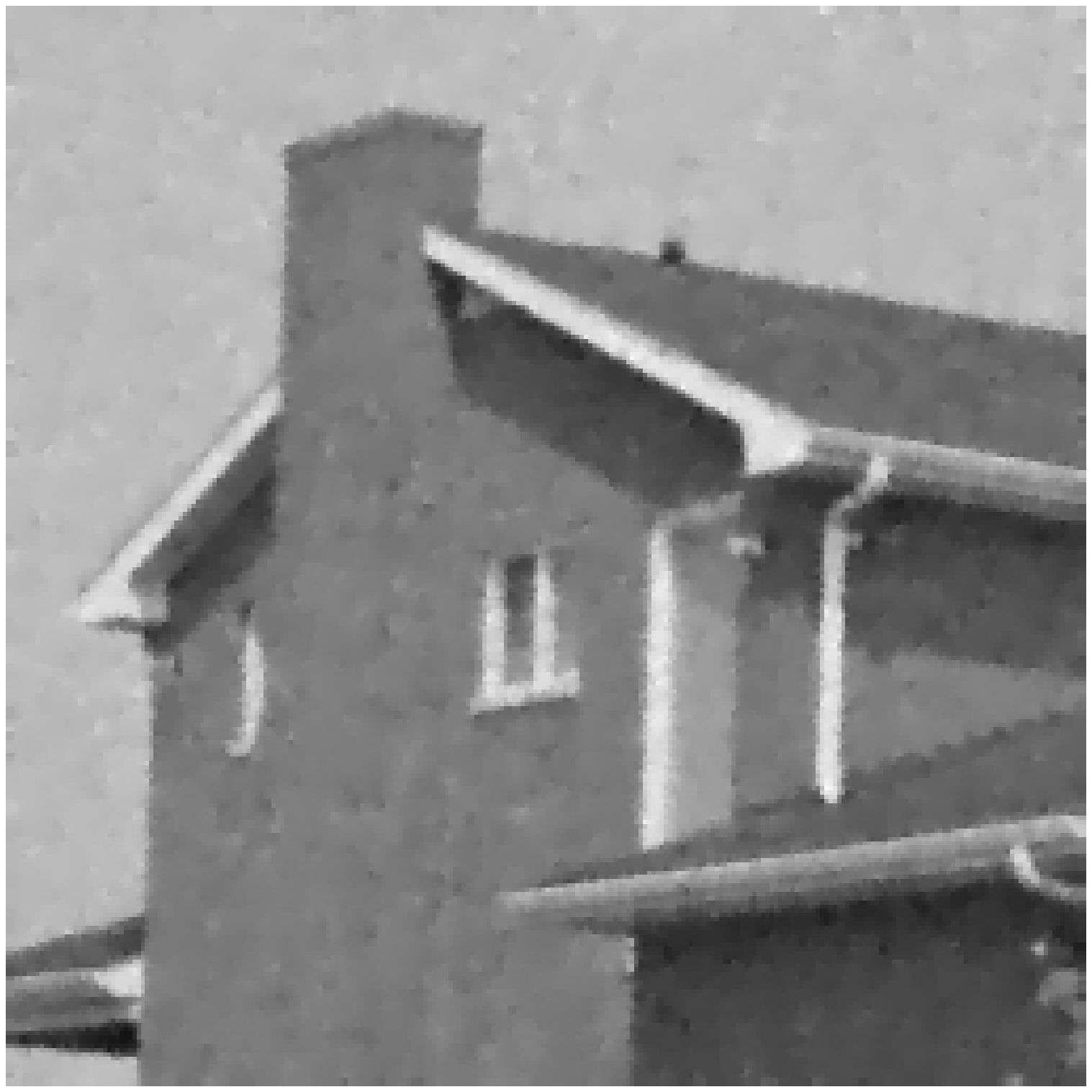}}%
}%
\caption{Denoising of {\em house} using the BGAPN algorithm. 
Notice that we do not have the staircasing effect that appears in the TV reconstruction. Because our model is linear we do not recover the texture and thus we get slightly inferior results compared to TV with respect to PSNR. Note that if we use a cubic overparameterization with BGAPN instead of linear we get PSNR (=31.81dB) better than that of TV.}
\label{fig:house_recovery}
\end{figure}

\begin{figure}[htb]
\centering
{\subfigure[Original Image Gradients]{\includegraphics[width=0.48\linewidth]{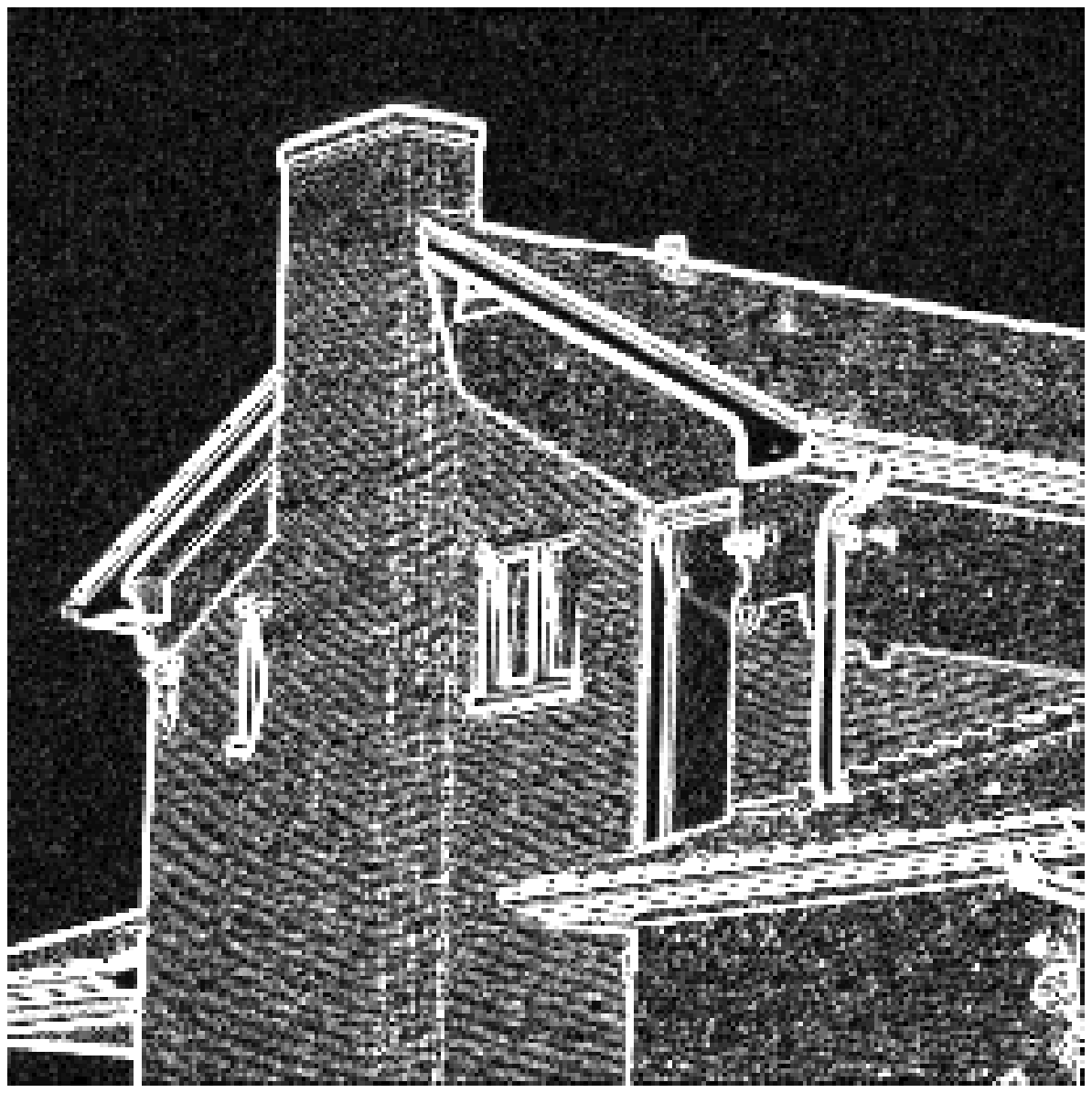}}%
\subfigure[Recovered Image Gradients]{\includegraphics[width=0.48\linewidth]{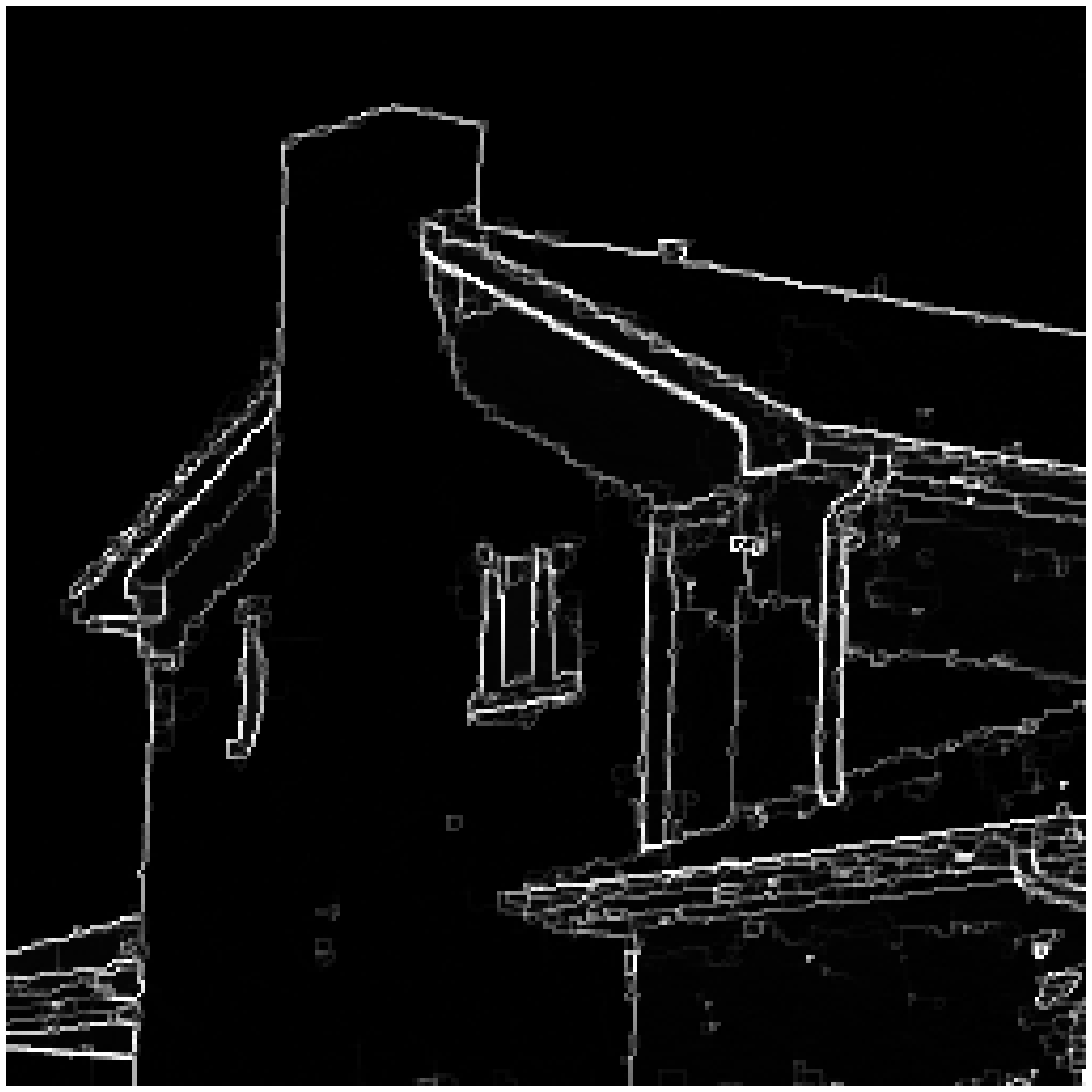}}
}%
\caption{Gradient map of the clean {\em house} image and our recovered image from Fig.~\ref{fig:house_recovery}.}
\label{fig:house_recovery_gradient}
\end{figure}

\subsection{Compressed Sensing of Piecewise Polynomial Functions}

We perform also a compressed sensing experiment in which we compare the performance of SSCoSAMP, with the optimal projection, and BGAPN for recovering a second order polynomial function with $6$ jumps from a small set of linear measurements. Each entry in the measurement matrix $\matr{M}$ is selected from an i.i.d normal distribution and then all columns are normalized to have a unit norm. The polynomial functions are selected as in the previous experiment but with two differences: (i) we omit the continuity constraint; and (ii) we normalize the signals to be with a unit norm. 

Fig.~\ref{fig:function_recovery_CS} presents the recovery rate (noiseless case $\sigma = 0$) of each program as a function of the number of measurements $m$. Note that for a very small or large number of samples BGAPN behaves better. However, in the middle range SSCoSaMP achieves a better reconstruction rate. Nonetheless, we may say that their performance is more or less the same.

\subsection{Cartoon Image Denoising}

We turn to evaluate the performance of our approach on images. 
A piecewise smooth model is considered to be a good model for images, and especially to \rg{the} ones with no texture, i.e., cartoon images \cite{Candes00Curvelets, Donoho01Sparse}. Therefore, we use a linear overparameterization of the two dimensional plane and \rg{employ} the two dimensional difference operator $\DIF$ that calculates the horizontal and vertical discrete derivatives of an image by applying the filters $[1, -1]$ and $[1, -1]^T$ on it. In this case, the problem in \eqref{eq:P0_analysis_e_overparam}  turns to be (notice that $\matr{M} = \matr{I}$)
\begin{eqnarray}
\label{eq:P0_analysis_e_overparam_denoising}
&& \hspace{-0.5in} \left[\hat{\vect{b}}_0^T, \hat{\vect{b}}_h^T, \hat{\vect{b}}_v^T \right]^T = \\ \nonumber && \min_{\tilde{\vect{b}}_0, \tilde{\vect{b}}_h, \tilde{\vect{b}}_v} \norm{\abs{\DIF\tilde{\vect{b}}_0}^2 + \abs{\DIF\tilde{\vect{b}}_h}^2 + \abs{\DIF\tilde{\vect{b}}_v}^2}_0 \\ \nonumber && \hspace{0.3in} s.t.  ~~ \norm{\vect{g} - \left[\matr{X}_0, \matr{X}_h, \matr{X}_v\right] \left[ \begin{array}{c}
\tilde{\vect{b}}_0 \\ \tilde{\vect{b}}_h \\ \tilde{\vect{b}}_v
\end{array}\right]}_2 \le \norm{\vect{e}}_2,
\end{eqnarray}
where $\matr{X}_0$, $\matr{X}_h, \matr{X}_v$ are the matrices that contain the DC, the horizontal and the vertical parameterizations respectively\rg{; and} $\hat{\vect{b}}_0, \hat{\vect{b}}_h, \hat{\vect{b}}_v$ are their corresponding space variables.

We apply this scheme for denoising two cartoon images,  {\em swoosh} and {\em sign}. 
We compare our results with the ones of TV denoising \cite{Rudin92Nonlinear}.
Figs.~\ref{fig:swoosh_recovery} and \ref{fig:sign_recovery} present the recovery of {\em swoosh} and {\em sign} from their noisy version contaminated with an additive white Gaussian noise with $\sigma = 20$.   Note that we achieve better recovery results than TV and do not suffer from its staircasing effect. We have \rg{tuned the parameters of TV separately for each image to optimize its output quality}, while we have used the same setup for our method in all the denoising experiments. To get a good quality with BGAPN, we run the algorithm several times with different set of parameters (which are the same for all images) and then provide as an output the average image of all the runs. Notice that using this technique with TV degrades its results.

To test whether our better denoising is just a result of using overparameterization or an outcome of our new framework, we compare also to TV with linear overparameterization \cite{Nir07Over}\footnote{Code provided by the authors.}. Notice that while plugging  overparameterization directly in TV improves the results in some cases \cite{Nir07Over}, this is not the case with the images here. Therefore, we see that our new framework that links sparsity with overparameterization has an advantage over the old approach that still acts within the variational scheme.

We could use other forms of overparameterizations such as cubical instead of planar or add other directions of the derivatives in addition to the horizontal and vertical ones. For example, one may apply our scheme also using an operator that calculates also the diagonal derivatives using the filters $\left[ \begin{array}{cc}
1 & 0 \\ 
0 &  -1 
\end{array}  \right]$ and $\left[ \begin{array}{cc}
0 & 1 \\ 
-1 &  0 
\end{array}  \right]$. 
Such choices may lead to an improvement in different scenarios. A future work should focus on learning the overparameterizations and the type of derivatives that should be used for denoising and for other tasks. We believe that such a learning has the potential to lead to state-of-the-art results.

\subsection{Image Segmentation}


\begin{figure*}[htb]
\centering
{\subfigure[Original Image]{\includegraphics[width=0.44\linewidth]{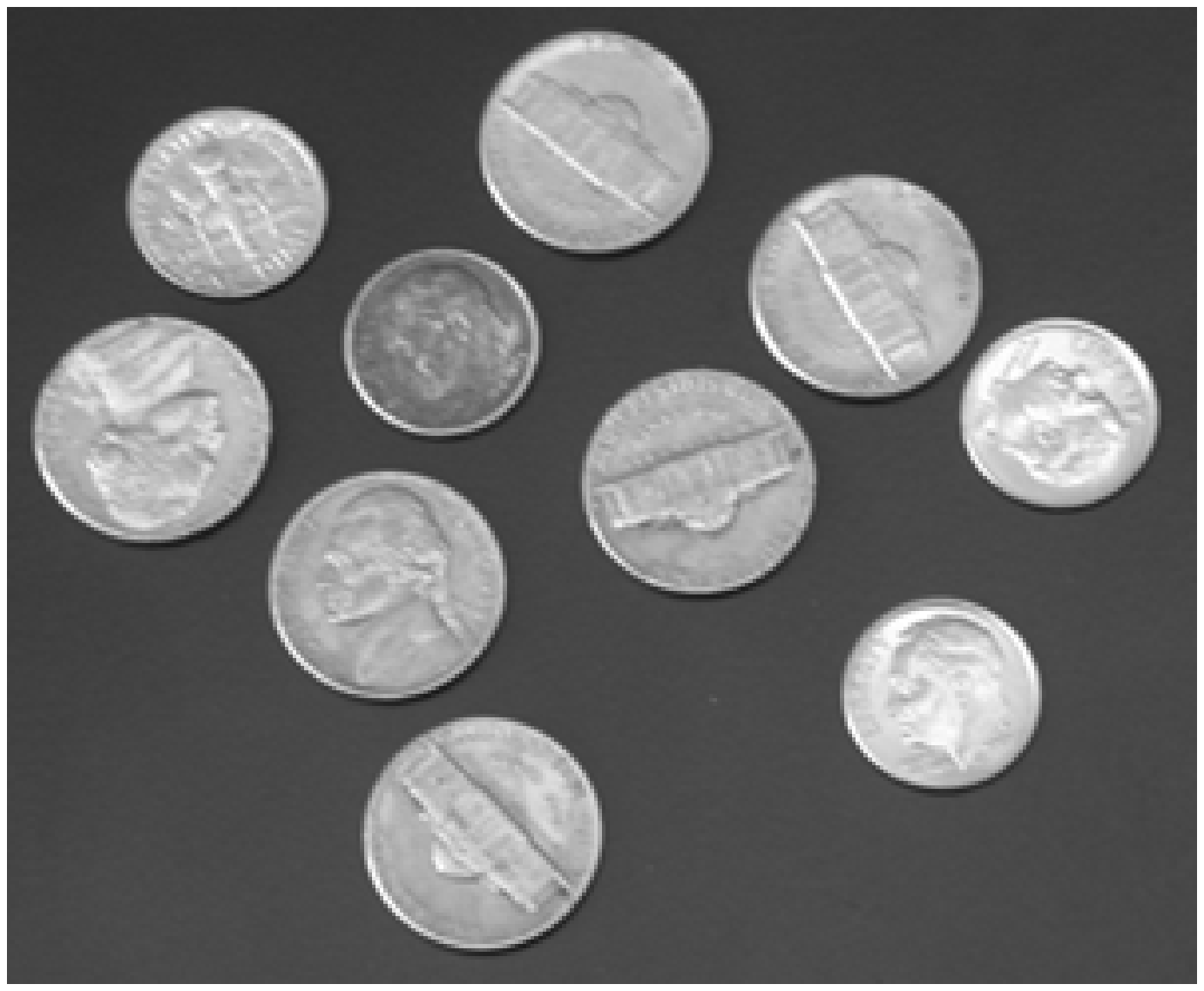}}%
\subfigure[Piecewise linear version of the image]{\includegraphics[width=0.44\linewidth]{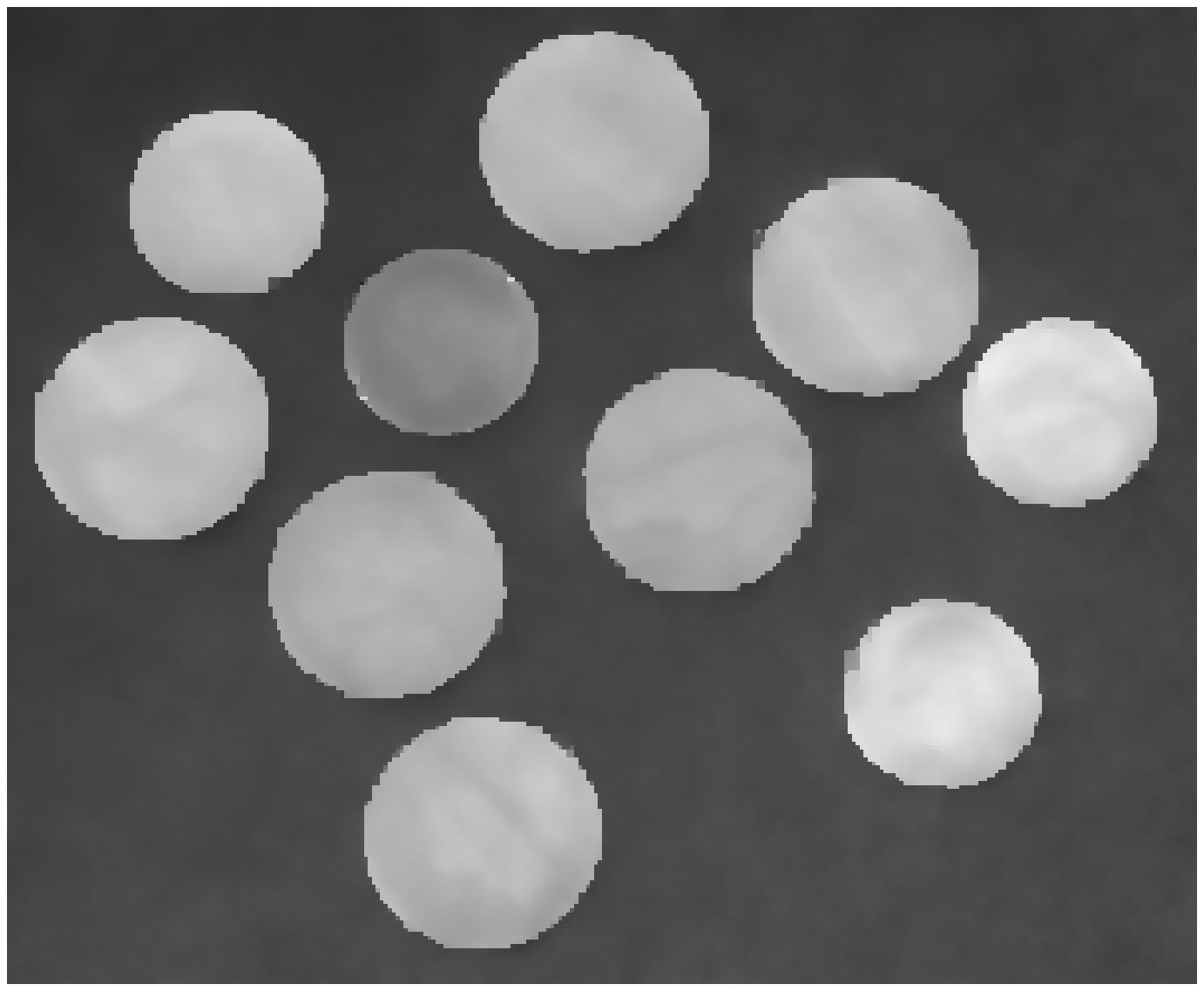}}
\subfigure[Image Segmentation]{\includegraphics[width=0.44\linewidth]{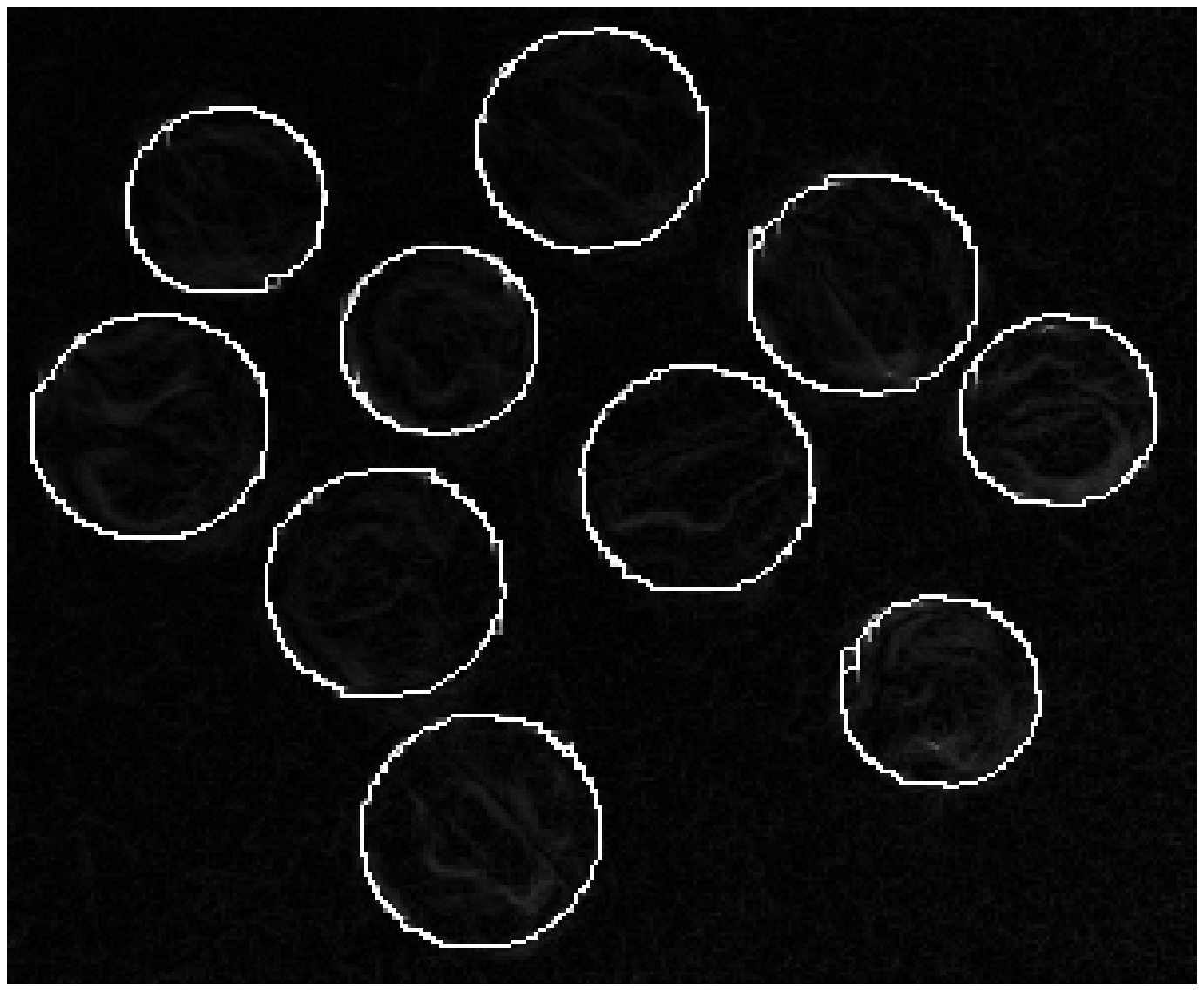}}
\subfigure[Image Segmentation using Graph-Cuts \cite{Felzenszwalb04Efficient}]{\includegraphics[width=0.44\linewidth]{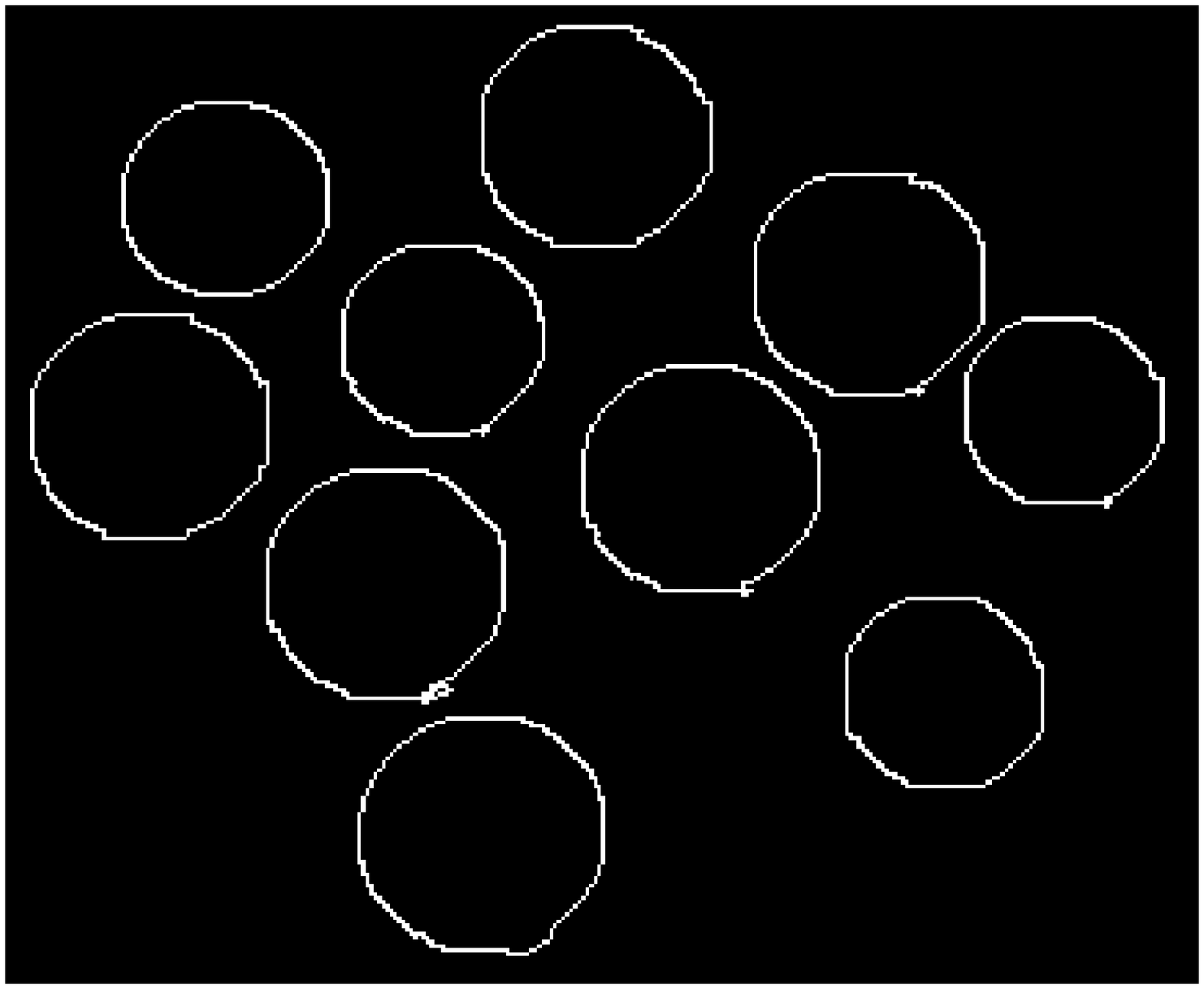}}
}%
\caption{Piecewise linear version of {\em coins} image together with the segmentation result. We compare to the popular graph-cuts based segmentation \cite{Felzenszwalb04Efficient}.}
\label{fig:coins_segmentation}
\end{figure*}

\begin{figure*}[htb]
\centering
{\subfigure[Original Image]{\includegraphics[width=0.44\linewidth]{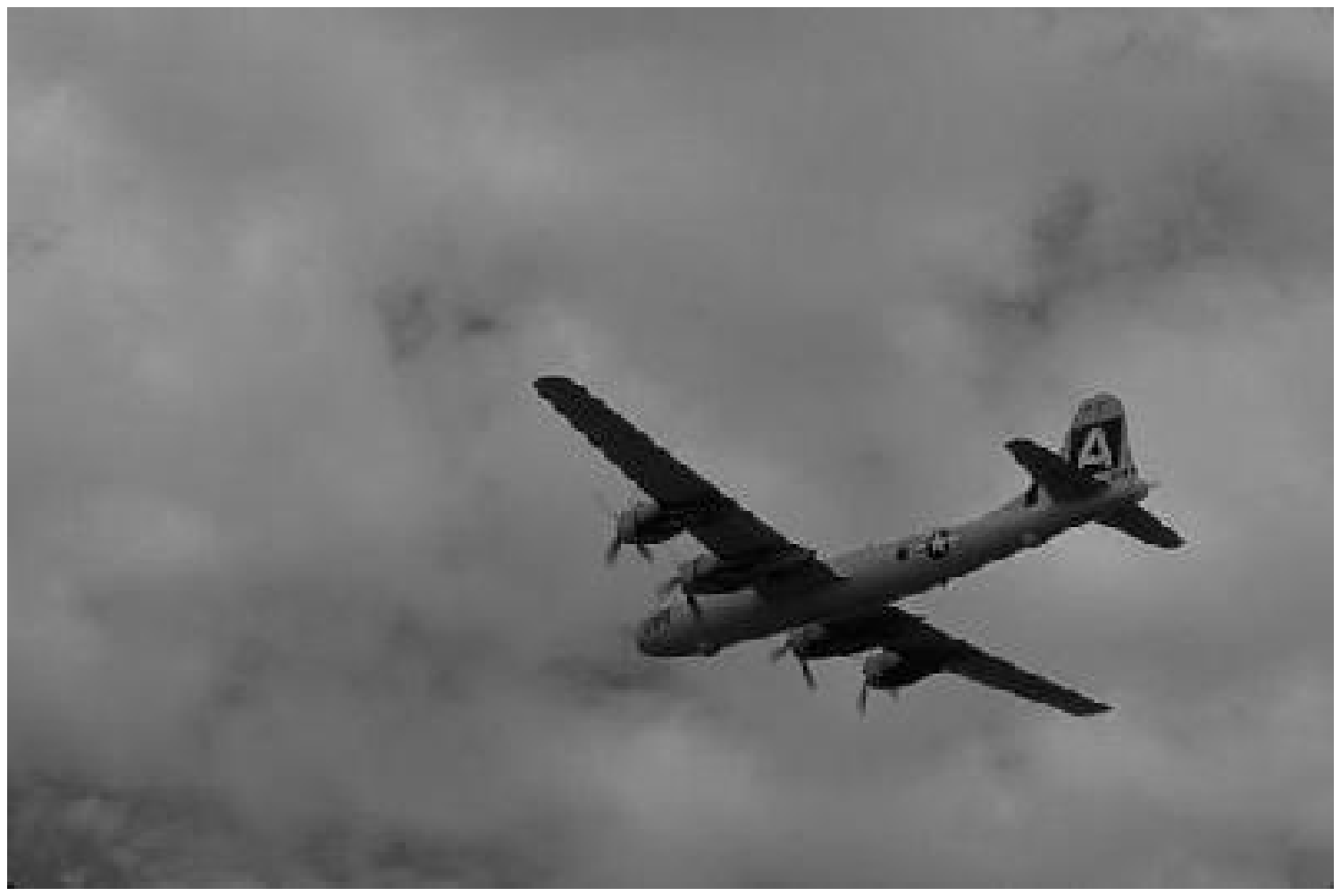}}%
\subfigure[Piecewise linear version of the image]{\includegraphics[width=0.44\linewidth]{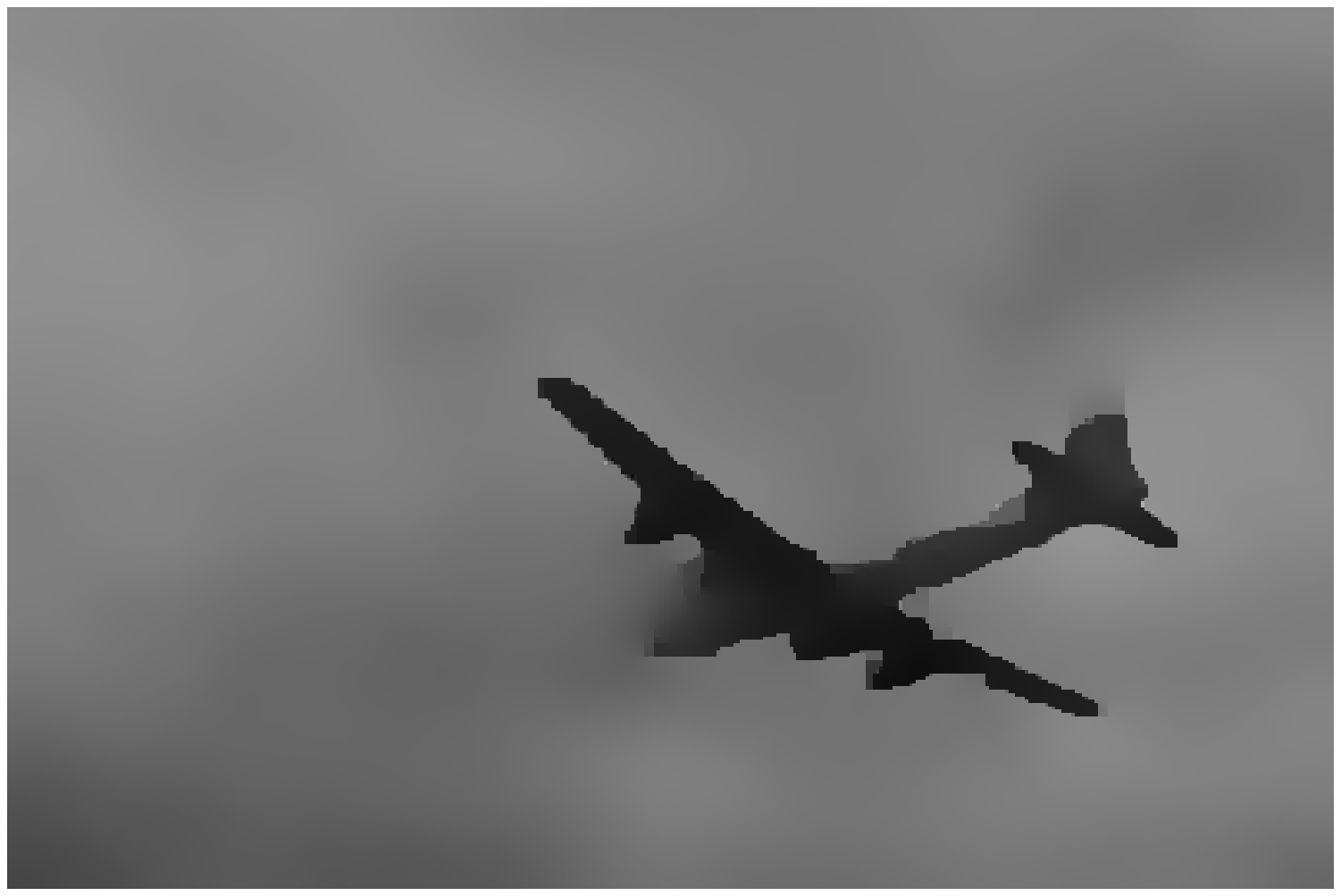}}
\subfigure[Image Segmentation]{\includegraphics[width=0.44\linewidth]{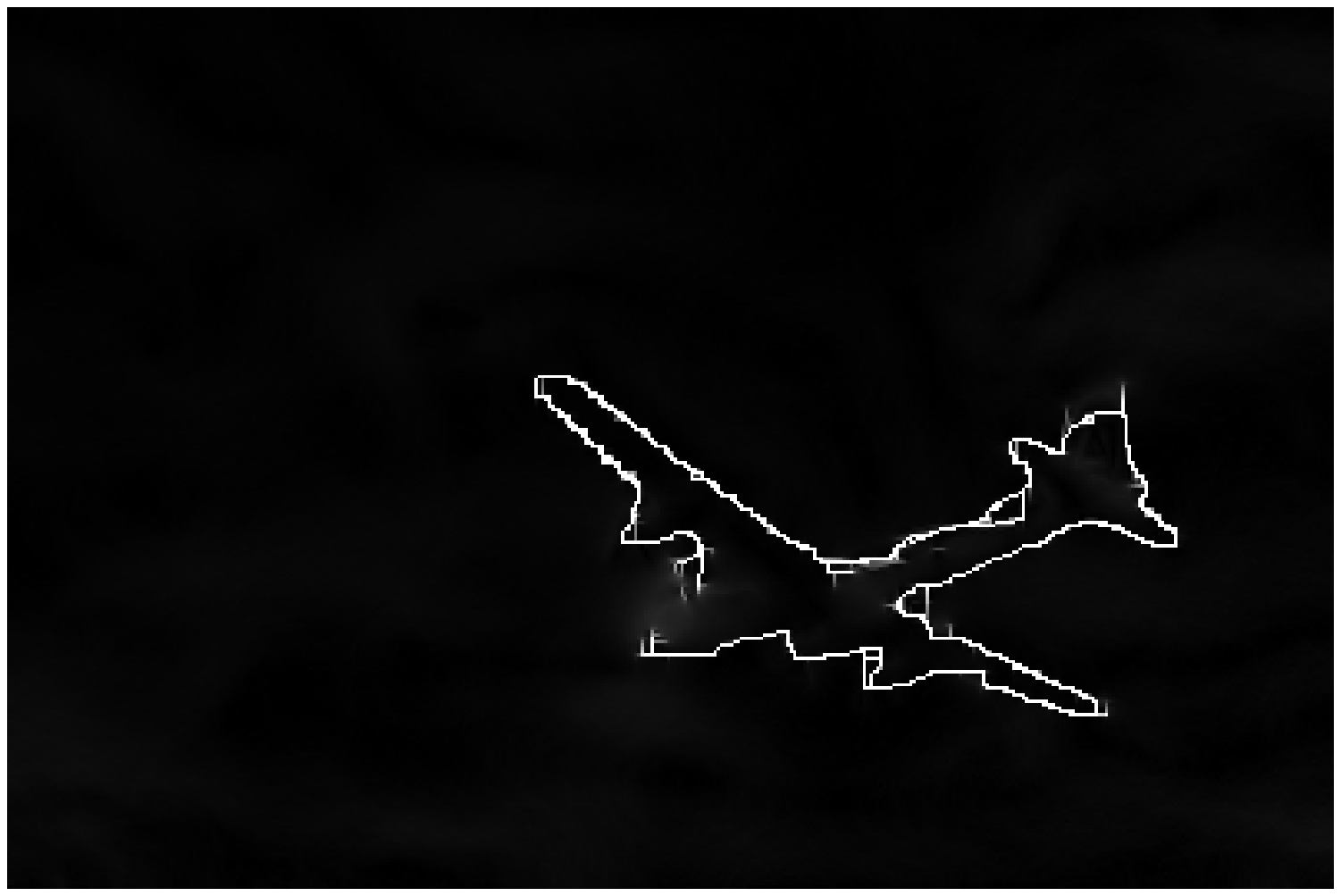}}
\subfigure[Image Segmentation using Graph-Cuts \cite{Felzenszwalb04Efficient}]{\includegraphics[width=0.44\linewidth]{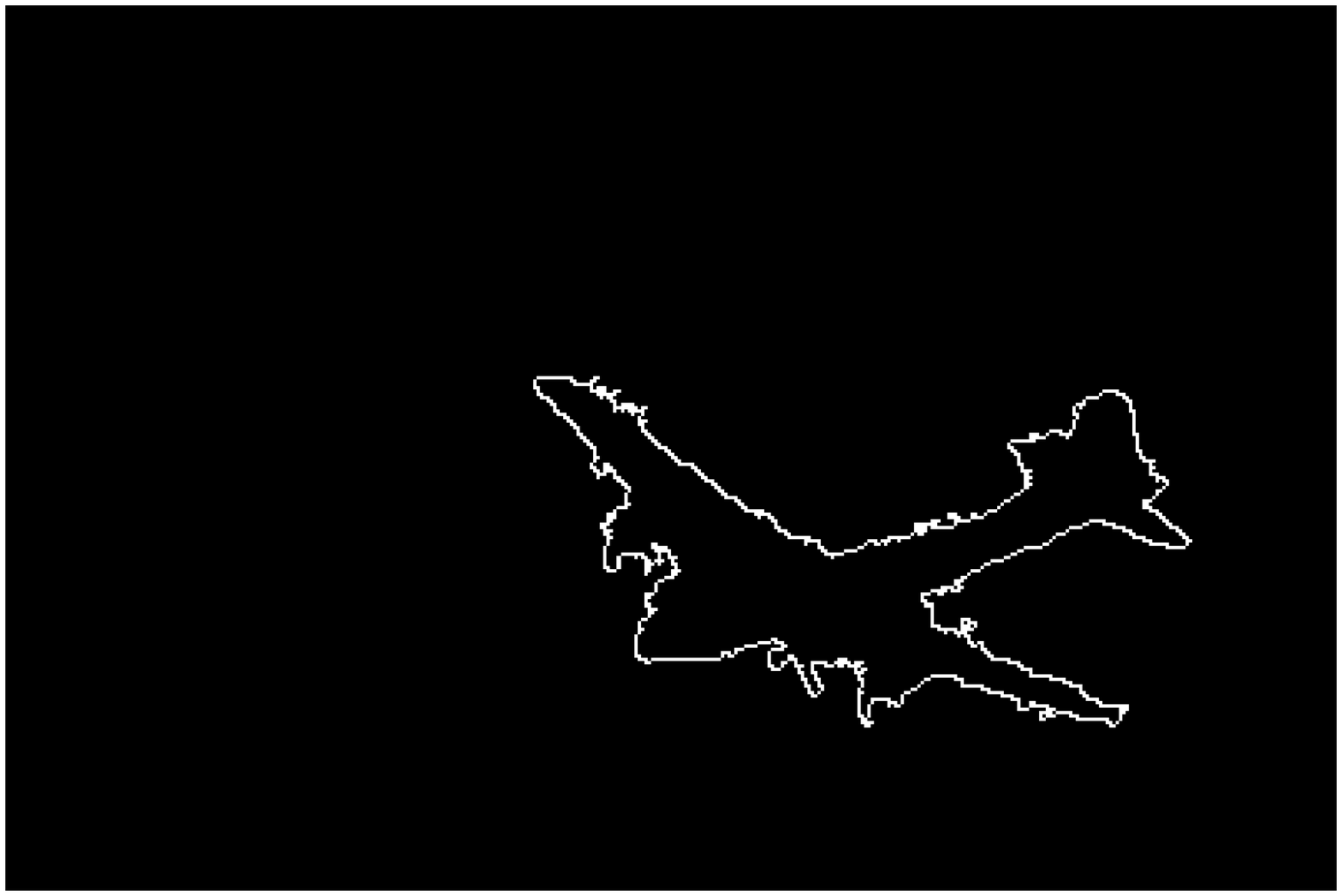}}
}%
\caption{Piecewise linear version of {\em airplane} image together with the segmentation result. We compare to the popular graph-cuts based segmentation \cite{Felzenszwalb04Efficient}.}
\label{fig:airplane_segmentation}
\end{figure*}

\begin{figure*}[htb]
\centering
{\subfigure[Original Image]{\includegraphics[width=0.44\linewidth]{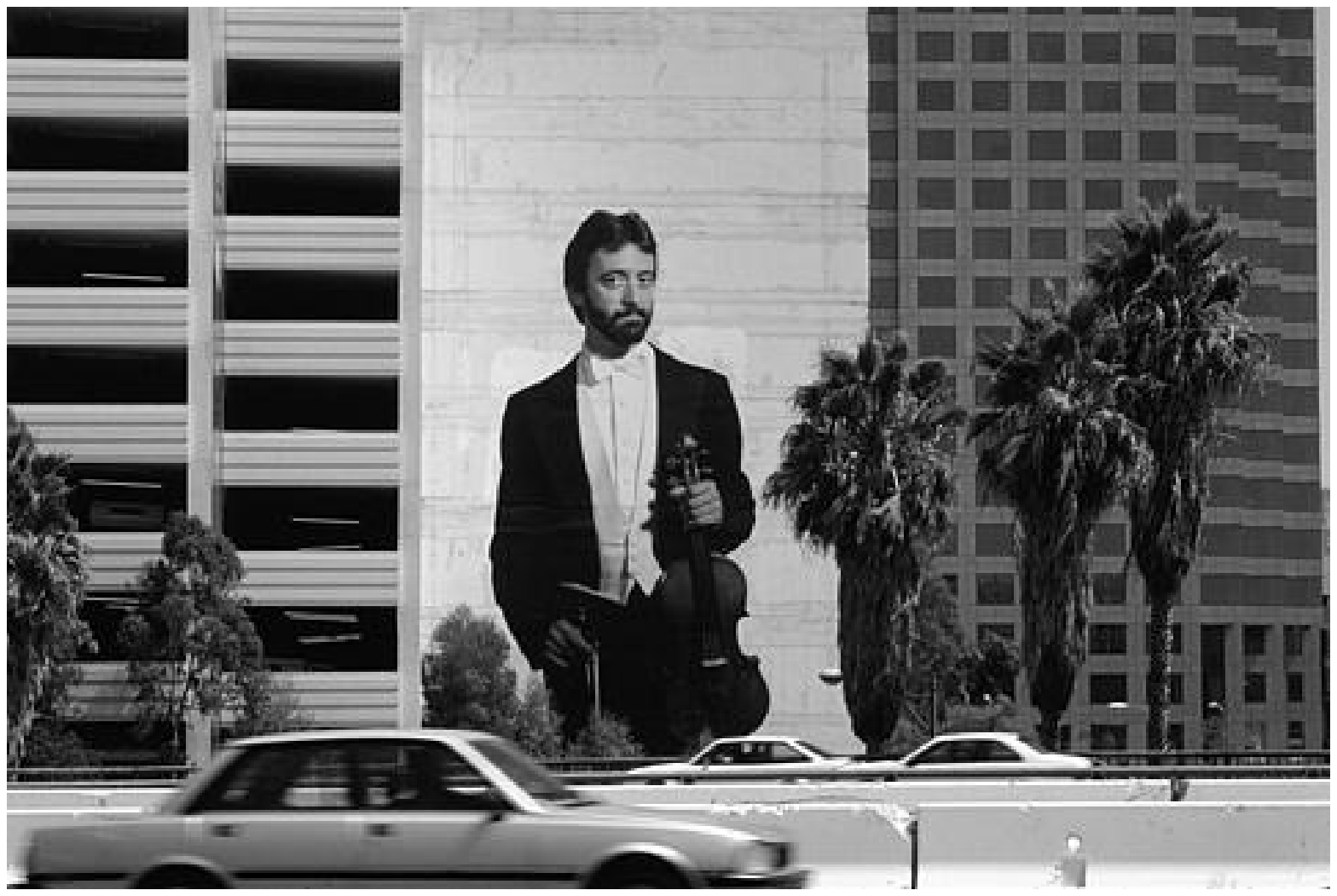}}%
\subfigure[Piecewise linear version of the image]{\includegraphics[width=0.44\linewidth]{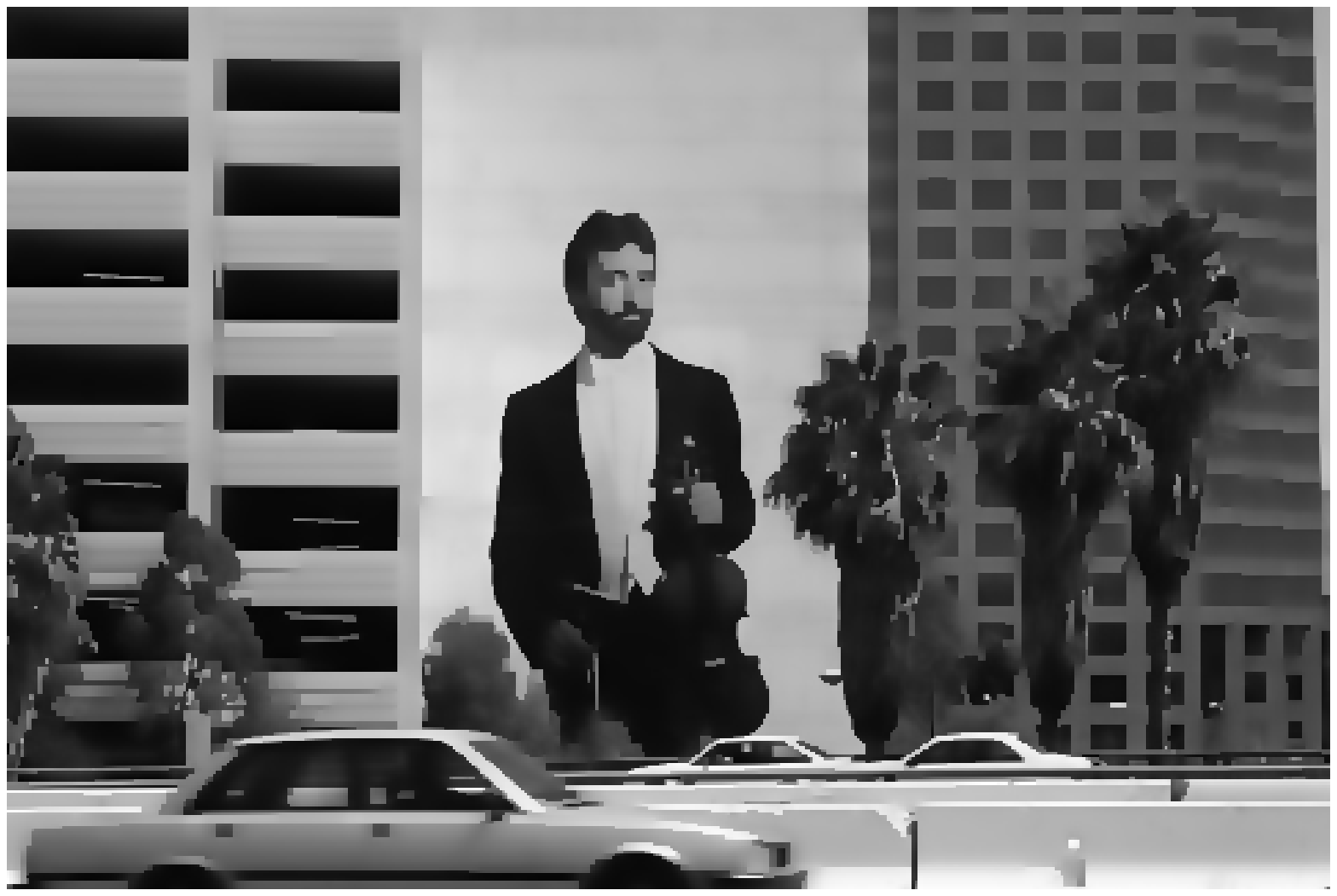}}
\subfigure[Image Segmentation]{\includegraphics[width=0.44\linewidth]{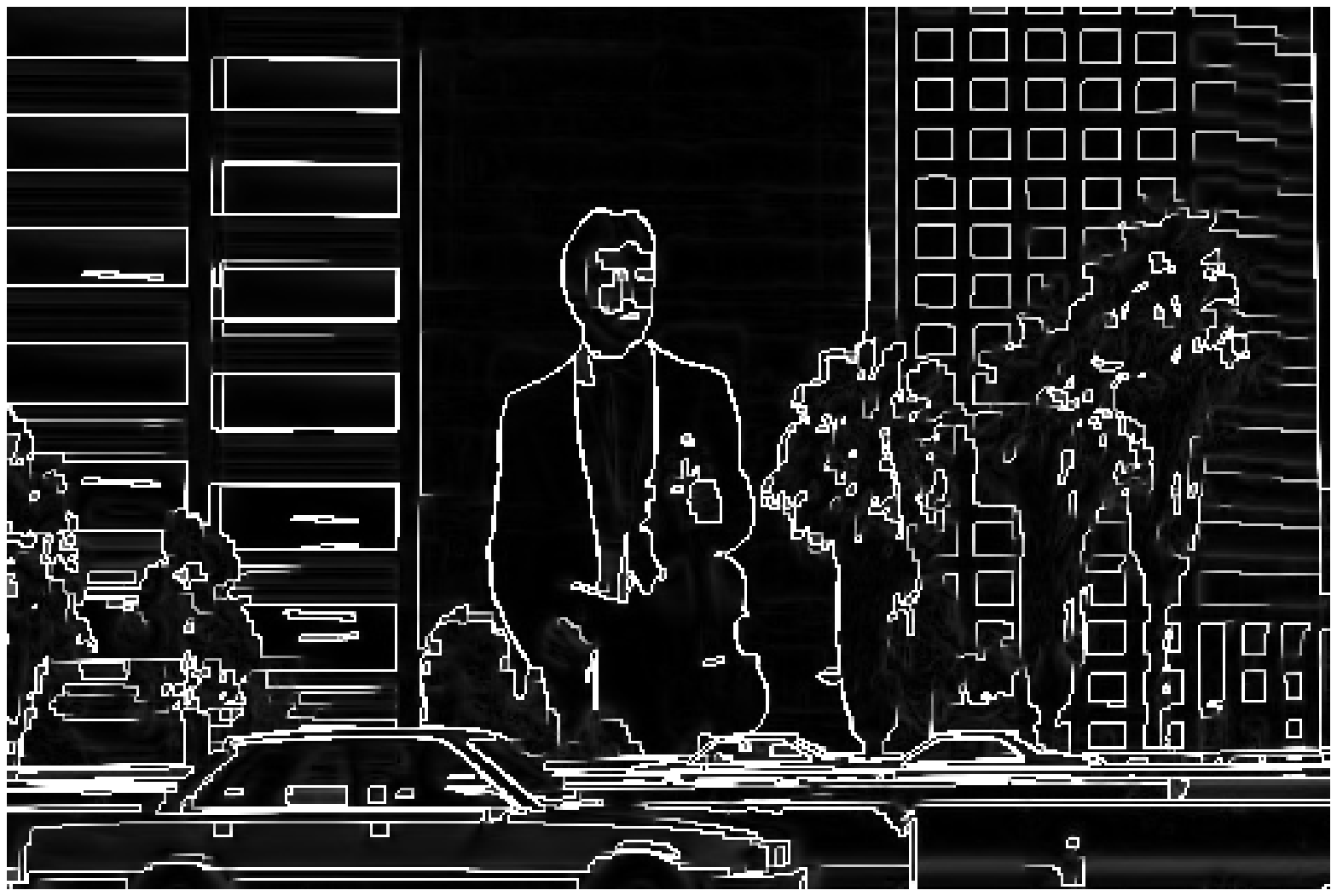}}
\subfigure[Image Segmentation using Graph-Cuts \cite{Felzenszwalb04Efficient}]{\includegraphics[width=0.44\linewidth]{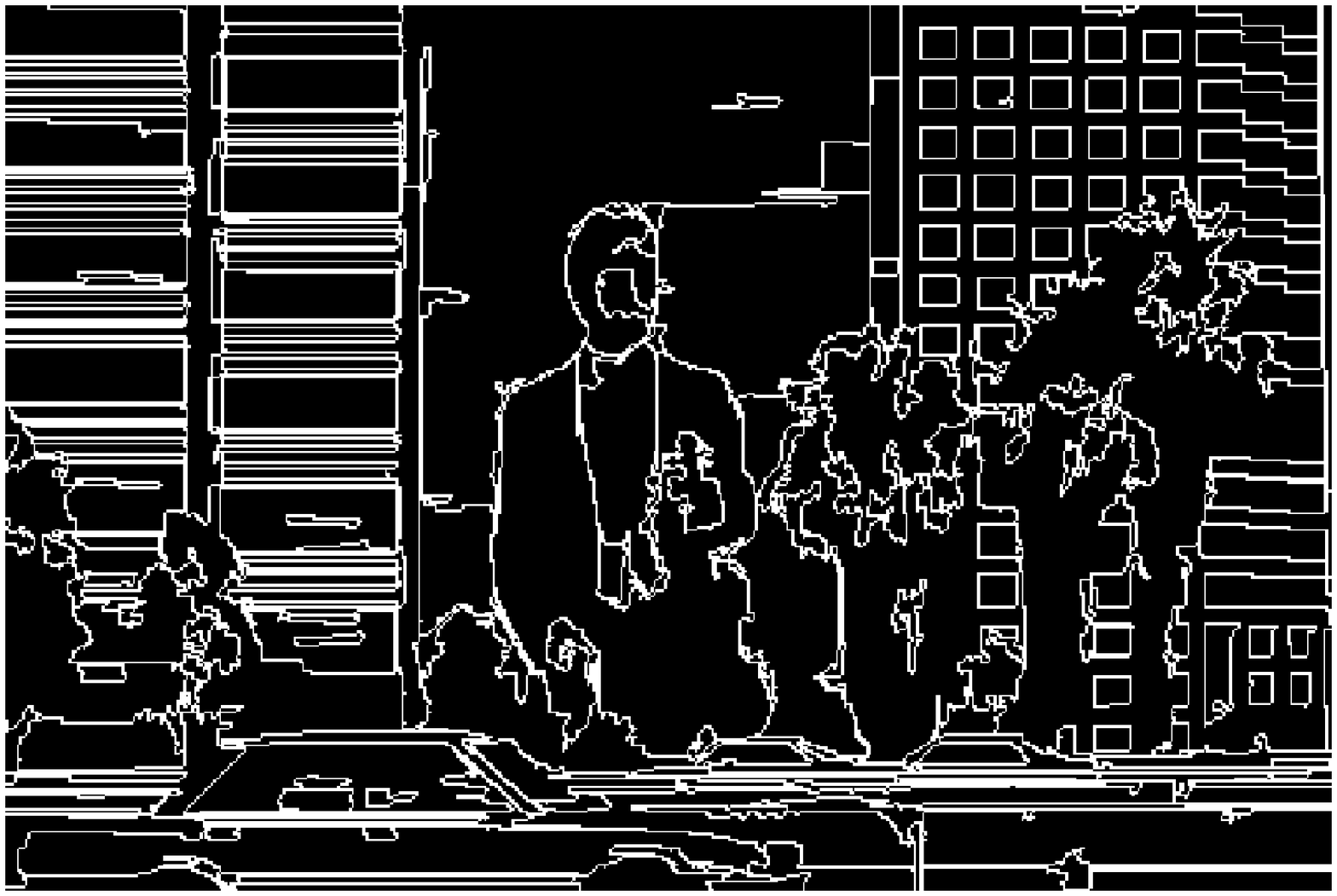}}
}%
\caption{Piecewise linear version of {\em man} image together with the segmentation result. We compare to the popular graph-cuts based segmentation \cite{Felzenszwalb04Efficient}.}
\label{fig:man_segmentation}
\end{figure*}

As a motivation for the task of segmentation we present the denoising of an image with a texture. We continue \rg{using} the model in \eqref{eq:P0_analysis_e_overparam_denoising} and \rg{consider} the {\em house} image as an example. 
Fig.~\ref{fig:house_recovery} demonstrates the denoising result we get for this image. Note that here as well we do not suffer from the staircasing effect that appears in the TV recovery. However, due to the nature of our model we loose the texture and therefore achieve an inferior PSNR compared to the TV denoising\footnote{The lower PSNR we get with our method is because our model is linear and therefore is less capable to adapt itself to the texture. By using a cubic overparameterization we get PSNR which is equal to the one of TV. Note also that for larger noise magnitudes the recovery performance of our algorithm in terms of PSNR becomes better than TV also with the linear model, \rg{as in these conditions, we tend to  loose the texture anyway.}}.  

Though the removal of texture is not favorable for the task of denoising,  
it makes the recovery of salient edges in the original image easier. In Fig.~\ref{fig:house_recovery_gradient} we present the gradient map of our recovered image and the one of the original image. It can be seen that while the gradients of the original image capture also the texture changes, \rg{our method finds} only the main edges\footnote{We have tested our approach also in the presence of a blur operator. The edges in this case are preserved as well.}. This motivates us to use our scheme for segmentation. 

Since our scheme divides the image into piecewise linear regions, we can view our strategy as an approach that minimizes the Mumford-Shah functional
\cite{Mumford89Optimal, Ambrosio90Approximation}.
On the other hand, if the image has only two regions, our segmentation result can be viewed as a solution of the Chan-Vese functional with the difference that we model each region by a polynomial function
instead of approximating it by a constant \cite{Chan01Active}.

We present our segmentation results for three images, and for each we display the  piecewise constant version of each image together with its boundary map. Our segmentation results appear in Figs.~\ref{fig:coins_segmentation}, \ref{fig:airplane_segmentation} and \ref{fig:man_segmentation}. 
We compare our results to the popular graph-cuts based segmentation \cite{Felzenszwalb04Efficient}. Notice that we achieve a comparable performance, where in some places our method behaves better and in others the strategy in \cite{Felzenszwalb04Efficient} provides a better result.

Though we get a good segmentation, it is clear that there is still a large room for  improvement compared to the current state-of-the-art. One direction for improvement is to use more filters within $\OM$. Another one is to calculate the gradients of the coefficients parameters and not of the recovered image as they are supposed to be truly piecewise constant. We leave these ideas to a future work.

\section{Conclusion and Future Work}
\label{sec:conc}

This work has presented a novel framework for solving the overparameterized variational problem using sparse representations. 
We have demonstrated how this framework can be used both for one dimensional and two dimensional functions, while a generalization to other higher dimensions (such as 3D) is straightforward.
We have solved the problem of line fitting for piecewise polynomial 1D signals and then shown how the new technique can be used for compressed sensing, denoising and segmentation.

Though this work has focused mainly on linear overparameterizations, the extension to other forms is straightforward. However, to keep the discussion as simple as possible, we have chosen to use simple forms of overparameterizations in the experiments section. As a future research, we believe that a learning process should be added to our scheme. It should adapt the functions of the space variables $\matr{X}_1, \dots, \matr{X}_n$ and the filters in $\OM$ to the signal at hand.
We believe that this has the potential to lead to state-of-the-art results in segmentation, denoising and other signal processing tasks. Combining of our scheme with the standard sparse representation approach may provide the possibility to add support to images with texture. This will lead to a scheme that works globally on the image for the cartoon part and locally for the texture part. 
Another route for future work is to integrate our scheme in the state-of-the-art overparameterized based algorithm for optical flow  in \cite{Nir08Over}.

\section*{Acknowledgment}
The authors would like to thank Tal Nir and Guy Roseman for providing their code for the experiments.
The research leading to these results has received funding from the European Research Council under European Unions Seventh Framework Program, ERC Grant agreement no. 320649.
This research is partially supported by AFOSR, ARO, NSF, ONR, and NGA.
The authors would like to thank the anonymous reviewers for their helpful and constructive comments that greatly contributed to improving  this paper.

\appendices 

\section{The SSCoSaMP Algorithm}
\label{sec:sscosamp}

\begin{algorithm}[t]
\caption{Signal Space CoSaMP (SSCoSaMP) for Piecewise Polynomial Functions} \label{alg:SSCoSaMP}
\begin{algorithmic}[l]

\REQUIRE $k, \matr{M}, \vect{g}, \gamma$, where $\vect{g} = \matr{M}\vect{f}
+ \vect{e}$, $\vect{f} = \left[\matr{I}, \matr{Z}, \matr{Z}^2, \dots, \matr{Z}^n \right] \left[ 
\vect{b}_0^T, \vect{b}_1^T, \dots, \vect{b}_n^T
\right]^T$ is a piecewise polynomial function of order $n$,  
 $k = \norm{\sum_{i=0}^n \abs{\DIF\vect{b}_i} }_0$ is the number of jumps in the representation coefficients of $\vect{f}$, $\vect{e}$ is
an additive noise and $\gamma$ is a parameter of the algorithm.  $S_n(\cdot, k)$ is a procedure that approximates a given signal by a piecewise polynomial function of order $n$ with $k$ jumps.

\ENSURE $\hat{\vect{f}}$: A piecewise polynomial with $k+1$ segments that approximates
$\vect{f}$.

\STATE $\bullet$ Initialize the jumps' locations $T^0 =\emptyset$, the residual $\vect{g}_r^0 = \vect{g}$ and set $t = 0$.

\WHILE{halting criterion is not satisfied}

\STATE $\bullet$ $t = t + 1$. 

\STATE $\bullet$ Find the parameterization $\vect{b}_{r,0}, \vect{b}_{r,1}, \dots, \vect{b}_{r,n}$ of the residual's polynomial approximation by calculating $S_n(\matr{M}^T\vect{g}^{t - 1}_r, \gamma k)$.

\STATE $\bullet$ Find new temporal jump locations: $T_\Delta = $ the support of $\sum_{i=0}^n \abs{\DIF\vect{b}_{r,i}} $.

\STATE $\bullet$ Update the jumps locations' indices: $\tilde{T}^t = T^{t -1} \cup
T_\Delta$.

\STATE $\bullet$ Compute temporal parameters: $\left[\vect{b}_{p,0},  \dots, \vect{b}_{p,n} \right] = \argmin_{\tilde{\vect{b}}_{0},  \dots, \tilde{\vect{b}}_n}\norm{\vect{g} -\matr{M} \left[\matr{I}, \matr{Z}, \matr{Z}^2 \dots, \matr{Z}^n \right] \left[ \begin{array}{c}
\tilde{\vect{b}}_0 \\ \tilde{\vect{b}}_1 \\ \vdots \\ \tilde{\vect{b}}_n
\end{array}\right]}_2^2$ $\text{ s.t. }$ \\ ~~~~~~ $(\DIF\tilde{\vect{b}}_0 )_{(\tilde{T}^t)^C} = 0$, $\dots$ $(\DIF\tilde{\vect{b}}_n )_{(\tilde{T}^t)^C} = 0$.

\STATE $\bullet$ Calculate a polynomial approximation of order $n$: \\~~${\vect{f}}^t =S_n(\left[\matr{I}, \matr{Z}, \matr{Z}^2, \dots, \matr{Z}^n \right] \left[ 
\vect{b}^T_{p,0}, \dots, \vect{b}_{p,n}^T
\right]^T,k)
$.

\STATE $\bullet$ Find new  jump locations: $T^t = $ the locations of the jumps in
the parameterization of $\vect{f}^t$.

\STATE $\bullet$ Update the residual:
$\vect{g}_r^t = \vect{g} - \matr{M}{\vect{f}}^t$.

\ENDWHILE
\STATE $\bullet$ Form final solution $\hat{\vect{f}} = {\vect{f}}^t$.
\end{algorithmic}
\end{algorithm}

For approximating \eqref{eq:P0_analysis_k_poly_overparam}, we use a block sparsity variant of SSCoSaMP \cite{Giryes14NearOracle} and adapt it to our model.
It is presented in Algorithm~\ref{alg:SSCoSaMP}. 
Due to the equivalence between $\matr{D}_{HS}$ and $\OM_{DIF}$, we use the latter in the algorithm. 

This method uses a projection $S_n(\cdot, k)$ that 
given a signal finds its closest piecewise polynomial functions with $k$ jump points. 
We calculate this projection using dynamic programming. Our strategy is a generalization of the one that appears in \cite{ Giryes14Greedy, Han04Optimal} and is presented in the next subsection.

The halting criterion we use in our work in Algorithm~\ref{alg:SSCoSaMP} is $\norm{\vect{g}_r^t}_2 \le \epsilon$ for a given small constant $\epsilon$. Other options for stopping criteria are discussed in \cite{Needell09CoSaMP}.

 \subsection{Optimal Approximation using Piecewise Polynomial Functions}
 \label{sec:opt_poly_func}

Our projection technique uses the fact that once the jump points are set, the optimal parameters of the polynomial in a segment $[t,l]$ can be calculated optimally by solving a least squares minimization problem
\begin{eqnarray}
\label{eq:Pn_Gtl_LS}
\norm{\left[\matr{I}[t,l], \matr{Z}[t,l], \dots, \matr{Z}^n[t,l] \right] \left[ \begin{array}{c}
\vect{b}_0[t,l] \\ \vect{b}_1[t,l] \\ \vdots \\ \vect{b}_n[t,l]
\end{array}\right]  - \vect{g}[t,l]}_2^2,
\end{eqnarray}
where $\vect{g}[t,l]$ is the sub-vector of $\vect{g}$ supported by the indices $t$ to $l$ ($t \le l$) and $\matr{Z}^i[t,l]$ is the (square) sub-matrix of $\matr{Z}^i$ corresponding to the indices $t$ to $l$.
We denote by $P_n(\vect{g}[t,l])$ the polynomial function we get by  solving \eqref{eq:Pn_Gtl_LS}.
Indeed, in the case that the size of the segment $[t,l]$ is smaller than the number of parameters, e.g. segment of size one for a linear function, the above minimization problem has infinitely many options for setting the parameters. However, all of them lead to the same result, which is keeping the values of the points in the segment, i.e., having $P_n(\vect{g}[t,l]) = \vect{g}[t,l]$.

Denote by $S_n(\vect{g}[1,\tilde{d}], k)$ the optimal approximation of the signal $\vect{g}[1,\tilde{d}]$ by a piecewise polynomial function with $k$ jumps. It can be calculated by solving the following recursive minimization problem  
\begin{eqnarray}
\label{eq:S_gk_i_recursion}
\hat{t} = \argmin_{1 \le t < \tilde{d}} && \norm{S_n(\vect{g}[1,t], k -1) - \vect{g}[1,t]}_2^2 \\ \nonumber &&  + \norm{P_n(\vect{g}[t+1,\tilde{d}]) - \vect{g}[t+1,\tilde{d}]}_2^2,
\end{eqnarray}
 and setting 
 \begin{eqnarray}
 \label{eq:S_gk_composition}
 S_n(\vect{g}[1,\tilde{d}], k) = \left[ \begin{array}{c}
S_n(\vect{g}[1,\hat{t}], k -1)  \\ P_n(\vect{g}[\hat{t}+1,\tilde{d}])
\end{array} \right].
\end{eqnarray}
The vectors $S_n(\vect{g}[1,t], k -1)$ can be calculated recursively using \eqref{eq:S_gk_i_recursion}. The recursion ends with the base case $S_n(\vect{g}[1,{t}], 0) = P_n(\vect{g}[1,{t}])$.

This leads us to the following algorithm for calculating an optimal approximation for a signal $\vect{g}$. Notice that this algorithm provides us also with the parameterization of a piecewise polynomial.
\begin{enumerate}
\item Calculate $S_n(\vect{g}[1,t], 0) = P_n(\vect{g}[1,{t}])$ for $1 \le t \le  d$. 
\item For $\tilde{k} = 1: k-1$ do
\begin{itemize}
\item Calculate $ S_n(\vect{g}[1,\tilde{d}], \tilde{k})$ for $ 1 \le \tilde{d} \le d$ using \eqref{eq:S_gk_i_recursion} and \eqref{eq:S_gk_composition}.
\end{itemize}
\item Calculate $ S_n(\vect{g}[1,d], k)$ using \eqref{eq:S_gk_i_recursion} and \eqref{eq:S_gk_composition}.
\end{enumerate}
Denoting by $T$ the worst case complexity of calculating $P_n(\vect{g}[t,l])$ for any pair $t,l$, we have that the complexity of step 1) is $O(dT)$; of step 2) is $O(kd^2(T+d))$, as the computation of the projection error is of complexity $O(d)$; and of step 3) $O(d(T+d))$.
Summing all together we get a total complexity of  $O(kd^2(T+d))$ for the algorithm, which is a polynomial complexity since $T$ is polynomial.

\section{The Block GAPN Algorithm}
\label{sec:bgapn}

For approximating \eqref{eq:P0_analysis_e_overparam}, we extend the GAPN technique \cite{Nam11GAPN} to block sparsity and adapt it to our model.
It is presented in Algorithm~\ref{alg:GAPN}. Notice that this program, unlike SSCoSaMP, does not assume the knowledge of $k$ or the existence of an optimal projection onto the signals' low dimensional union of subspaces.
Note also that it suits a general form of overparameterization and not only 1D piecewise polynomial functions.
It is possible to accelerate BGAPN for highly scaled problems by removing from the cosupport several elements at a time instead of one in the {\em update cosupport} stage. 

Ideally, we would expect that after several iterations of updating the cosupport in BGAPN we would have $\OM_\Lambda\hat{\vect{b}}_i= 0$. However, many signals are only nearly cosparse, i.e., have $k$ significantly large values in $\OM\vect{b}_i$ while the rest  are smaller than a small constant $\epsilon$. Therefore, a natural stopping criterion in this case would be to stop when the maximal value in $\abs{\OM\hat{\vect{b}}_i}$ is smaller than $\epsilon$. This is the stopping criterion we use throughout this paper for BGAPN. 
Of course, this is not the only option for a stopping criterion, e.g. one may look at the relative solution change in each iteration or use a constant number of iterations if $k$ is foreknown.

\begin{algorithm}[t]
\caption{The Block GAPN Algorithm} \label{alg:GAPN}
\begin{algorithmic}

\REQUIRE $\matr{M}, \vect{g}, \OM$, where $\vect{g} = \matr{M}\vect{f}
+ \vect{e}$,  $\vect{f} = \left[\matr{X}_1, \dots, \matr{X}_n \right]\left[ \vect{b}_1^T, \dots, \vect{b}_n^T \right]^T$ such that $\sum_{i}^n\abs{\matr{\OM}\vect{b}_i}$ is sparse, and $\vect{e}$ is
an additive noise.

\ENSURE $\hat{\vect{f}} = \left[\matr{X}_1, \dots, \matr{X}_n \right]\left[ \hat{\vect{b}}_1^T, \dots, \hat{\vect{b}}_n^T \right]^T$: an estimate for $\vect{f}$ such that $\sum_{i}^n\abs{\matr{\OM}\hat{\vect{b}}_i}$ is sparse.

\STATE Initialize cosupport $\Lambda = \left\{1, \dots, p \right\}$ and set $t = 0$.

\WHILE{halting criterion is not satisfied}

\STATE $t = t + 1$.

\STATE Calculate a new estimate: \begin{eqnarray}
\label{eq:gapn_min}
&& \hspace{-0.3in}  \left[{\hat{\vect{b}}}_1^T, \dots, \hat{\vect{b}}_n^T\right]^T =  \argmin_{\tilde{\vect{b}}_1, \dots, \tilde{\vect{b}}_n} \sum_{i=1}^n\norm{\OM_\Lambda\tilde{\vect{b}}_i}_2^2 \\ \nonumber && s.t.  ~~ \norm{\vect{g} - \matr{M}\left[\matr{X}_1, \dots \matr{X}_n\right] \left[ \begin{array}{c}
\tilde{\vect{b}}_1 \\ \vdots \\ \tilde{\vect{b}}_n
\end{array}\right]}_2 \le \norm{\vect{e}}_2.
\end{eqnarray}

\STATE Update cosupport: $\Lambda = \Lambda \setminus \left\{ \argmax_j \sum_{i=1}^n\norm{\OM_j\hat{\vect{b}}_i}_2^2 \right\}$.

\ENDWHILE

\STATE Form an estimate for the original signal: $\hat{\vect{f}} = \left[\matr{X}_1, \dots, \matr{X}_n \right]\left[\hat{\vect{b}}_1^T, \dots,  \hat{\vect{b}}_n^T \right]^T$.

\end{algorithmic}
\end{algorithm}

We present also a modified version of BGAPN in Algorithm~\ref{alg:GAPN_continue} that imposes a continuity constraint on the change points. This is done by creating a binary diagonal matrix $\matr{W} = \diag(w_1, \dots, w_p )$ such that in each iteration of the program the $i$-th element $w_i$ is $1$ if it corresponds to a change point and zero otherwise. This matrix serves as a weights matrix to penalize discontinuity in the change point.
This is done by adding the regularizing term $$\gamma \norm{\matr{W}\OM\left[\matr{X}_1, \dots \matr{X}_n\right] \left[ \begin{array}{c}
\tilde{\vect{b}}_1 \\ \vdots \\ \tilde{\vect{b}}_n
\end{array}\right]}_2^2$$ to the minimization problem in \eqref{eq:gapn_min} (Eq. \eqref{eq:gapn_cont_min} in Algorithm~\ref{alg:GAPN_continue}), which leads to the additional step (Eq. \eqref{eq:cont_min_poly}) in the modified program. 

\begin{algorithm}[t]
\caption{The Block GAPN Algorithm with Continuity Constraint} \label{alg:GAPN_continue}
\begin{algorithmic}

\REQUIRE  $\matr{M}, \vect{g}, \OM$, $\gamma$, where $\vect{g} = \matr{M}\vect{f}
+ \vect{e}$,  $\vect{f} = \left[\matr{X}_1, \dots, \matr{X}_n \right]\left[ \vect{b}_1^T, \dots, \vect{b}_n^T \right]^T$ such that $\sum_{i}^n\abs{\matr{\Omega}\vect{b}_i}$ is sparse, $\vect{e}$ is
an additive noise, and $\gamma$ is a weight for the continuity constraint.

\ENSURE $\hat{\vect{f}} = \left[\matr{X}_1, \dots, \matr{X}_n \right]\left[ \hat{\vect{b}}_1^T, \dots, \hat{\vect{b}}_n^T \right]^T$: an estimate for $\vect{f}$ such that $\sum_{i}^n\abs{\matr{\OM}\hat{\vect{b}}_i}$ is sparse.

\STATE Initialize cosupport $\Lambda = \left\{1, \dots, p \right\}$ and set $t = 0$.

\WHILE{halting criterion is not satisfied}

\STATE $t = t + 1$.

\STATE Calculate a new estimate: \begin{eqnarray}
\label{eq:gapn_cont_min}
&& \left[{\hat{\vect{b}}}_1^T, \dots, \hat{\vect{b}}_n^T\right]^T =  \argmin_{\tilde{\vect{b}}_1, \dots, \tilde{\vect{b}}_n} \sum_{i=1}^n\norm{\OM_\Lambda\tilde{\vect{b}}_i}_2^2 \\ \nonumber && \hspace{0.3in} s.t.  ~~ \norm{\vect{g} - \matr{M}\left[\matr{X}_1, \dots \matr{X}_n\right] \left[ \begin{array}{c}
\tilde{\vect{b}}_1 \\ \vdots \\ \tilde{\vect{b}}_n
\end{array}\right]}_2 \le \norm{\vect{e}}_2.
\end{eqnarray}

\STATE Update cosupport: $\Lambda = \Lambda \setminus \left\{ \argmax_j \sum_{i=1}^n\norm{\OM_j\hat{\vect{b}}_i}_2^2 \right\}$.

\STATE Create Weight Matrix: $\matr{W} = \diag(w_1, \dots, w_p )$, where $w_i = 0$ if $i \in \Lambda$ or $w_i = 1$ otherwise.

\STATE Recalculate the estimate: \begin{eqnarray}
\label{eq:cont_min_poly}
&& \hspace{-0.3in} \left[{\hat{\vect{b}}}_1^T, \dots, \hat{\vect{b}}_n^T\right]^T =  \argmin_{\tilde{\vect{b}}_1, \dots, \tilde{\vect{b}}_n} \sum_{i=1}^n\norm{\OM_\Lambda\tilde{\vect{b}}_i}_2^2 \\ \nonumber && \hspace{0.4in}+ \gamma \norm{\matr{W}\OM\left[\matr{X}_1, \dots \matr{X}_n\right] \left[ \begin{array}{c}
\tilde{\vect{b}}_1 \\ \vdots \\ \tilde{\vect{b}}_n
\end{array}\right]}_2^2\\ \nonumber && \hspace{0.3in} s.t.  ~~ \norm{\vect{g} - \matr{M}\left[\matr{X}_1, \dots \matr{X}_n\right] \left[ \begin{array}{c}
\tilde{\vect{b}}_1 \\ \vdots \\ \tilde{\vect{b}}_n
\end{array}\right]}_2 \le \norm{\vect{e}}_2.
\end{eqnarray}

\ENDWHILE

\STATE Form an estimate for the original signal: $\hat{\vect{f}} = \left[\matr{X}_1, \dots, \matr{X}_n \right]\left[\hat{\vect{b}}_1^T, \dots,  \hat{\vect{b}}_n^T \right]^T$.

\end{algorithmic}
\end{algorithm}

{\small
\bibliographystyle{IEEEtran}
\bibliography{LineFitting}
}

\end{document}